\documentclass{article} 
\usepackage{iclr2027_conference,times}
\iclrpreprint

\usepackage{amsmath,amsfonts,bm}

\def\eqref#1{equation~\ref{#1}}

\def\1{\bm{1}}

\DeclareMathAlphabet{\mathsfit}{\encodingdefault}{\sfdefault}{m}{sl}
\SetMathAlphabet{\mathsfit}{bold}{\encodingdefault}{\sfdefault}{bx}{n}

\usepackage{comment}
\usepackage{rotating}
\usepackage{etoc}
\usepackage{csvsimple}
\usepackage{pgf}
\usepackage{booktabs}
\usepackage{multirow}
\usepackage[table]{xcolor}
\usepackage{tikz}
\usepackage{amssymb}
\usetikzlibrary{arrows.meta,calc,positioning,shapes.geometric,backgrounds}
\usetikzlibrary{decorations.pathreplacing}
\usepackage{pgfplots}
\usepackage{caption}
\usepackage{subcaption}
\usepackage{graphicx}
\pgfplotsset{compat=1.18}
\usepackage{xcolor}
\usepackage{array}
\usepackage{tabularx}
\usepackage{wrapfig}
\usepackage[table]{xcolor}
\usepackage{pgfplots}
\usepackage{pgfplotstable}
\usepackage{multirow}
\usepackage[export]{adjustbox}
\usepackage{xcolor}
\usepackage{pgf}
\usepackage{etoolbox}
\usepackage{csvsimple}
\usepgfplotslibrary{groupplots}
\usepackage{xfp}
\usepackage{hyperref}
\usepackage{url}
\usepackage{graphicx}
\usepackage{amsthm}
\usepackage{booktabs}
\usepackage{float}
\usepackage{multirow}
\usepackage{subcaption}
\usepackage{tcolorbox}
\newtheorem{theorem}{Theorem}[section]

\usepackage{algorithm}
\usepackage{algpseudocode}

\definecolor{gtgreen}{RGB}{55,140,70}
\definecolor{advred}{RGB}{190,55,45}
\definecolor{otherblue}{RGB}{45,105,170}

\definecolor{maniBorder}{RGB}{80,170,100}

\definecolor{origGreen}{RGB}{60, 140, 80}
\definecolor{isoBlue}{HTML}{526F91}
\definecolor{maniPurple}{RGB}{125, 85, 165}

\definecolor{manigrey}{gray}{0.92}
\definecolor{resultgreen}{RGB}{55,150,65}

\definecolor{gaincolor}{RGB}{55,160,75}
\definecolor{losscolor}{RGB}{130,130,130}
\definecolor{mutedred}{RGB}{170,85,85}

\newcommand{\upgain}[1]{%
    {\scriptsize\textcolor{gaincolor}{#1 $\uparrow$}}%
}

\newcommand{\downgain}[1]{%
    {\scriptsize\textcolor{mutedred}{#1$\downarrow$}}%
}

\newcommand{\retimg}[1]{%
\includegraphics[
    width=0.105\textwidth,
    height=1.5cm,
    keepaspectratio
]{#1}
}

\title{Randomized Manifold Smoothing for Formal Robustness to Semantic Perturbations}

\usepackage{authblk}

\author{%
  Daniya Niazi$^{1}$ \quad
  Alaa Anani$^{2,3}$ \quad
  Bernt Schiele$^{2}$ \quad
  Jonas Fischer$^{2}$ \\
  {\normalfont
  $^{1}$Saarland University, Saarland Informatics Campus, Germany \\
  $^{2}$Max Planck Institute for Informatics, Saarland Informatics Campus, Saarbr\"ucken, Germany \\
  $^{3}$CISPA Helmholtz Center for Information Security, Saarbr\"ucken, Germany}
}

\definecolor{jonasorange}{rgb}{0.75, 0.38, 0.0}
\definecolor{crimson}{rgb}{0.6, 0.0, 0.1}
\definecolor{berntblue}{rgb}{0.08, 0.38, 0.74}

\newcommand{\norm}[1]{\lVert #1 \rVert}
\iclrfinalcopy 
\begin{document}

\maketitle
\lhead{Preprint. Under review.}   



\begin{abstract}
Model predictions can be sensitive to small, unrelated variations in the input, which questions the reliability of the model.
Randomized smoothing provides robustness guarantees against small, adversarial perturbations, giving a certifiably robust predictor for such perturbations.
However, these perturbations are applied across all directions of the ambient space, irrespective of the underlying data mani\-fold. 
Consequently, its guarantees largely cover perturbations of the input that are not seen in reality rather than semantically meaningful variations in the data. 
To address this, we introduce \textbf{Randomized Manifold Smoothing}, which adapts the smoothing distribution to local data geometry by shaping Gaussian perturbations along principal directions of a locally estimated subspace. 
This yields anisotropic certificates over locally supported variations while retaining the statistical certification machinery of randomized smoothing. 
We evaluate manifold smoothing across image classification, semantic segmentation, interpretable concept representations, and vision-language retrieval tasks, including semantic distribution shifts. Across these settings, we show that manifold smoothing consistently preserves task performance while providing stronger robustness guarantees along semantically meaningful perturbations compared to standard randomized smoothing in high-dimensional data and representation spaces.

\end{abstract}

\section{Introduction}

Machine learning models often achieve strong predictive performance, yet accuracy alone does not guarantee that their predictions remain stable at deployment, where small changes in data acquisition, appearance, or representation can alter a model's prediction. 
For safety-critical applications,
it is therefore important
to establish not only whether a prediction is correct, but whether it is
\emph{provably stable} under natural variations of the input.

Randomized smoothing (RS) provides scalable, provable robustness guarantees
\citep{lecuyer2019certified,cohen2019certified} by certifying predictions
within an $\ell_2$ ball around a given input using isotropic Gaussian perturbations. 
While well-suited for norm-bounded adversarial robustness, this isotropic certified region around the input is a poor match for natural variations in the data, i.e., the semantically meaningful changes of the kind observed between real samples. Randomized smoothing perturbs inputs in a $d$-dimensional space, either the raw input space (e.g., pixels) or a learned representation space (e.g., CLIP embeddings), which we refer to as the \emph{ambient space} $\mathbb{R}^d$. Here, data typically lie near a much lower-dimensional manifold \citep{fefferman2016testing,pope2021intrinsic}. Natural data variations follow this manifold, which means they are anisotropic in the ambient space: large along a few directions and negligible along most others.
Standard randomized smoothing using isotropic noise, however, perturbs all ambient space directions equally, which suits worst-case (adversarial) perturbations rather than natural variations. This creates a gap between \emph{what
randomized smoothing certifies} and \emph{the variations we seek robustness
against}. We therefore ask:

\begin{quote} \emph{Can we certify robustness to variations supported by the data distribution, rather than to arbitrary perturbations in the ambient space?} \end{quote} 

We introduce \textbf{Randomized Manifold Smoothing}, which adapts the smoothing distribution to the local data geometry (Fig.~\ref{fig:teaser}). Around each input $x$, we estimate local variation from neighboring data, identify its principal directions and scales, and shape Gaussian perturbations accordingly. This induces an input-dependent ellipsoidal geometry that stretches along directions of greater local variation and contracts along directions with little local support. Through a whitening transformation, we retain the statistical certification machinery of randomized smoothing while obtaining guarantees over this locally adapted geometry. Our main contributions are:
\begin{itemize}

    \item Introduction of \textbf{Randomized Manifold Smoothing}, a 
    framework for formal robustness certification against semantically
    meaningful perturbations.

    \item \textbf{Manifold-aligned ellipsoidal certificates.} We estimate
    the local geometry from neighboring data points to define an anisotropic smoothing distribution, and use whitening to certify an ellipsoidal region aligned with this geometry.


     \item \textbf{Empirical validation.} We evaluate manifold smoothing on image classification, semantic segmentation, sparse concept spaces, and vision--language retrieval, spanning pixel and representation spaces.

\end{itemize}

Across tasks, manifold smoothing matches isotropic smoothing at small noise and substantially outperforms it at larger noise, where isotropic smoothing collapses. 
Our results indicate that concentrating perturbations along data-supported directions instead of along all ambient directions, makes manifold smoothing an effective approach for \emph{certifying robustness to natural variations in the data}.

\input{tikz_fig/teaser/teaser17}

\section{Related Work}
\label{sec:related-work}

A robustness certificate is a worst-case guarantee of prediction invariance: given a base predictor $f$ and an input $x$, the certified predictor $g$ returns a radius $r$ such that $g(x')=g(x)$ for every $x'$ within distance $r$ of $x$, or abstains when no such $r$ can be established. Randomized smoothing (RS)~\citep{lecuyer2019certified,cohen2019certified} provides scalable black-box guarantees by perturbing the input at random, aggregating the resulting predictions, and turning how consistently the top class wins into a certified radius. The region this radius carves out around $x$, and thus the perturbations the certificate covers, has been explored along several lines.

\textbf{$\ell_p$ certification.} Standard Gaussian RS certifies an isotropic $\ell_2$-ball~\citep{cohen2019certified}, with subsequent work extending smoothing to other $\ell_p$ regions~\citep{lee2019tight,levine2020randomized,yang2020randomized}. These regions are fixed independently of $x$ and can be misaligned with the low-dimensional directions of natural data variations. Our region instead adapts its shape to the local data geometry.

\textbf{Input-dependent certification.} A second line relaxes isotropy by letting the smoothing covariance depend on the input. \citet{alfarra2022data} optimize a per-input noise variance, and ANCER~\citep{eiras2022ancer} generalizes this to full anisotropy through per-example volume maximization over ellipsoidal regions. \citet{sukenik2022intriguing} show that input-dependent variance can break soundness unless constrained to vary slowly in $x$. Unlike prior input-dependent methods, which optimize the smoothing covariance against the certificate, we estimate it non-parametrically from local geometry and keep it fixed at $x$ for certification; Theorems~\ref{thm:cov-convergence} \&~\ref{thm:main-ellipsoid-convergence} establish its convergence to the true tangent-space covariance.

\textbf{Transformation-specific certification.} A third line certifies structured, semantically meaningful changes using predefined transformation families~\citep{fischer2020certified,li2021tss,alfarra2022deformrs}. These guarantees are interpretable, but restricted to transformations fixed in advance. Our approach instead recovers the directions of variation directly from the local data density.

\textbf{Beyond classification.} RS has also been extended to structured outputs, including object detection~\citep{chiang2020detection}, general metric output spaces~\citep{kumar2021center}, semantic segmentation~\citep{Fischer2021ScalableSegmentation,anani2024adaptive}, attributions \citep{anani2025pixel} and circuit discovery \citep{anani2026certified}. These extensions concern \emph{what} is certified and are orthogonal to the geometry in which robustness is measured.

\textbf{Summary.} Existing RS certificates either measure $r$ in a fixed norm independently of the data, adapt the region by optimizing against the radius itself, or restrict it to a transformation family specified in advance. Manifold smoothing instead derives the region from the local geometry of the data: an ellipsoid aligned with the estimated tangent space at $x$, requiring neither a parametric family nor optimization of the smoothing distribution.
\section{Randomized Manifold Smoothing}

\begin{figure*}[t!]
    \centering
    \includegraphics[width=\textwidth]{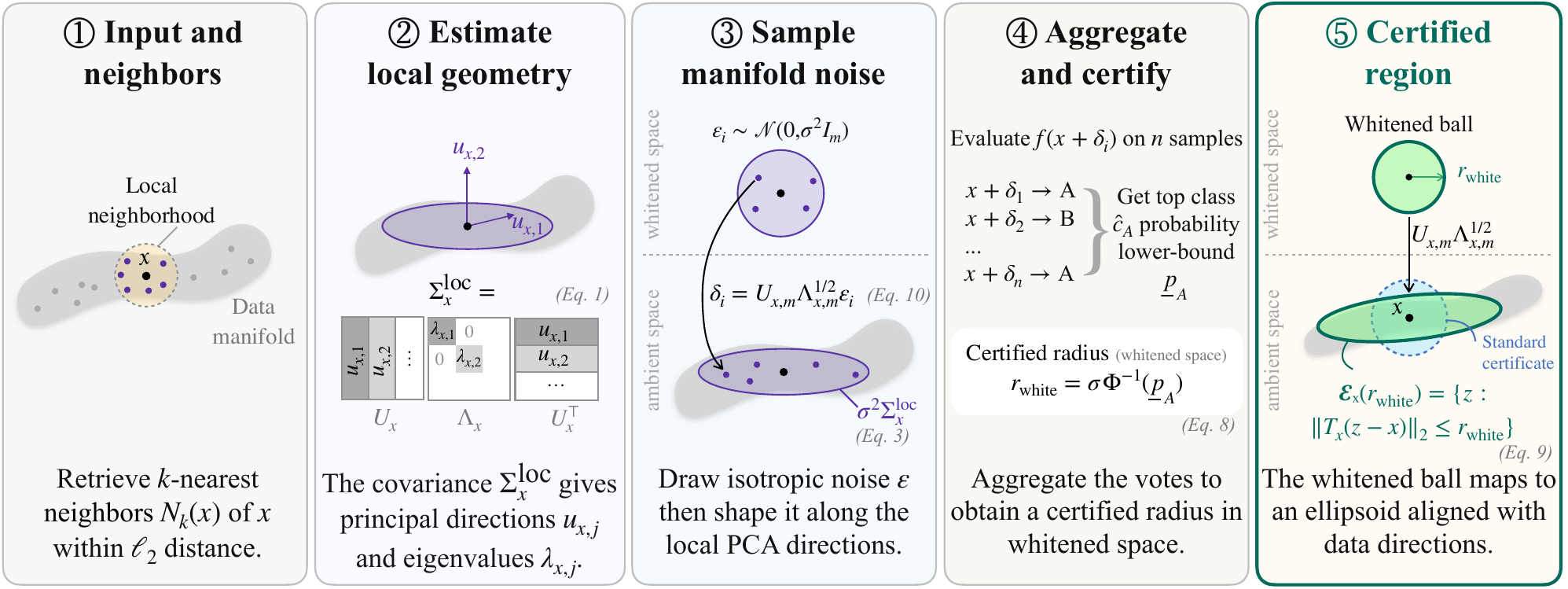}
\caption{ \textbf{Manifold-aligned certification through local geometry estimation. } Given an input $x$, (1) we retrieve its $k$ nearest neighbors $N_k(x)$, (2) estimate local geometry through principal component analysis (PCA) (Eq.~\ref{eq:covar}), and (3) sample isotropic Gaussian noise $\varepsilon_i$ in whitened space and shape it into $\delta_i$ along the local principal directions (Eq.~\ref{eq:sample}). (4) We then aggregate predictions over $n$ perturbed Monte Carlo samples $x + \delta_i$ and certify the predicted class $\hat{c}_A$ from the lower confidence bound $\underline{p}_A$, obtaining a certified radius $r_{\mathrm{white}}$ in whitened space (Eq.~\ref{eq:radius}). (5) Mapping the standard certified $\ell_2$-ball from the whitened space to the ambient space through the local covariance results in a certified region that is a manifold-aligned ellipsoid $\mathcal{E}_x(r_{\mathrm{white}})$ around $x$ (Eq.~\ref{eq:ellipsoid}).}

\label{fig:method}
\vspace{-4mm}
\end{figure*}

The central idea of our approach is to produce noise along directions reflecting the data geometry, shifting the goal from robustness against \textit{adversarial} perturbations to robustness against \textit{semantic} perturbations. 
Thus, instead of considering isotropic Gaussian noise, which might be added through a technical artifact or adversary but has no direct semantic meaning, we provide a sense for robustness to semantic change, such as a lesion in an X-ray being slightly larger or smaller. Here, we determine semantic changes in a data-driven manner, considering the variation along directions in the ambient space around the target sample $x$. Crucially, this lets us carry over the formalism of standard RS while modeling semantic variations, in the sense of natural image changes observed in the data.
We then model noise \textit{proportional to these observed variations}, drawing Gaussian noise around $x$ whose covariance is the local covariance of the $k$-neighborhood of $x$, scaled by a global noise level. Instead of an $\ell_2$-ball we now obtain an ellipsoid that stretches far along the principal directions with large eigenvalues, and little along those with small eigenvalues. Directions with larger eigenvalues correspond to greater observed local variation. Before we discuss this approach more formally, we introduce the necessary notation.

\subsection{Notation}
\label{sec:notation}

We consider inputs $\mathcal{X}=\mathbb{R}^d$, labels $\mathcal{Y}=\{1,\dots,C\}$, a training set $\mathcal{D}=\{(x_i,y_i)\}_{i=1}^{n_{\mathcal{D}}}\subset\mathcal{X}\times\mathcal{Y}$, and a base predictor $f:\mathcal{X}\to\mathcal{Y}$ to certify. For $x\in\mathcal{X}$, $N_k(x)=\{i_1,\dots,i_k\}$ indexes its $k$ nearest neighbors in ambient $\ell_2$ distance, which we assume approximates the manifold distance locally. The local mean, covariance, and eigen-decomposition of the covariance over this neighborhood are

\begin{equation}\label{eq:covar}
\mu_x = \frac{1}{k}\sum_{j=1}^k x_{i_j},
\qquad
\Sigma_x^{\text{loc}} = \frac{1}{k-1}\sum_{j=1}^k (x_{i_j}-\mu_x)(x_{i_j}-\mu_x)^\top = U_x \Lambda_x U_x^\top,
\end{equation}

where the columns $u_{x,1},\dots,u_{x,d}$ of the orthogonal matrix $U_x$ are the eigenvectors,
i.e., the local principal directions, and $\Lambda_x = \operatorname{diag}(\lambda_{x,1},\dots,\lambda_{x,d})$
holds the eigenvalues $\lambda_{x,1} \ge \dots \ge \lambda_{x,d} \ge 0$, i.e., the variances along
these directions; together they give a local linear approximation of the data manifold around $x$.

\subsection{Noise along the data manifold}

We now introduce the noise sampling along principal directions more formally. Let $\sigma > 0$ be a global scaling parameter for the noise level, as common in RS. In the local principal basis, we draw noise independently along each principal direction with variance proportional to its eigenvalue,
\begin{equation}\label{eq:noisepc}
\eta_j \sim \mathcal{N}(0,\ \sigma^2 \lambda_{x,j}), \quad j=1,\dots,d, 
    \qquad\text{that is} \qquad
    \eta \sim \mathcal{N}\left(0,\ \sigma^2 \Lambda_x\right).
\end{equation}

Rotating $\eta$ back to the original (pixel) coordinates with $U_x$ gives
$\delta_x = U_x \eta \sim \mathcal{N}(0,\Sigma_x)$,
where $\Sigma_x = U_x (\sigma^2 \Lambda_x) U_x^\top$.
Thus, the noise covariance is a scaled copy of the local empirical covariance,

\begin{equation}\label{eq:noise}
\delta_x \sim \mathcal{N}\big(0,\ \sigma^2 \Sigma_x^{\text{loc}}\big).
\end{equation}

Looking at Eq.~\ref{eq:noisepc}, the per-direction variance $\sigma^2\lambda_{x,j}$ ties the noise to the local data variations: perturbations are large along high-variance principal directions and vanish along directions where the neighbors barely differ. Since off-manifold directions carry near-zero eigenvalues, the noise stays close to the manifold by construction. Using this noise, we define the manifold-smoothed classifier

\begin{equation}\label{eq:smoothpred}
g_x(z) = \arg\max_{c \in \mathcal{Y}} \pi_{x,c}(z),
\qquad
\pi_{x,c}(z) =
\Pr_{\delta_x \sim \mathcal{N}(0,\,\sigma^2\Sigma_x^{\text{loc}})}
\big[f(z+\delta_x) = c\big].
\end{equation}

This is the exact analogue of standard RS~\citep{cohen2019certified}
with the isotropic covariance $\sigma^2 I$ replaced by the \textit{anisotropic} $\sigma^2\Sigma_x^{\text{loc}}$. The noise is shaped by the local PCA on the $k$NNs of $x$. Lifting the formal guarantees to anisotropic noise is discussed next.

\subsection{Formal guarantees through whitening}

To directly transfer the formal guarantees of vanilla RS, we need isotropic noise.
We achieve this through coordinate transformation by \textit{whitening} with respect to the local covariance.

\textbf{PCA-based whitening.} \label{sec:whitening}
To obtain isotropic noise, we map points $x'$ near $x$ to whitened coordinates
\begin{equation}\label{eq:whitening}
\widetilde{x}' = T_x (x' - x),
\qquad
T_x = \Lambda_x^{-1/2} U_x^\top.
\end{equation}
Since the empirical covariance may be rank-deficient, in practice we retain the leading \(m\) principal directions and floor their eigenvalues for numerical stability, performing whitening within the retained subspace (Sec. \ref{sec:algorithm}).
Now we analyze what happens to noise $\delta_x \sim \mathcal{N}(0,\ \sigma^2 \Sigma_x^{\text{loc}})$ in this space.
The whitened noise is
\[
   \widetilde{\delta}_x = T_x \delta_x
   = \Lambda_x^{-1/2} U_x^\top \delta_x\;.
\]

Since $\delta_x = U_x \eta,\ \eta \sim \mathcal{N}(0,\ \sigma^2 \Lambda_x)$, we thus get
$\widetilde{\delta}_x = \Lambda_x^{-1/2} U_x^\top U_x \eta = \Lambda_x^{-1/2} \eta$.
As $\eta \sim \mathcal{N}(0,\ \sigma^2 \Lambda_x)$ we have
$\widetilde{\delta}_x \sim \mathcal{N}(0,\ \sigma^2 I)$,
meaning that in whitened coordinates the \textit{noise is isotropic Gaussian}, allowing standard RS guarantees in the whitened space.

\textbf{Smoothed classifier.} With the whitened classifier $\widetilde{f}_x(\widetilde{x}') := f\big(x + U_x \Lambda_x^{1/2} \widetilde{x}'\big)$, the smoothed classifier becomes 
\begin{equation}\label{eq:smoothwhite}
g_x(z)
= \arg\max_{c \in \mathcal{Y}}~
\Pr_{\widetilde{\delta}_x \sim \mathcal{N}(0,\ \sigma^2 I)}
\big[ \widetilde{f}_x(T_x(z-x)+\widetilde{\delta}_x) = c \big]\;.
\end{equation}
Hence, PCA–eigenvalue–scaled smoothing collapses to standard RS, \textit{but in coordinates that are whitened} by the local covariance.

\textbf{The induced certified region.}
\label{sec:certified-region}
Following standard RS~\citep{cohen2019certified}, the guarantee in the whitened coordinates is an $\ell_2$-ball around the origin, which translates to an \textit{ellipsoid} around $x$ in the ambient space. Fix a candidate class $c_A \in \mathcal{Y}$, and let $\underline{p}_A$ be a lower bound on $\Pr[\widetilde f_x(\widetilde\delta_x) = c_A]$ and $\overline{p}_B$ an upper bound on the runner-up class probability. When $\underline{p}_A \geq \overline{p}_B$, the standard RS (with noise variance $\sigma^2$) certified radius in the whitened space is
\begin{equation}\label{eq:radius_general}
r_{\text{white}}
= \frac{\sigma}{2}
\big(\Phi^{-1}(\underline{p}_A)-\Phi^{-1}(\overline{p}_B)\big),
\end{equation}
where $\Phi^{-1}$ denotes the inverse standard normal CDF.
In practice, we follow the one-sided \textsc{Certify} procedure of
\citet{cohen2019certified}: we estimate only the lower confidence bound
$\underline{p}_A$ and set $\overline{p}_B = 1-\underline{p}_A$, a valid upper bound on the runner-up since class probabilities sum to one. With $\Phi^{-1}(1-p) = -\Phi^{-1}(p)$, the radius reduces to
\begin{equation}\label{eq:radius} r_{\text{white}} = \sigma\,\Phi^{-1}(\underline{p}_A), \end{equation} which is positive exactly when $\underline{p}_A > 0.5$; otherwise we abstain. Hence, smoothing $f$ for any $z = x + U_x\Lambda_x^{1/2}\widetilde{\delta}$ with $\norm{\widetilde{\delta}}_2 \le r_{\text{white}}$ using the noise of Eq.~\ref{eq:noise}, i.e., with the covariance fixed at $x$, still predicts $c_A$ as the reference input $x$.
%
The certificate above in Eq.~\ref{eq:radius} is defined in whitened coordinates.
In the retained PCA
subspace used for smoothing (Eq.~\ref{eq:sample}),
let $U_{x,m}$ and $\Lambda_{x,m}$ denote the retained $m$
eigenvectors and corresponding eigenvalues, respectively.
Restricting $z-x$ to this subspace, the condition
$\|T_x(z-x)\|_2 \le r_{\mathrm{white}}$ is equivalently
expressed by $z=x+U_{x,m}\Lambda_{x,m}^{1/2}v$,
with $\|v\|_2\le r_{\mathrm{white}}$.
The resulting certified region is

\begin{equation}\label{eq:ellipsoid}
\mathcal{E}_x(r_{\mathrm{white}})
=
\left\{
z=x+U_{x,m}\Lambda_{x,m}^{1/2}v
:
\|v\|_2\le r_{\mathrm{white}}
\right\}.
\end{equation}

This is an ellipsoid centered at $x$ with principal axes $u_{x,j}$ within the retained PCA subspace  and semi-axis lengths $r_{\text{white}}\sqrt{\lambda_{x,j}}$ for $j=1,\ldots,m$. The certificate thus extends far along directions in which the neighbors of $x$ vary strongly and contracts along directions with little local variation, i.e., it is a Mahalanobis ball in the metric of the inverse local covariance. As in standard RS, the smoothing distribution is held fixed while the input moves, rather than being re-estimated at $z$. Thus, the resulting certification guarantee is:

\begin{equation}\label{eq:rms-guarantee}
c_A=g_x(x)=g_x(z),
\qquad
\forall z\in\mathcal{E}_x(r_{\mathrm{white}}).
\end{equation}
App.~\ref{sec:ellipsoid-convergence} further discusses the certified region for the empirical ellipsoid estimated from the neighborhood of $x$.

\subsection{Estimating the Manifold-Smoothed Classifier}
\label{sec:algorithm}

Fig.~\ref{fig:method} (Algorithm in  App.~\ref{app:pseudocode}) illustrates the manifold certification procedure. Since the probabilities in Eq.~\ref{eq:smoothpred} are intractable, we estimate them by Monte Carlo sampling. 
Following the \textsc{Certify} procedure of \citet{cohen2019certified}, given $x$, we retrieve its $k$ nearest neighbors $N_k(x)$ (Fig.~\ref{fig:method} \textcircled{1}), compute the local covariance (Fig.~\ref{fig:method} \textcircled{2}) $\Sigma^{\mathrm{loc}}_{x}$ (Eq.~\ref{eq:covar}), and eigendecompose it, keeping the top-$m$ pair $(U_{x,m},\Lambda_{x,m})$.  Noise is then drawn as (Fig.~\ref{fig:method} \textcircled{3})
\begin{equation}\label{eq:sample}
\delta_i = U_{x,m}\Lambda_{x,m}^{1/2}\,\varepsilon_i,
\qquad
\varepsilon_i \sim \mathcal{N}(0,\sigma^2 I_m),
\end{equation}
and $f$ is evaluated on each $x+\delta_i$ (Fig.~\ref{fig:method} \textcircled{4}). Each $\varepsilon_i$ is a draw of the whitened noise $\widetilde{\delta}_x$ of Sec.~\ref{sec:whitening}, restricted to the retained $m$ directions, so shaping it is equivalent to sampling $\delta_x$ directly. A one-sided binomial test certifies the top class $\hat{c}_A$ when its lower confidence bound exceeds a threshold $\tau\in[0.5,1)$ with confidence $1-\alpha$, and abstains otherwise ; with $c_A = \hat{c}_A$, the certified region is then the ellipsoid (Fig.~\ref{fig:method} \textcircled{5})  in Eq.~\ref{eq:ellipsoid} with radius $r_{\mathrm{white}}$ from Eq.~\ref{eq:radius} for classification ($\tau = 0.5$), and the fixed radius $\sigma\,\Phi^{-1}(\tau)$ for segmentation~\citep{Fischer2021ScalableSegmentation}, where $\underline{p}_A$ is the one-sided lower confidence bound on $\pi_{\hat{c}_A}(x)$. Since only the sampling distribution differs from standard RS (Eq.~\ref{eq:smoothwhite}), the statistical test is unchanged, and the guarantee transfers directly.
\setlength{\textfloatsep}{4pt plus 1pt minus 1pt}
\setlength{\intextsep}{4pt plus 1pt minus 1pt}
\setlength{\floatsep}{4pt plus 1pt minus 1pt}

\newtcolorbox{takeawaybox}{
    colback=maniPurple!8,
    colframe=maniPurple,
    boxrule=0pt,
    leftrule=3pt,
    arc=0pt,
    left=6pt,
    right=6pt,
    top=5pt,
    bottom=5pt
}

\newcommand{\takeaway}[1]{%
\begin{takeawaybox}
#1
\end{takeawaybox}
}

\definecolor{gaincolor}{RGB}{55,160,75}
\definecolor{losscolor}{RGB}{130,130,130}
\definecolor{mutedred}{RGB}{170,85,85}

\begin{table*}[t]
\centering
\caption{
\textbf{Certified accuracy across representative tasks}
using the certification protocols of \citet{cohen2019certified}
and \citet{Fischer2021ScalableSegmentation}.
Subscripts $\mathrm{I}$ and $\mathrm{M}$ denote isotropic and manifold
smoothing, respectively. $\Delta$ denotes the absolute difference between
manifold and isotropic certified accuracy. 
}
\label{tab:main_certification}

\tiny
\setlength{\tabcolsep}{0pt}
\renewcommand{\arraystretch}{1.12}

\resizebox{\textwidth}{!}{%
\begin{tabular}{
@{}
c
@{\hspace{10pt}}
c@{\hspace{3pt}}c@{\hspace{3pt}}c
@{\hspace{12pt}}
c@{\hspace{3pt}}c@{\hspace{3pt}}c
@{\hspace{18pt}}
c@{\hspace{3pt}}c@{\hspace{3pt}}c
@{\hspace{18pt}}
c@{\hspace{3pt}}c@{\hspace{3pt}}c
@{}
}

\toprule

\multirow{2}{*}{$\sigma$}
&
\multicolumn{3}{c}{\textbf{Classification} (CelebA)}
&
\multicolumn{3}{c}{\textbf{Classification} (CelebA-HQ)}
&
\multicolumn{3}{c}{\textbf{Segmentation} (CelebAMask-HQ)}
&
\multicolumn{3}{c}{\textbf{Concept Space} (ImageNet)}
\\

\cmidrule(lr){2-4}
\cmidrule(lr){5-7}
\cmidrule(lr){8-10}
\cmidrule(lr){11-13}

&
$\mathrm{CAcc}_{\mathrm I}\uparrow$
&
$\mathrm{CAcc}_{\mathrm M}\uparrow$
&
$\Delta$
&
$\mathrm{CAcc}_{\mathrm I}\uparrow$
&
$\mathrm{CAcc}_{\mathrm M}\uparrow$
&
$\Delta$
&
$\mathrm{CAcc}_{\mathrm I}\uparrow$
&
$\mathrm{CAcc}_{\mathrm M}\uparrow$
&
$\Delta$
&
$\mathrm{CAcc}_{\mathrm I}\uparrow$
&
$\mathrm{CAcc}_{\mathrm M}\uparrow$
&
$\Delta$
\\

\midrule

0.02
&
\textbf{93.0}
&
92.0
&
\downgain{1.0}
&
\textbf{94.0}
&
93.0
&
\downgain{1.0}
&
94.30
&
\textbf{95.09}
&
\upgain{0.79}
&
67.08
&
\textbf{68.08}
&
\upgain{1.00}
\\

0.05
&
87.0
&
\textbf{92.0}
&
\upgain{5.0}
&
91.0
&
\textbf{92.0}
&
\upgain{1.0}
&
92.82
&
\textbf{94.74}
&
\upgain{1.92}
&
64.00
&
\textbf{68.06}
&
\upgain{4.06}
\\

0.10
&
72.0
&
\textbf{92.0}
&
\upgain{20.0}
&
83.0
&
\textbf{92.0}
&
\upgain{9.0}
&
90.46
&
\textbf{93.99}
&
\upgain{3.53}
&
45.8
&
\textbf{68.6}
&
\upgain{22.8}
\\

0.50
&
45.0
&
\textbf{92.0}
&
\upgain{47.0}
&
50.0
&
\textbf{93.0}
&
\upgain{43.0}
&
26.87
&
\textbf{84.27}
&
\upgain{57.40}
&
0.0
&
\textbf{63.8}
&
\upgain{63.8}
\\

1.00
&
45.0
&
\textbf{86.0}
&
\upgain{41.0}
&
50.0
&
\textbf{86.0}
&
\upgain{36.0}
&
26.99
&
\textbf{67.77}
&
\upgain{40.78}
&
0.0
&
\textbf{47.2}
&
\upgain{47.2}
\\

\bottomrule
\end{tabular}%
}

\end{table*}

\section{Experimental Setup}
\textbf{Certification.}
Across all experiments, we compare manifold smoothing against standard
isotropic randomized smoothing, under identical certification settings.
For classification, we follow the \textsc{Certify} procedure of ~\citet{cohen2019certified}, while for semantic
segmentation we adapt \textsc{SegCertify}~\citep{Fischer2021ScalableSegmentation}
for pixel-wise certification. For certification, we use the standard $\underline{p}_A>0.5$ criterion for classification and $\tau=0.75$ for semantic segmentation. We use $n_0=30$ Monte Carlo samples for class selection,
$n=100$ for certification, and failure probability $\alpha=0.001$. For manifold smoothing, we retrieve $k=500$ nearest neighbors via
approximate nearest-neighbor search and estimate the local PCA basis $m$  as $min(m,d)$, $U$, and the covariance spectrum $\Lambda$. For the main experiments, we retain the full PCA basis of selected neighbours $m=k=500$ and use
the resulting covariance to shape Gaussian noise, and perturb the input as $x'=x+\delta_i$. We use the same global noise scale $\sigma$ for
isotropic and manifold smoothing; for manifold smoothing, $\sigma$ scales
the locally estimated covariance $\Sigma_x^{\mathrm{loc}}$. Main certification results use a randomly selected
subset of 100 test examples. We ablate neighborhood size $k$, retained dimension $m$, Monte Carlo sample size $n$, and test subset size $N$ in App. ~\ref{app:ablation_study}.

\textbf{Models and datasets.} 
For image classification, we evaluate ResNet-18 smile prediction on
CelebA and CelebA-HQ, including distribution shifts on
CelebA. For segmentation, we use pretrained
BiSeNet~\citep{yu2018bisenet} on CelebAMask-HQ. We apply smoothing only at certification time, without model retraining, to isolate perturbation geometry. For concept-space classification, we
evaluate a pretrained ImageNet classifier on a smooth sparse SAE concept
representation (CFM~\citep{Wittenmayer2026CFM}). For image-to-text
retrieval, we smooth CLIP ViT-B/32 image embeddings and evaluate on
RoCOCO~\citep{Park2024RoCOCO}. Implementation details are provided in
App. Sec.~\ref{app:experiment_details}.


\textbf{Evaluation metrics.}
We evaluate robustness with task-specific metrics. For classification, we report certified accuracy
($\mathrm{CAcc}\uparrow$) that is, the fraction of examples that are both
correctly classified and certified, and abstention rate
($\varnothing\downarrow$). For semantic segmentation, pixel-wise CAcc,
certified mIoU ($\mathrm{mIoU}\uparrow$), and pixel abstention rate
($\varnothing\downarrow$), the fraction of pixels for which no certified
prediction is returned.
For image-to-text retrieval, we report Recall@1
($\mathrm{R@1}\uparrow$), Retrieval Safety Misalignment Score
($\mathrm{RSMS}\downarrow$), and retrieval drop
($\mathrm{Drop}\downarrow$). Subscripts $\mathrm I$ and $\mathrm M$
denote isotropic and manifold smoothing. Evaluation metrics details are discussed in
App. Sec.~\ref{app:evaluation}.

\section{Empirical results}

We evaluate manifold smoothing across four tasks, each probing an aspect of \textbf{robustness certification} under meaningful
\textbf{semantic perturbations} across different \textbf{representation spaces}. Hence, we investigate:
(i) \textit{Image Classification}: how far can the perturbation magnitude
increase while preserving certified predictive structure, including under
semantic distribution shift? (Sec.~\ref{sec:classification});
(ii) \textit{Semantic Segmentation}: do these guarantees extend from
image-level predictions to dense, pixel-level certification?
(Sec.~\ref{sec:segmentation});
(iii) \textit{Interpretable Concept Representations}: does smoothing along
the local manifold preserve semantic structure in sparse concept spaces?
(Sec.~\ref{sec:concepts}); and
(iv) \textit{Image-to-Text Retrieval}: does preserving local geometry
maintain semantic neighborhoods and cross-modal alignment in multimodal
embedding spaces? (Sec.~\ref{sec:rococo_main}).

\subsection{OOD-Robust Image Classification}
\label{sec:classification}

\definecolor{oodteal}{HTML}{1B9E8A}
\definecolor{oodorange}{HTML}{D95F02}
\definecolor{oodpurple}{HTML}{7570B3}
\definecolor{oodpink}{HTML}{E7298A}
\definecolor{oodgreen}{HTML}{66A61E}
\definecolor{oodblue}{HTML}{1F78B4}

\newcommand{\oodPlot}[3]{%
   
    \ifnum#1=0\relax
    \def\oodDataFile{data/ood_certification_results.csv}%
    \else
        \def\oodDataFile{data/non_ood_certification_results.csv}%
    \fi
    \addplot[
        thin,
        mark=*,
        mark size=1pt,
        #3,
        restrict expr to domain={\thisrow{SmoothID}}{#1:#1},
        restrict expr to domain={\thisrow{AttrID}}{#2:#2},
        unbounded coords=discard,
    ]
    table[
        col sep=comma,
        x=sigma,
        y expr={100*\thisrow{CertAcc}},
    ] {\oodDataFile};
}

\newcommand{\idPlot}[2]{%
    \addplot[
        line width=1.0pt,
        densely dotted,
        no marks,
        #2,
        restrict expr to domain={\thisrow{SmoothID}}{#1:#1},
        restrict expr to domain={\thisrow{AttrID}}{0:0},
        unbounded coords=discard,
    ]
    table[
        col sep=comma,
        x=sigma,
        y expr={100*\thisrow{CertAcc}},
    ] {data/ood_certification_results.csv};
}

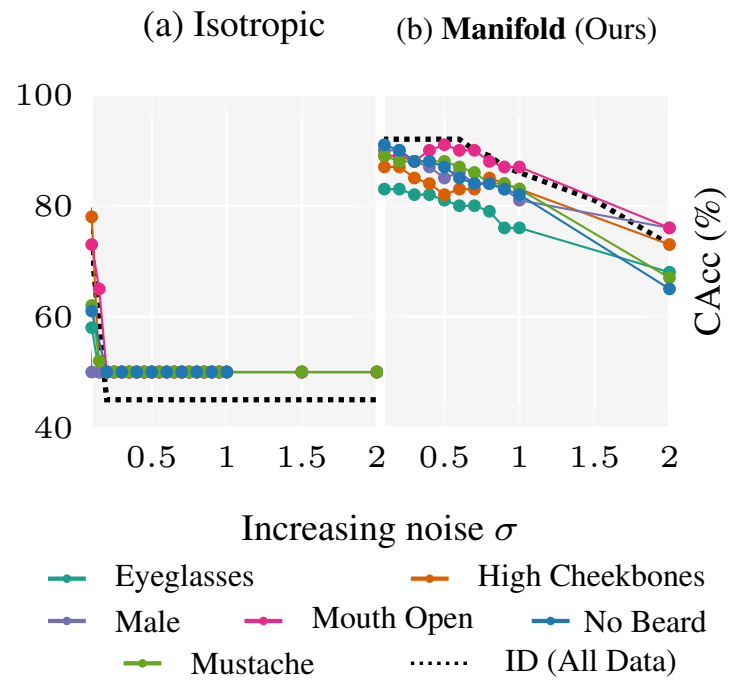
\begin{wrapfigure}{r}{0.35\textwidth}
\vspace{-6pt}
\centering

\begin{tikzpicture}


\begin{axis}[
    name=isoaxis,
    width=0.135\textwidth,
    height=2.2cm,
    scale only axis,
    xmin=0.1,
    xmax=2.0,
    ymin=40,
    ymax=100,
    xtick={0.5,1.0,1.5,2.0},
    ytick={40,60,80,100},
    xlabel={},
    ylabel={},
    title={(a) Isotropic},
    axis background/.style={fill=gray!8},
    grid=major,
    major grid style={draw=white,line width=0.5pt},
    axis line style={draw=white},
    tick style={draw=gray!10},
    tick label style={font=\tiny},
    label style={font=\scriptsize},
    title style={font=\scriptsize},
]

\idPlot{0}{black}

\oodPlot{0}{1}{oodteal}
\oodPlot{0}{2}{oodorange}
\oodPlot{0}{3}{oodpurple}
\oodPlot{0}{4}{oodpink}
\oodPlot{0}{5}{oodgreen}
\oodPlot{0}{6}{oodblue}

\end{axis}


\begin{axis}[
    name=maniaxis,
    at={(isoaxis.east)},
    anchor=west,
    xshift=0.5mm,
    width=0.135\textwidth,
    height=2.2cm,
    scale only axis,
    xmin=0.1,
    xmax=2.0,
    ymin=40,
    ymax=100,
    xtick={0.5,1.0,1.5,2.0},
    ytick={40,60,80,100},
    yticklabels=\empty,
    xlabel={},
    ylabel={},
    title={(b) \textbf{Manifold} (Ours)},
    axis background/.style={fill=gray!8},
    grid=major,
    major grid style={draw=white,line width=0.5pt},
    axis line style={draw=white},
    tick style={draw=gray!10},
    tick label style={font=\tiny},
    label style={font=\tiny},
    title style={font=\tiny},
]

\idPlot{1}{black}

\oodPlot{1}{1}{oodteal}
\oodPlot{1}{2}{oodorange}
\oodPlot{1}{3}{oodpurple}
\oodPlot{1}{4}{oodpink}
\oodPlot{1}{5}{oodgreen}
\oodPlot{1}{6}{oodblue}

\end{axis}


\node[
    font=\scriptsize,
    anchor=north
]
at ([yshift=-4.5mm]$(isoaxis.south)!0.5!(maniaxis.south)$)
{Increasing noise $\sigma$};


\node[
    rotate=90,
    font=\scriptsize,
    anchor=south
]
at ([xshift=5.0mm]maniaxis.east)
{CAcc (\%)};


\coordinate (legendbase) at
([yshift=-10mm]$(isoaxis.south west)!0.5!(maniaxis.south east)$);

\def\legline{1mm}


\draw[oodteal,thick]
([xshift=-22mm]legendbase)
--
([xshift=-19.5mm]legendbase);
\fill[oodteal]
([xshift=-20.75mm]legendbase) circle (0.8pt);
\node[anchor=west,font=\tiny]
at ([xshift=-18.8mm]legendbase)
{Eyeglasses};

\draw[oodorange,thick]
([xshift=2mm]legendbase)
--
([xshift=4.5mm]legendbase);
\fill[oodorange]
([xshift=3.25mm]legendbase) circle (0.8pt);
\node[anchor=west,font=\tiny]
at ([xshift=5.2mm]legendbase)
{High Cheekbones};


\coordinate (legendrowtwo) at
([yshift=-2.8mm]legendbase);

\draw[oodpurple,thick]
([xshift=-22mm]legendrowtwo)
--
([xshift=-19.5mm]legendrowtwo);
\fill[oodpurple]
([xshift=-20.75mm]legendrowtwo) circle (0.8pt);
\node[anchor=west,font=\tiny]
at ([xshift=-18.8mm]legendrowtwo)
{Male};

\draw[oodpink,thick]
([xshift=-9mm]legendrowtwo)
--
([xshift=-6.5mm]legendrowtwo);
\fill[oodpink]
([xshift=-7.75mm]legendrowtwo) circle (0.8pt);
\node[anchor=west,font=\tiny]
at ([xshift=-5.8mm]legendrowtwo)
{Mouth Open};

\draw[oodblue,thick]
([xshift=10mm]legendrowtwo)
--
([xshift=12.5mm]legendrowtwo);
\fill[oodblue]
([xshift=11.25mm]legendrowtwo) circle (0.8pt);
\node[anchor=west,font=\tiny]
at ([xshift=12.2mm]legendrowtwo)
{No Beard};


\coordinate (legendrowthree) at
([yshift=-5.6mm]legendbase);

\draw[oodgreen,thick]
([xshift=-17mm]legendrowthree)
--
([xshift=-14.5mm]legendrowthree);
\fill[oodgreen]
([xshift=-15.75mm]legendrowthree) circle (0.8pt);
\node[anchor=west,font=\tiny]
at ([xshift=-13.8mm]legendrowthree)
{Mustache};

\draw[
    black,
    densely dotted,
    line width=0.65pt
]
([xshift=2mm]legendrowthree)
--
([xshift=6mm]legendrowthree);

\node[anchor=west,font=\tiny]
at ([xshift=7mm]legendrowthree)
{ID (All Data)};

\end{tikzpicture}


\captionsetup{
    font=scriptsize,
    labelfont=bf,
    skip=2pt
}

\caption{
\textbf{Certified robustness under attribute-OOD CelebA shifts.}
The dotted curve denotes the in-distribution certified accuracy.
}
\label{fig:ood_certification}

\vspace{-8pt}
\end{wrapfigure}

\textbf{Robustness guarantees.}
\textit{Can manifold smoothing provide guarantees at larger, semantically
meaningful perturbation levels?}
At small noise ($\sigma=0.02$), Tab. \ref{tab:main_certification} shows that, both smoothers achieve comparable certified accuracy on CelebA and Celeb-HQ. At $\sigma=1.0$, manifold smoothing retains $86\%$ (\textcolor{origGreen}{$\uparrow91.1\%$}) certified accuracy
versus $45\%$ in isotropic collapsing to "No-Smile" prediction; we analyzed this in App. Sec. \ref{app:class_wise_celeba}, showing that larger perturbations can be certified along the data manifold without unnecessarily extending the certified region into off-manifold directions. We provide extended noise scale perturbation results in
App. Sec.~\ref{app:noise_sweeps_all} 

\input{tikz_fig/celeba_main_certification}

\textbf{Perturbation structure.}
\textit{What do these larger perturbations correspond to in image
space?}
Fig.~\ref{fig:celeba_noise_sweep} reveals how 
isotropic noise produces  diffuse
corruption while manifold
perturbations instead induce coherent variations in \textit{altered smiles, open/closed mouths, eye shape and color, skin tone, hair, and lip shape}, while
preserving the meaningful facial structure at even higher perturbations. This highlights the contrast between the two
approaches: randomized smoothing was designed to provide robustness
guarantees in settings typical of adversarial perturbations, whereas here
we seek guarantees when semantically meaningful perturbations occur.

\textbf{Robustness under semantic distribution shift.}
\textit{Do guarantees over semantic variations persist under semantic
distribution shifts?}
We evaluate six attribute-defined CelebA shifts by training with a target
attribute-absent ($a=0$) and evaluating where it is present ($a=1$). Similarly, the local covariance comes from the attribute-absent ($a=0$) neighborhood.
Fig.~\ref{fig:ood_certification} shows that manifold smoothing preserves
non-trivial certification across all OOD shifts over larger
perturbations. Relative to the in-distribution reference, the OOD curves
exhibit varying degradation, most prominently for \textit{Eyeglasses, High Cheekbones and No Beard},
showing that the underlying distribution shift remains challenging. Nevertheless, manifold smoothing robustness guarantees persist under OOD shift.

\begin{takeawaybox}
Manifold smoothing provides better \textbf{robustness guarantees}, preserves \textbf{semantic coherence} across
\textit{larger perturbations},
and remains \textbf{robustly certifiable} under
\textit{semantic distribution shift}.
\end{takeawaybox}

\subsection{Image Segmentation}
\label{sec:segmentation}

\begin{figure}[h]
\centering

\begin{minipage}[t]{0.59\textwidth}
    \vspace{0pt}

    \hspace*{-2mm}%
    \resizebox{\linewidth}{!}{%

\begingroup

\setlength{\tabcolsep}{0pt}
\renewcommand{\arraystretch}{1.0}

\definecolor{maniBorder}{RGB}{80,170,100}

\newcommand{\segimage}[1]{%
    \includegraphics[width=0.115\columnwidth]{#1}%
}

\newcommand{\maniimage}[1]{%
    \begingroup
    \setlength{\fboxsep}{0pt}%
    \setlength{\fboxrule}{1.0pt}%
    \fcolorbox{maniBorder}{white}{%
        \includegraphics[width=0.115\columnwidth]{#1}%
    }%
    \endgroup
}

\newcommand{\segcell}[2]{%
    \begin{tikzpicture}[baseline=(img.base)]
        \node[
            inner sep=0pt,
            outer sep=0pt
        ] (img) {#1};

        \node[
            overlay,
            font=\tiny,
            anchor=south,
            yshift=-2.6pt
        ] at (img.north) {#2};
    \end{tikzpicture}%
}


\begin{tabular}{
    c@{\hspace{5pt}}
    c@{\hspace{5pt}}
    c@{\hspace{2pt}}
    c@{\hspace{2pt}}
    c@{\hspace{2pt}}
    c@{\hspace{2pt}}
    c@{\hspace{2pt}}
    c
}




&
&
{\scriptsize $0.02$}
&
{\scriptsize $0.05$}
&
{\scriptsize $0.10$}
&
{\scriptsize $0.30$}
&
{\scriptsize $0.50$}
&
{\scriptsize $1.00$}
\\[-1pt]


\segcell{
    \segimage{
        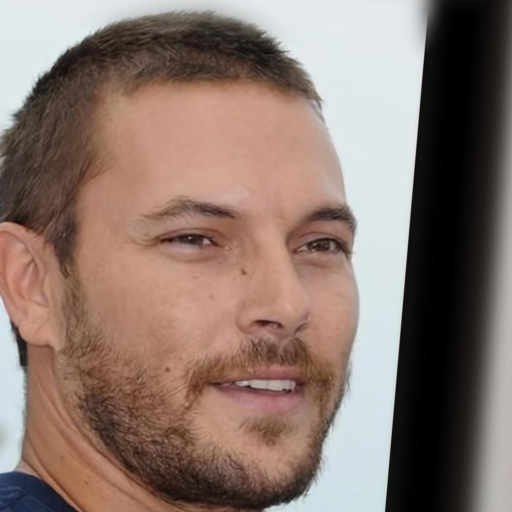
    }
}{Sample $x$}
&
\raisebox{0.7\height}{%
    \rotatebox[origin=c]{90}{\scriptsize Isotropic}%
}
&
\segimage{
    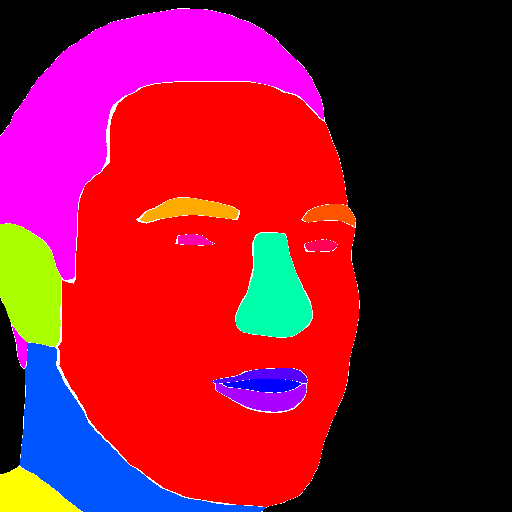
}
&
\segimage{
    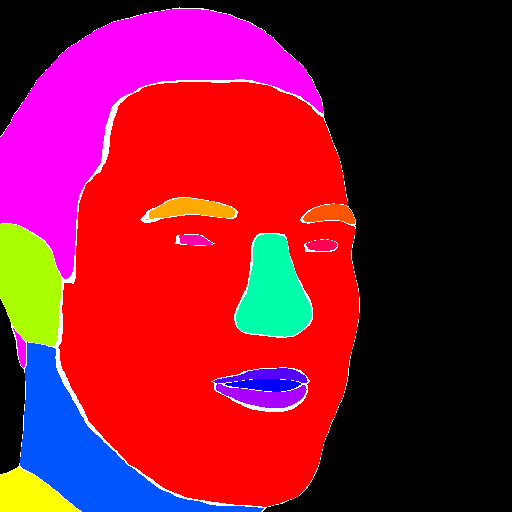
}
&
\begin{tikzpicture}
    \node[
        inner sep=0pt,
        outer sep=0pt
    ] (iso010) {
        \includegraphics[
            width=0.115\columnwidth
        ]{
            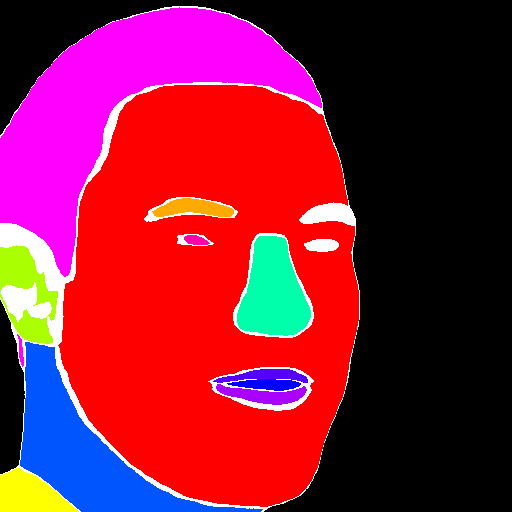
        }
    };

    \node[
        font=\tiny,
        text=white,
        anchor=west,
        fill=black,
        fill opacity=0.55,
        text opacity=1,
        inner sep=1pt
    ] (browlabel)
    at ([xshift=-9mm,yshift=-2mm]iso010.north east)
    {R. Eye};

    \draw[
        ->,
        line width=0.40pt,
        white
    ]
    (browlabel.east)
    --
    ([xshift=1.8mm,yshift=1mm]iso010.center);

\end{tikzpicture}
&
\segimage{
    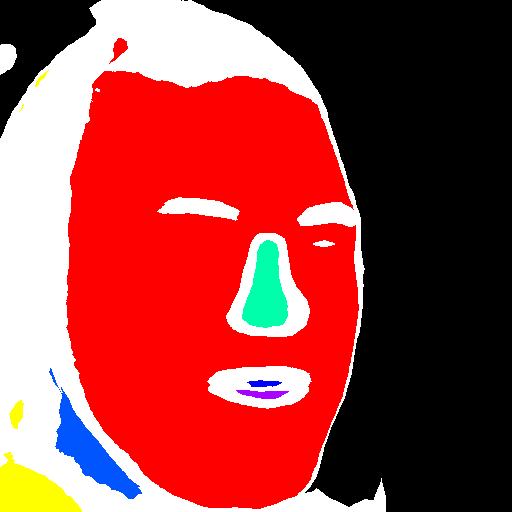
}
&
\segimage{
    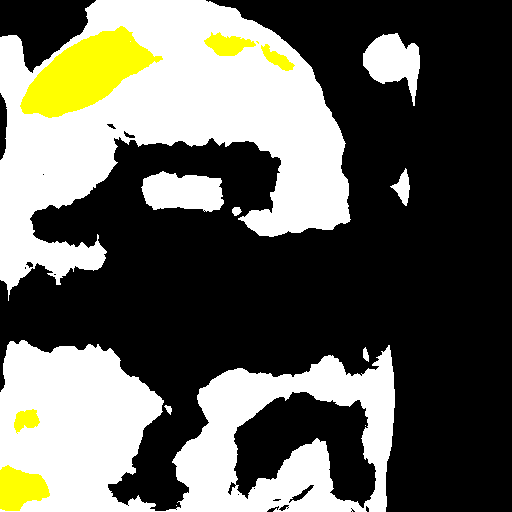
}
&
\segimage{
    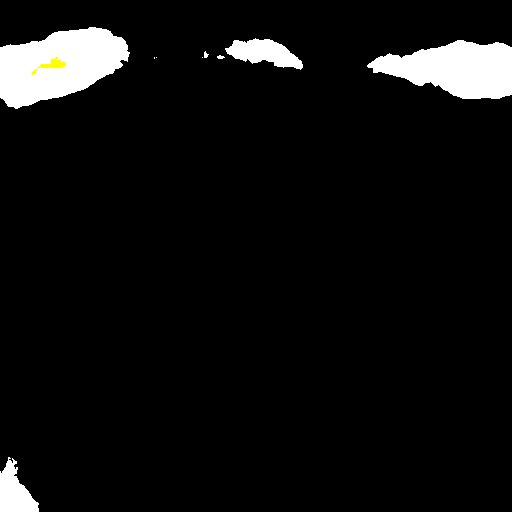
}
\\[2pt]


\segcell{
    \segimage{
        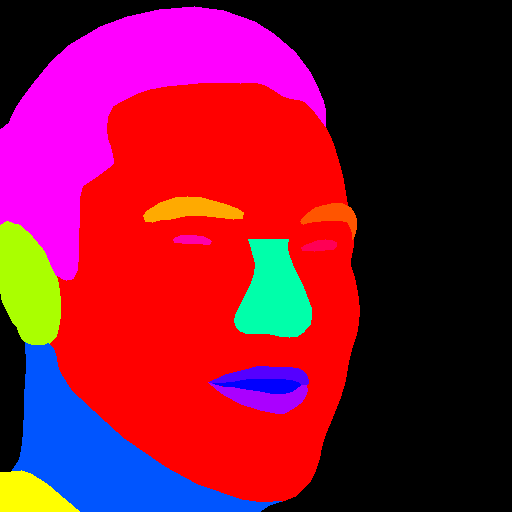
    }
}{GT-Mask}
&
\raisebox{0.92\height}{%
    \rotatebox[origin=c]{90}{\scriptsize \textbf{Ours}}%
}
&
\maniimage{
    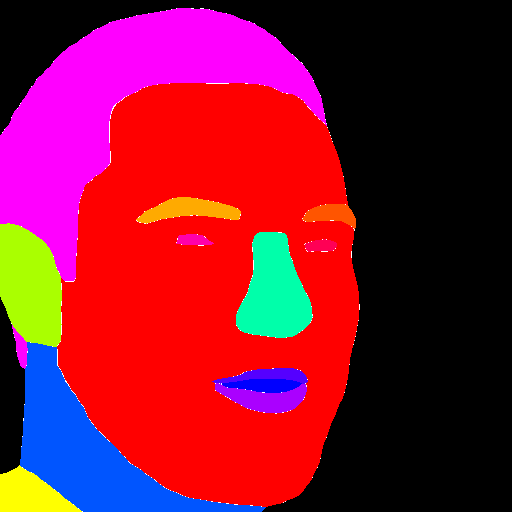
}
&
\maniimage{
    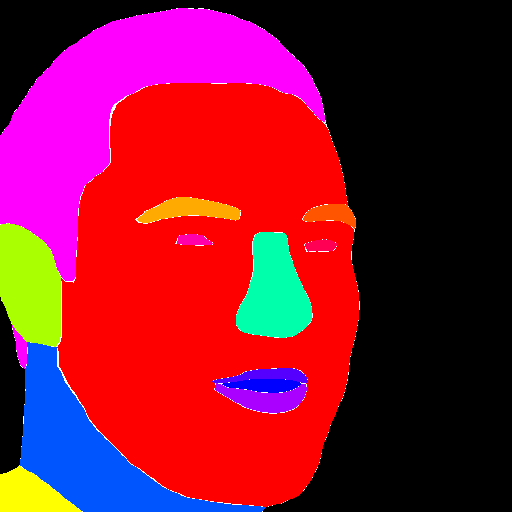
}
&
\maniimage{
    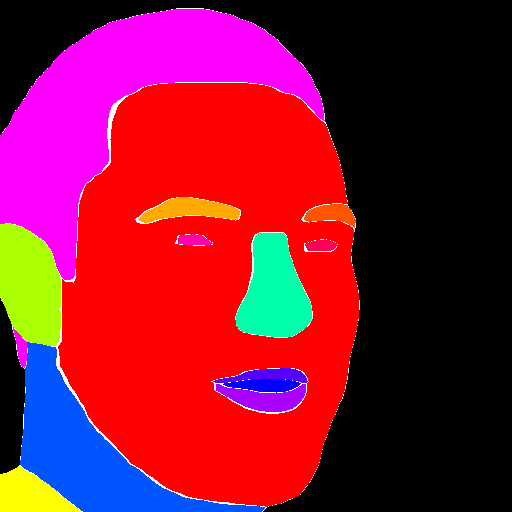
}
&
\maniimage{
    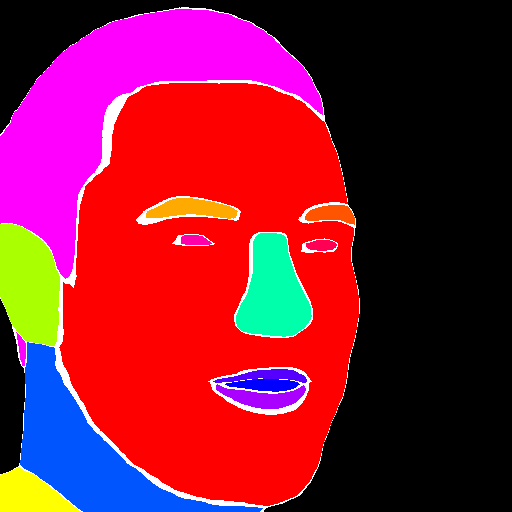
}
&
\maniimage{
    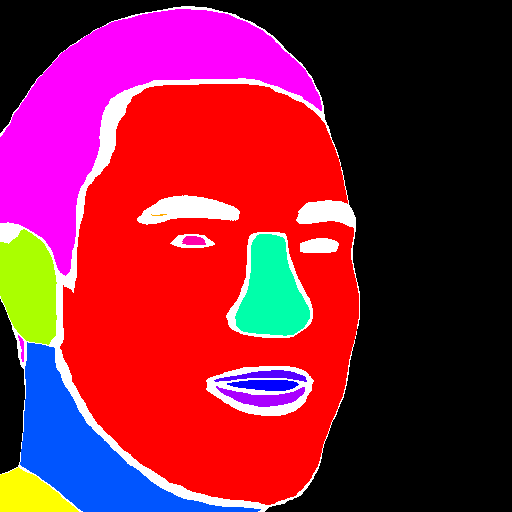
}
&
\maniimage{
    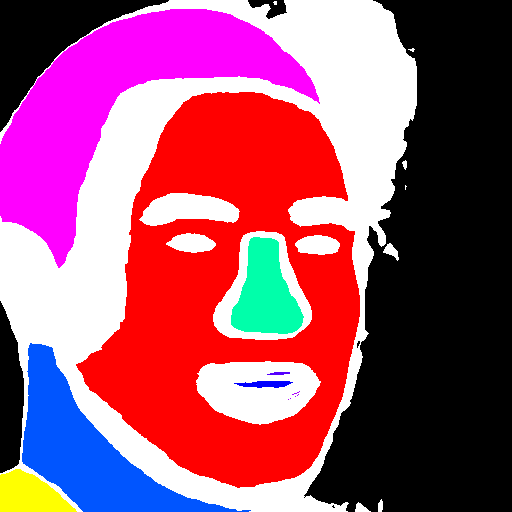
}

\end{tabular}

\endgroup
    }

    \vspace{-3pt}
    \captionof{figure}{
        \textbf{Certified segmentation under increasing
        perturbation levels.}
        CelebAMask-HQ sample under isotropic and
        manifold smoothing; white pixels denote abstention.
    }
    \label{fig:segmentation_comparison}
\end{minipage}
\hfill
\begin{minipage}[t]{0.369\textwidth}    \centering
    \vspace{0.8pt}

    \captionsetup{
        type=table,
        font=scriptsize,
    }

    \captionof{table}{
        \textbf{Dense certification on CelebAMask-HQ.}
         Certified mIoU (\%) and pixel abstention rate ($\varnothing$) under
    isotropic (I) and manifold (M) smoothing.
    }
    \label{tab:seg_miou}

    \vspace{-2pt}

    \resizebox{\linewidth}{!}{%
        \begin{tabular}{@{}c c c c c@{}}
        \toprule
        $\sigma$
        & $\mathrm{mIoU}_{\mathrm I}\uparrow$
        & $\mathrm{mIoU}_{\mathrm M}\uparrow$
        & $\varnothing_{\mathrm I}\downarrow$
        & $\varnothing_{\mathrm M}\downarrow$
        \\
        \midrule

        0.02 & 74.03 & \textbf{74.60} & 1.88 & \textbf{0.48} \\
        0.05 & 73.10 & \textbf{74.54} & 3.60 & \textbf{1.19} \\
        0.10 & 60.76 & \textbf{73.37} & 3.60 & \textbf{2.45} \\
        0.30 & 29.87 & \textbf{67.62} & 23.61 & \textbf{7.52} \\
        0.50 & 5.39  & \textbf{57.44} & 40.07 & \textbf{13.81} \\
        1.00 & 2.12  & \textbf{35.58} & \textbf{8.81} & 30.60 \\
        \bottomrule
        \end{tabular}%
    }

\end{minipage}

\vspace{-3pt}
\end{figure}

\textbf{Dense Certification and Semantic Preservation.}
\textit{Does the advantage of manifold smoothing extend from image-level
classification to dense prediction?}
Tab.~\ref{tab:main_certification} shows that it does.
At $\sigma=0.5$, manifold smoothing provides
$84.27\%$ certified pixel accuracy versus $26.87\%$ isotropically
(Tab.~\ref{tab:main_certification}), and certified mIoU remains $57.44\%$
versus $5.39\%$, with substantially lower pixel abstention
($13.81\%$ vs.\ $40.07\%$; Tab.~\ref{tab:seg_miou}).
At $\sigma=1.0$, isotropic abstention decreases despite near-zero mIoU,
revealing prediction collapse rather than meaningful robustness. Qualitatively, Fig.~\ref{fig:segmentation_comparison} reveals
\textit{which \textbf{semantic} structures remain preserved}: isotropic smoothing
first loses \textit{fine features} (eyes, brows, lips) and, at higher noise,
\textit{broader facial regions}, while manifold smoothing preserves both. Class-wise $\Delta$IoU gains across the full perturbation range
are provided in App. Sec.~\ref{app:segmentation_iou_class_wise}

\begin{takeawaybox}
Manifold smoothing provides \textbf{dense robustness guarantees} over
\textbf{semantic variation} at \textit{larger perturbation scales}, from
\textit{fine to broader (facial) features}.
\end{takeawaybox}

\input{tikz_fig/fig_concepts}

\subsection{Interpretable Latent Spaces Smoothing}
\label{sec:concepts}

Unlike the preceding pixel-space experiments, here we smooth the sparse SAE
concept representation of the input image $x$ while retaining the ImageNet classification task, to study:
\textit{Can robustness guarantees be maintained under perturbations
in \textbf{interpretable sparse concept spaces}?}
Tab.~\ref{tab:main_certification} shows that sparse concept representations are much more sensitive to isotropic noise, limiting certification at larger perturbations. Isotropic
CAcc reaches $0\%$ by $\sigma=0.5$, whereas manifold smoothing respects local
embedding geometry, retaining $63.8\%$ CAcc at $\sigma=0.5$ and $47.2\%$ at
$\sigma=1.0$. This extends robustness guarantees over larger
manifold-aligned variations in concept space.

\begin{figure}[H]
\centering


\newcommand{\retrievaldir}{
figures/experiment/cfm/NN_SAE_retrieval/sigma_1.0/idx36958_1.0_noise_cfm}


\newcommand{\queryimg}[1]{%
    \includegraphics[
        width=.82cm,
        height=.82cm
    ]{#1}%
}

\newcommand{\nnimg}[1]{%
    \includegraphics[
        width=.74cm,
        height=.74cm
    ]{#1}%
}


\resizebox{\linewidth}{!}{%
\begin{tikzpicture}

\node[
    inner sep=0pt
] (query) at (0,1)
{
    \queryimg{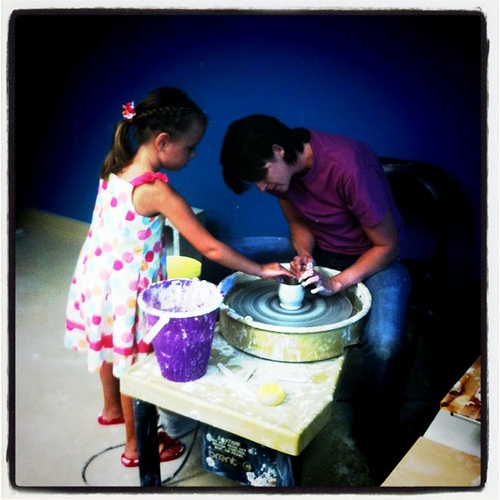}
};

\node[
    above=5pt of query,
    font=\tiny
]
(querytitle)
{Query};

\node[
    anchor=south,
    font=\fontsize{3.8}{4.2}\selectfont,
    inner sep=0pt
]
at ($(query.north)+(0,1.5pt)$)
{potter's wheel};


\node[inner sep=0pt] (i1) at (1.18,1)
{\nnimg{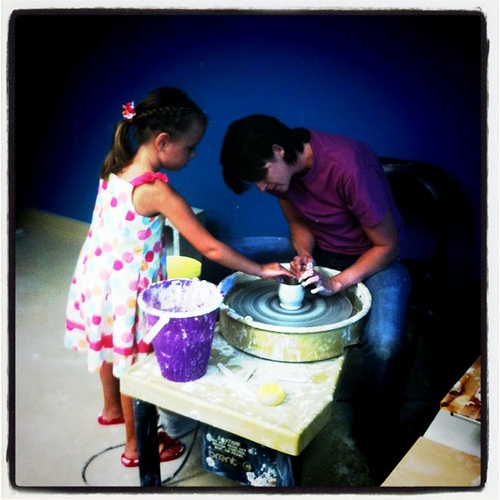}};

\node[inner sep=0pt] (i2) at (1.95,1)
{\nnimg{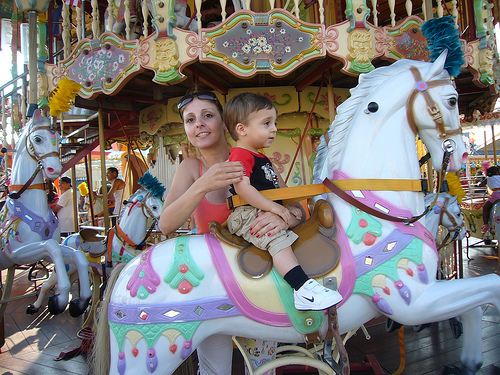}};

\node[inner sep=0pt] (i3) at (2.72,1)
{\nnimg{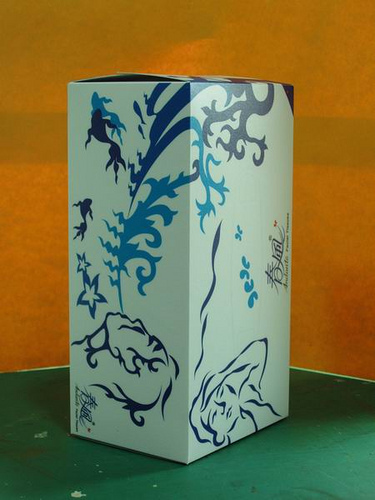}};

\node[inner sep=0pt] (i4) at (3.49,1)
{\nnimg{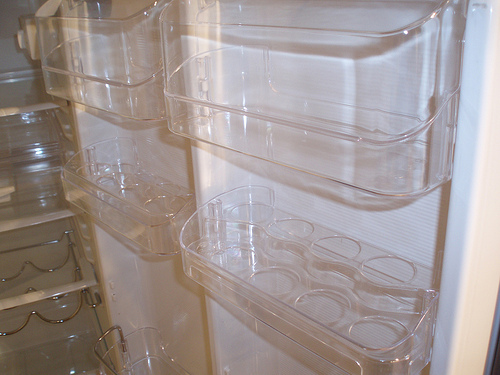}};

\node[inner sep=0pt] (i5) at (4.26,1)
{\nnimg{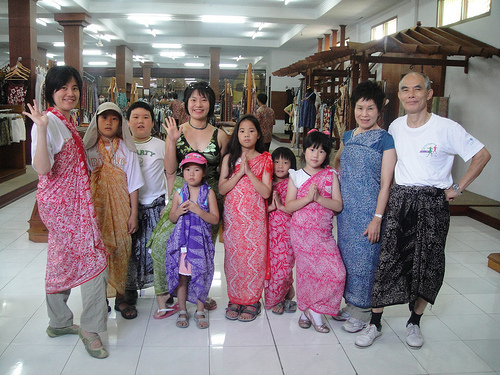}};


\draw[
    decorate,
    decoration={
        brace,
        amplitude=2pt
    },
    draw=isoBlue,
    line width=.42pt
]
($(i1.north west)+(0,.07)$)
--
($(i5.north east)+(0,.07)$);


\node[
    anchor=south,
    align=center,
    font=\tiny,
    text=black
]
at ($(i3.north)+(0,.34)$)
{
    Isotropic \quad ($\ell_2=90.27$)
};


\node[
    anchor=south,
    align=center,
    text width=4cm,
    font=\fontsize{4}{3.5}\selectfont,
    text=black,
    inner sep=0pt
]
at ($(i3.north)+(0,.17)$)
{
    potter's wheel, carousel, carton, refrigerator, sarong
};


\node[inner sep=0pt] (m1) at (5.35,1)
{\nnimg{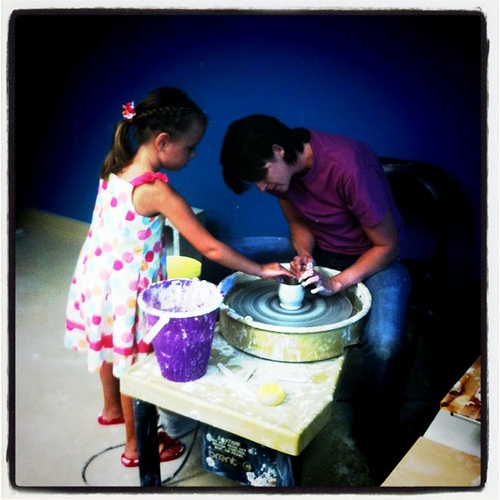}};

\node[inner sep=0pt] (m2) at (6.12,1)
{\nnimg{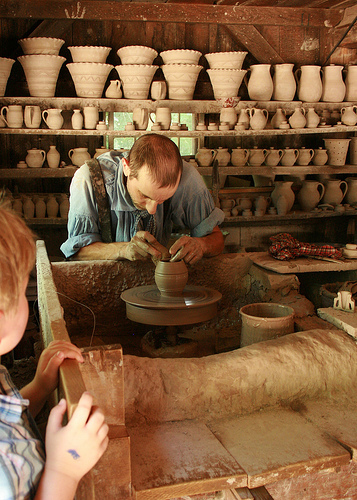}};

\node[inner sep=0pt] (m3) at (6.89,1)
{\nnimg{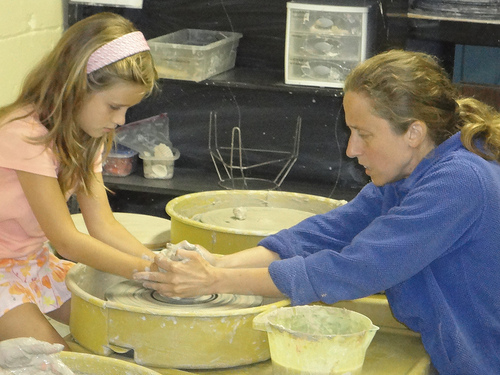}};

\node[inner sep=0pt] (m4) at (7.66,1)
{\nnimg{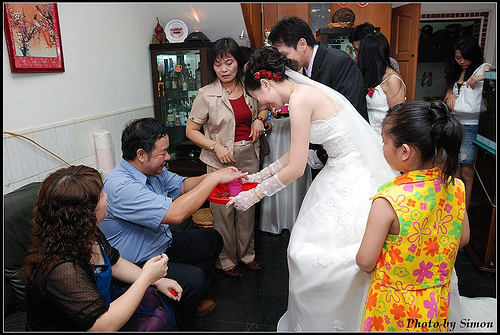}};

\node[inner sep=0pt] (m5) at (8.43,1)
{\nnimg{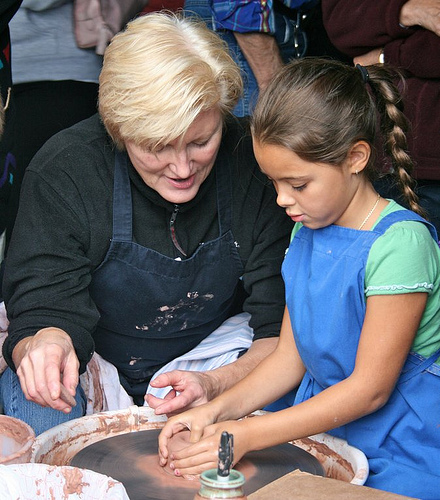}};


\foreach \im in {m1,m2,m3,m4,m5}{
    \draw[
        draw=origGreen,
        line width=.60pt,
        rounded corners=.4pt
    ]
    (\im.south west) rectangle (\im.north east);
}


\draw[
    decorate,
    decoration={
        brace,
        amplitude=2pt
    },
    draw=maniPurple,
    line width=.42pt
]
($(m1.north west)+(0,.07)$)
--
($(m5.north east)+(0,.07)$);


\node[
    anchor=south,
    align=center,
    font=\tiny,
    text=black
]
at ($(m3.north)+(0,.34)$)
{
    \textbf{Manifold (Ours)}\quad ($\ell_2=5.15$)
};


\node[
    anchor=south,
    align=center,
    text width=4cm,
    font=\fontsize{4}{3.5}\selectfont,
    text=black,
    inner sep=0pt
]
at ($(m3.north)+(0,.17)$)
{
    potter's wheel, potter's wheel, potter's wheel, groom, potter's wheel
};

\end{tikzpicture}%
}


\caption{
\textbf{Nearest-neighbor retrieval after smoothing in SAE space.}
At $\sigma=1.0$, manifold smoothing stays closer to the original representation
($\ell_2=5.15$ vs.\ $90.27$) and retrieves more semantically matching top classes (potter's wheel retrieved as 4/5) than isotropic smoothing.
}
\label{fig:sae_nn_retrieval}

\end{figure}

\textbf{Semantic concept consistency.}
\textit{Does prediction-level robustness reflect stability of the underlying
semantic representation?}
Fig.~\ref{fig:sae_concept_comparison} shows that, despite identical top-1
predictions, isotropic perturbations activate weak or absent concepts, while
manifold perturbations remain closer to the original activation profile.
Consistently, $z_{\mathrm M}$ remains much closer to the original embedding
than $z_{\mathrm I}$ ($\ell_2$: $5.15$ vs.\ $90.27$) and better preserves its
nearest-neighbor structure (Fig.~\ref{fig:sae_nn_retrieval}) indicating semantic consistency under
manifold smoothing. Results for initially inactive concepts are provided in App.~\ref{app:top_least_concepts}.

\takeaway{
Manifold-aligned \textbf{guarantees} extend \textbf{semantic coherence}
in interpretable sparse spaces.
}

\begin{wraptable}{r}{0.4\textwidth}
\vspace{-5pt}
\centering
\captionsetup{
    font=scriptsize,
    labelfont=bf,
    skip=3pt
}

\caption{
\textbf{RoCOCO Danger retrieval under smoothing.}
Higher R@1 and lower RSMS and Drop are preferred.
}
\label{tab:rococo_danger_main}

\tiny
\setlength{\tabcolsep}{2.5pt}
\renewcommand{\arraystretch}{0.95}

\begin{tabular}{@{}c ccc ccc@{}}
\toprule
&
\multicolumn{3}{c}{Isotropic}
&
\multicolumn{3}{c}{\textbf{Manifold} (Ours)}
\\
\cmidrule(lr){2-4}
\cmidrule(lr){5-7}

$\sigma$
& R@1 $\uparrow$
& RSMS $\downarrow$
& Drop $\downarrow$
& R@1 $\uparrow$
& RSMS $\downarrow$
& Drop $\downarrow$
\\
\midrule

0.02
& 35.23 & 26.46 & 29.70
& \textbf{41.26} & \textbf{22.98} & \textbf{17.70}
\\

0.05
& 8.54 & 32.20 & 66.00
& \textbf{41.34} & \textbf{22.90} & \textbf{17.60}
\\

0.10
& 3.08 & 32.96 & 93.90
& \textbf{41.52} & \textbf{22.80} & \textbf{17.20}
\\

0.50
& 0.02 & 27.48 & 100.00
& \textbf{35.14} & \textbf{25.26} & \textbf{29.90}
\\

1.00
& 0.02 & 25.00 & 100.00
& \textbf{21.32} & \textbf{28.14} & \textbf{57.50}
\\

\bottomrule
\end{tabular}

\end{wraptable}

\subsection{Safety Alignment in Foundation Models}
\label{sec:rococo_main}

Vision-language models rely on semantic alignment in a shared
embedding space. We ask \textit{whether manifold smoothing can
preserve this \textbf{alignment} under semantic perturbations}.
We evaluate CLIP ViT-B/32 on RoCOCO~\citep{Park2024RoCOCO}, which augments
COCO with semantically altered captions. We report the \textit{Danger}
split, where security-related concepts replace low-influence nouns;
all corruption bank sweeps are provided in
App. Sec.~\ref{app:rococo_extended}. At $\sigma = 0.1$, manifold smoothing limits the retrieval drop to
$17.20\%$, compared with $93.90\%$ under isotropic smoothing,
indicating substantially better preservation of retrieval-relevant
semantic information in embedding space. Notably, smoothing can itself increase RSMS relative to unsmoothed CLIP,
but this increase is consistently smaller under manifold smoothing. We provide a qualitative example to visualize semantic alignment in
the App. Fig.~\ref{app:rococo_caption_qual_eg}.



\takeaway{
Local covariance smoothing retains \textbf{cross-modal retrieval robustness.}
}





\section{Discussion and Conclusion}
\label{discussion}

\textbf{Discussion.} Manifold smoothing outperforms isotropic smoothing across four tasks and three representation spaces (pixels, sparse concepts, multimodal embeddings), including under semantic distribution shifts, with lower abstention rates, while preserving concept activations in sparse spaces and maintaining cross-modal retrieval alignment. This supports our hypothesis that isotropic certificates spend their budget on off-manifold directions rather than along meaningful variations in the data.

We would like to note that using the same noise scale $\sigma$ does not imply equal perturbation magnitudes. We conducted matched-norm baselines in App. \ref{app:basline} that suggest that restricting perturbations to the locally supported PCA subspace is an important factor, while manifold smoothing further benefits from scaling noise according to the observed variance in the data. Hence, when robustness to meaningful data variations is the goal, restricting perturbations to locally supported directions yields stronger guarantees than standard randomized smoothing. 

The effectiveness of the approach also depends on how reliably finite data neighbours can estimate the local subspace. In domains where the intrinsic dimensionality is substantially smaller than the ambient dimensionality, such as typical vision or language representations, a moderate number of local samples may suffice to estimate the feasible covariance, provided the neighborhood is sufficiently dense and locally concentrated. Conversely, higher intrinsic dimensionality requires more samples for reliable covariance estimation. 
While we treat the retained PCA dimension $m$ as a hyperparameter, practitioners may use the effective rank as a diagnostic proxy for local effective dimensionality and can consider larger neighborhoods for higher-dimensional manifolds, where sufficient samples are needed for reliable covariance estimation preserving locality.

\textbf{Limitations.} Manifold smoothing adds per-input computation cost from neighborhood retrieval and
local eigen-decomposition, both computed once and reused across all Monte
Carlo samples, reducing cost. Approximate neighbor search reduces the retrieval cost.
In App. Sec. ~\ref{app:runtime} we discuss a single-input runtime cost. The method further assumes a canonical input space in which ambient $\ell_2$ distance is locally meaningful. This assumption is more natural for some learned representation spaces and normalized data domains such as CelebA than for raw natural images. For particular inputs, such as natural images, further research has to be done to ensure that local perturbations are equivariant to transformations such as translations of objects in the fixed pixel grid.
Similar to standard Randomized Smoothing, certification also becomes harder as the number of classes grows, since exceeding $\tau$ is less attainable when probability mass spreads over many classes.




\textbf{Summary.}
We introduced Randomized Manifold Smoothing, a framework that provides formal robustness guarantees against perturbations along meaningful data variations. By shaping the smoothing noise with the local covariance of an input's nearest neighbors, our method certifies an ellipsoid aligned with the data manifold rather than an isotropic ball stretching into the ambient space without any relation to the data. 
Empirically, we showed that manifold smoothing preserves task performance and certified accuracy at perturbation scales where isotropic smoothing collapses. Hence, we show that certification is not limited to adversarial threat models, but practitioners can obtain guarantees over meaningful input variations, bridging formal robustness and real-world reliability.

\clearpage
\section{Use of LLMs}


The research methodology, experimental design, analysis, and scientific
conclusions presented in this work were developed by the authors. We used
large language models to generate text snippets for selected sections,
which were subsequently reviewed and developed into the final text by the
authors. LLMs were also used for language editing to improve clarity and
conciseness and for \LaTeX{} formatting of figures and tables. During
implementation, LLMs assisted with project configuration and writing modular code pipelines. All generated code was reviewed and adapted before use.

\section{Reproducibility Statement}
All datasets used in this work are publicly available. We provide references
to the pretrained models used for the classification and CFM experiments.
Detailed experimental settings, including model, smoothing, sampling, and
certification parameters, are provided in the appendix, together with
pseudocode for our method. We will release the complete implementation code upon acceptance to facilitate the reproduction of our
results.

\bibliography{bib}

\begin{thebibliography}{27}
\providecommand{\natexlab}[1]{#1}
\providecommand{\url}[1]{\texttt{#1}}
\expandafter\ifx\csname urlstyle\endcsname\relax
  \providecommand{\doi}[1]{doi: #1}\else
  \providecommand{\doi}{doi: \begingroup \urlstyle{rm}\Url}\fi

\bibitem[Alfarra et~al.(2022{\natexlab{a}})Alfarra, Bibi, Khan, Torr, and Ghanem]{alfarra2022deformrs}
Motasem Alfarra, Adel Bibi, Naeemullah Khan, Philip H.~S. Torr, and Bernard Ghanem.
\newblock {DeformRS}: Certifying input deformations with randomized smoothing.
\newblock In \emph{AAAI Conference on Artificial Intelligence (AAAI)}, 2022{\natexlab{a}}.

\bibitem[Alfarra et~al.(2022{\natexlab{b}})Alfarra, Bibi, Torr, and Ghanem]{alfarra2022data}
Motasem Alfarra, Adel Bibi, Philip H.~S. Torr, and Bernard Ghanem.
\newblock Data dependent randomized smoothing.
\newblock In \emph{Conference on Uncertainty in Artificial Intelligence (UAI)}, 2022{\natexlab{b}}.

\bibitem[Anani et~al.(2024)Anani, Lorenz, Schiele, and Fritz]{anani2024adaptive}
Alaa Anani, Tobias Lorenz, Bernt Schiele, and Mario Fritz.
\newblock Adaptive hierarchical certification for segmentation using randomized smoothing.
\newblock In \emph{International Conference on Machine Learning (ICML)}, 2024.

\bibitem[Anani et~al.(2025)Anani, Lorenz, Fritz, and Schiele]{anani2025pixel}
Alaa Anani, Tobias Lorenz, Mario Fritz, and Bernt Schiele.
\newblock Pixel-level certified explanations via randomized smoothing.
\newblock In \emph{International Conference on Machine Learning (ICML)}, 2025.

\bibitem[Anani et~al.(2026)Anani, Lorenz, Schiele, Fritz, and Fischer]{anani2026certified}
Alaa Anani, Tobias Lorenz, Bernt Schiele, Mario Fritz, and Jonas Fischer.
\newblock Certified circuits: Stability guarantees for mechanistic circuits.
\newblock In \emph{International Conference on Machine Learning (ICML)}, 2026.

\bibitem[Bernhardsson(2018)]{bernhardsson2018annoy}
Erik Bernhardsson.
\newblock Annoy: Approximate nearest neighbors in {C++}/{Python}.
\newblock \url{https://github.com/spotify/annoy}, 2018.
\newblock GitHub repository.

\bibitem[Chiang et~al.(2020)Chiang, Curry, Abdelkader, Kumar, Dickerson, and Goldstein]{chiang2020detection}
Ping-yeh Chiang, Michael~J. Curry, Ahmed Abdelkader, Aounon Kumar, John Dickerson, and Tom Goldstein.
\newblock Detection as regression: Certified object detection with median smoothing.
\newblock In \emph{Advances in Neural Information Processing Systems (NeurIPS)}, 2020.

\bibitem[Cohen et~al.(2019)Cohen, Rosenfeld, and Kolter]{cohen2019certified}
Jeremy Cohen, Elan Rosenfeld, and Zico Kolter.
\newblock Certified adversarial robustness via randomized smoothing.
\newblock In \emph{International Conference on Machine Learning (ICML)}, 2019.

\bibitem[Deng et~al.(2009)Deng, Dong, Socher, Li, Li, and Fei-Fei]{deng2009imagenet}
Jia Deng, Wei Dong, Richard Socher, Li-Jia Li, Kai Li, and Li~Fei-Fei.
\newblock {ImageNet}: A large-scale hierarchical image database.
\newblock In \emph{IEEE Conference on Computer Vision and Pattern Recognition (CVPR)}, pp.\  248--255, 2009.

\bibitem[Eiras et~al.(2022)Eiras, Alfarra, Torr, Kumar, Dokania, Ghanem, and Bibi]{eiras2022ancer}
Francisco Eiras, Motasem Alfarra, Philip Torr, M.~Pawan Kumar, Puneet~K. Dokania, Bernard Ghanem, and Adel Bibi.
\newblock {ANCER}: Anisotropic certification via sample-wise volume maximization.
\newblock \emph{Transactions on Machine Learning Research (TMLR)}, 2022.

\bibitem[Fefferman et~al.(2016)Fefferman, Mitter, and Narayanan]{fefferman2016testing}
Charles Fefferman, Sanjoy Mitter, and Hariharan Narayanan.
\newblock Testing the manifold hypothesis.
\newblock \emph{Journal of the American Mathematical Society}, 2016.

\bibitem[Fischer et~al.(2020)Fischer, Baader, and Vechev]{fischer2020certified}
Marc Fischer, Maximilian Baader, and Martin Vechev.
\newblock Certified defense to image transformations via randomized smoothing.
\newblock In \emph{Advances in Neural Information Processing Systems (NeurIPS)}, 2020.

\bibitem[Fischer et~al.(2021)Fischer, Baader, and Vechev]{Fischer2021ScalableSegmentation}
Marc Fischer, Maximilian Baader, and Martin Vechev.
\newblock Scalable certified segmentation via randomized smoothing.
\newblock In \emph{Proceedings of the 38th International Conference on Machine Learning (ICML)}, 2021.

\bibitem[Karras et~al.(2018)Karras, Aila, Laine, and Lehtinen]{karras2018progressive}
Tero Karras, Timo Aila, Samuli Laine, and Jaakko Lehtinen.
\newblock Progressive growing of {GANs} for improved quality, stability, and variation.
\newblock In \emph{International Conference on Learning Representations (ICLR)}, 2018.

\bibitem[Kumar \& Goldstein(2021)Kumar and Goldstein]{kumar2021center}
Aounon Kumar and Tom Goldstein.
\newblock Center smoothing: Certified robustness for networks with structured outputs.
\newblock In \emph{Advances in Neural Information Processing Systems (NeurIPS)}, 2021.

\bibitem[Lecuyer et~al.(2019)Lecuyer, Atlidakis, Geambasu, Hsu, and Jana]{lecuyer2019certified}
Mathias Lecuyer, Vaggelis Atlidakis, Roxana Geambasu, Daniel Hsu, and Suman Jana.
\newblock Certified robustness to adversarial examples with differential privacy.
\newblock In \emph{IEEE Symposium on Security and Privacy (S\&P)}, 2019.

\bibitem[Lee et~al.(2020)Lee, Liu, Wu, and Luo]{CelebAMask-HQ}
Cheng-Han Lee, Ziwei Liu, Lingyun Wu, and Ping Luo.
\newblock {MaskGAN}: Towards diverse and interactive facial image manipulation.
\newblock In \emph{IEEE Conference on Computer Vision and Pattern Recognition (CVPR)}, 2020.

\bibitem[Lee et~al.(2019)Lee, Yuan, Chang, and Jaakkola]{lee2019tight}
Guang-He Lee, Yang Yuan, Shiyu Chang, and Tommi Jaakkola.
\newblock Tight certificates of adversarial robustness for randomly smoothed classifiers.
\newblock In \emph{Advances in Neural Information Processing Systems (NeurIPS)}, 2019.

\bibitem[Levine \& Feizi(2020)Levine and Feizi]{levine2020randomized}
Alexander Levine and Soheil Feizi.
\newblock Robustness certificates for sparse adversarial attacks by randomized ablation.
\newblock In \emph{AAAI Conference on Artificial Intelligence (AAAI)}, pp.\  4585--4593, 2020.

\bibitem[Li et~al.(2021)Li, Weber, Xu, Rimanic, Kailkhura, Xie, Zhang, and Li]{li2021tss}
Linyi Li, Maurice Weber, Xiaojun Xu, Luka Rimanic, Bhavya Kailkhura, Tao Xie, Ce~Zhang, and Bo~Li.
\newblock {TSS}: Transformation-specific smoothing for robustness certification.
\newblock In \emph{ACM SIGSAC Conference on Computer and Communications Security (CCS)}, pp.\  535--557, 2021.

\bibitem[Liu et~al.(2015)Liu, Luo, Wang, and Tang]{liu2015faceattributes}
Ziwei Liu, Ping Luo, Xiaogang Wang, and Xiaoou Tang.
\newblock Deep learning face attributes in the wild.
\newblock In \emph{Proceedings of International Conference on Computer Vision (ICCV)}, December 2015.

\bibitem[Park et~al.(2024)Park, Um, Yoon, Chun, Yun, and Choi]{Park2024RoCOCO}
Seulki Park, Daeho Um, Hajung Yoon, Sanghyuk Chun, Sangdoo Yun, and Jin~Young Choi.
\newblock {RoCOCO}: Robustness benchmark of ms-coco to stress-test image-text matching models.
\newblock \emph{arXiv preprint arXiv:2304.10727}, 2024.

\bibitem[Pope et~al.(2021)Pope, Zhu, Abdelkader, Goldblum, and Goldstein]{pope2021intrinsic}
Phillip Pope, Chen Zhu, Ahmed Abdelkader, Micah Goldblum, and Tom Goldstein.
\newblock The intrinsic dimension of images and its impact on learning.
\newblock In \emph{International Conference on Learning Representations (ICLR)}, 2021.

\bibitem[S{\'u}ken{\'i}k et~al.(2022)S{\'u}ken{\'i}k, Kuvshinov, and G{\"u}nnemann]{sukenik2022intriguing}
Peter S{\'u}ken{\'i}k, Aleksei Kuvshinov, and Stephan G{\"u}nnemann.
\newblock Intriguing properties of input-dependent randomized smoothing.
\newblock In \emph{International Conference on Machine Learning (ICML)}, pp.\  20697--20743, 2022.

\bibitem[Wittenmayer et~al.(2026)Wittenmayer, Rao, Parchami-Araghi, Schiele, and Fischer]{Wittenmayer2026CFM}
Kai Wittenmayer, Sukrut Rao, Amin Parchami-Araghi, Bernt Schiele, and Jonas Fischer.
\newblock {CFM}: Language-aligned concept foundation model for vision.
\newblock \emph{arXiv preprint arXiv:2601.13798}, 2026.

\bibitem[Yang et~al.(2020)Yang, Duan, Hu, Salman, Razenshteyn, and Li]{yang2020randomized}
Greg Yang, Tony Duan, J.~Edward Hu, Hadi Salman, Ilya Razenshteyn, and Jerry Li.
\newblock Randomized smoothing of all shapes and sizes.
\newblock In \emph{International Conference on Machine Learning (ICML)}, pp.\  10693--10705, 2020.

\bibitem[Yu et~al.(2018)Yu, Wang, Peng, Gao, Yu, and Sang]{yu2018bisenet}
Changqian Yu, Jingbo Wang, Chao Peng, Changxin Gao, Gang Yu, and Nong Sang.
\newblock {BiSeNet}: Bilateral segmentation network for real-time semantic segmentation.
\newblock In \emph{ECCV}, 2018.

\end{thebibliography}
\bibliographystyle{iclr2027_conference}

\clearpage

\appendix
\appendix

\section*{Overview of the Appendix}

In this appendix, we provide additional theoretical analysis, experimental
details, ablations, and extended results. Theoretical analysis and proofs
are provided in Sec.~\ref{app:theory},
experimental details in Sec.~\ref{app:experiment_details},
evaluation metrics in Sec.~\ref{app:evaluation}, and ablation studies in
Sec.~\ref{app:ablation_study}. Extended quantitative and qualitative
results are provided in Secs.~\ref{app:quantitative}--\ref{app:qualitative}.

\section{Discussion on Theory}
\label{app:theory}
In this section we discuss formally what the implications of the local covariance estimator are on how well we model the underlying data geometry. We first give a more abstract outline of the theorems in the next section and then provide the detailed proofs in the following sections.

\subsection{Outline for the Convergence of the Empirical Manifold Ellipsoid}
\label{sec:ellipsoid-convergence}

Our certificate relies on the empirical local covariance matrix $\Sigma_x^{\mathrm{loc}}$ Eq.~\ref{eq:covar} computed from the $k$ nearest neighbors $N_k(x)=\{i_1,\ldots,i_k\}$ of $x$.
The resulting certified region is the ellipsoid
\[
    E_x(r)=\left\{z\in \mathbb R^d:(z-x)^\top U_x\Lambda_x^{-1}U_x^\top(z-x)\le r^2\right\},
\]
where $\Sigma_x^{\mathrm{loc}}=U_x\Lambda_xU_x^\top$ and $r$ the certified radius.
In this section, we state a convergence result showing that this empirical
ellipsoid approaches an ideal local manifold ellipsoid as the number of local
samples increases, provided the neighborhood remains sufficiently local.

To distinguish the empirical quantity from its population analogue, we write $\widehat \Sigma_{x,k}^{\mathrm{loc}}:=\Sigma_x^{\mathrm{loc}}.$
Let $\mathcal M\subset \mathbb R^d$ denote the data manifold and let
$T_x\mathcal M$ be its tangent space at $x$. We define the $k$-NN radius
\[
\rho_k(x):=\max_{j\le k}\|x_{i_j}-x\|_2.
\]
We can define the ideal local tangent covariance by projecting all points in the neighborhood of $x$ on the tangent plane at $x$ as
\[
    \Sigma_{x,\rho_k}^{\star}:=\operatorname{Cov}\left(P_{T_x\mathcal M}(X-x)\,\middle|\,X\in \mathcal M\cap B(x,\rho_k(x))\right),
\]
viewed as a $d\times d$ matrix by extending it with zeros in directions normal
to $T_x\mathcal M$. Here, $P_{T_x\mathcal M}$ is the projection onto the tangent plane of the manifold at $x$ and $B(x,\rho_k(x))$ is the euclidean ball with radius $\rho_k(x)$ centered at $x$.

Now let
\[
    \Sigma_{x,\rho_k}^{\star}=U_x^\star \Lambda_x^\star U_x^{\star\top}
\]
be its eigendecomposition, where the nonzero eigenvalues are
\[
    \lambda_{x,1}^\star\ge \cdots \ge\lambda_{x,m}^\star>0.
\]
The \textit{tangent--normal eigengap} is
\[
    \gamma_x:=\lambda_{x,m}^\star-\lambda_{x,m+1}^\star=\lambda_{x,m}^\star,
\]
since $\Sigma_{x,\rho_k}^{\star}$ has rank $m$.

We first show the convergence of covariance and local PCA.
For that, we assume that the data lie locally on a smooth $m$-dimensional manifold
$\mathcal M\subset \mathbb R^d$ whose reach is at least $\tau$, and that the
$k$ nearest neighbors of $x$ lie in a sufficiently small ball
$B(x,\rho_k(x))$ where the manifold is well approximated by its tangent space.
We also assume that the local samples are well-behaved, e.g. sub-Gaussian, in
tangent coordinates.

Under these assumptions, the empirical local covariance
$\widehat\Sigma_{x,k}^{\mathrm{loc}}$ computed from the $k$ nearest neighbors
converges to the ideal tangent-space covariance
$\Sigma_{x,\rho_k}^{\star}$ at a rate controlled by a finite-sample term and a
curvature-bias term. Consequently, the empirical eigenvalues and top-$m$ PCA
subspace converge to their ideal tangent-space counterparts, with the subspace
error additionally controlled by the tangent--normal eigengap $\gamma_x$.

\begin{theorem}[Covariance and local PCA convergence]
\label{thm:cov-convergence}
Assume that the data lie on an $m$-dimensional $C^2$ manifold
$\mathcal M\subset \mathbb R^d$ with reach at least $\tau>0$.
Assume further that the local samples in $N_k(x)$ are sub-Gaussian in tangent
coordinates and that $\rho_k(x)\le c\tau$ for a sufficiently small universal
constant $c>0$.

Then, with probability at least $1-\delta$,
\[
    \left\|\widehat \Sigma_{x,k}^{\mathrm{loc}}-\Sigma_{x,\rho_k}^{\star}\right\|_2\le\eta_k(x,\delta),
\]
where
\[
    \eta_k(x,\delta)=C_1\rho_k(x)^2\left(\sqrt{\frac{m+\log(1/\delta)}{k}}+\frac{m+\log(1/\delta)}{k}\right)+C_2\frac{\rho_k(x)^3}{\tau}.
\]
\end{theorem}
The first term is the finite-sample covariance estimation error, while the
second term is the local curvature bias incurred by replacing the manifold by
its tangent space.
If $\eta_k(x,\delta)\le \gamma_x/4$, then the empirical PCA subspace
$\widehat U_{x,m}$ spanned by the top $m$ eigenvectors of
$\widehat \Sigma_{x,k}^{\mathrm{loc}}$ satisfies
\[
    \left\|\sin \Theta(\widehat U_{x,m},U_x^\star)\right\|_2\le\frac{2\eta_k(x,\delta)}{\gamma_x}.
\]
Here, \(\Theta(\widehat U_{x,m},U_x^\star)\) denotes the matrix of principal
angles between the empirical PCA subspace \(\operatorname{span}(\widehat U_{x,m})\)
and the ideal tangent subspace \(\operatorname{span}(U_x^\star)=T_x\mathcal M\).
Thus,
\(\|\sin\Theta(\widehat U_{x,m},U_x^\star)\|_2\) is the sine of the largest
principal angle between these two subspaces and quantifies the error in the
estimated local manifold directions.
Moreover, for each $j=1,\ldots,m$,
\[
    |\widehat \lambda_{x,j}-\lambda_{x,j}^\star|\le\eta_k(x,\delta).
\]

In summary, Theorem~\ref{thm:cov-convergence} shows that the empirical
ellipsoid is accurate when three quantities are favorable: the number of
neighbors $k$ is large, the local neighborhood radius $\rho_k(x)$ is small
relative to the manifold reach $\tau$, and the tangent--normal eigengap
$\gamma_x$ is not too small. In particular, the eigengap controls the stability
of the local PCA directions.

The preceding result controls the covariance and tangent subspace. We next
state the corresponding consequence for the Mahalanobis metric used by the
certificate.
For that we assume the conditions of Theorem~\ref{thm:cov-convergence} and additionally require that the
covariance estimation error is small compared to the smallest nonzero ideal
eigenvalue $\lambda_{x,m}^{\star}$. This ensures that the inverse covariance,
and hence the Mahalanobis metric that defines the certificate, is stable.

Let the empirical rank-$m$ Mahalanobis metric be
\[
    \widehat A_x:=\widehat U_{x,m}\widehat \Lambda_{x,m}^{-1}\widehat U_{x,m}^\top,
\]
and the ideal tangent Mahalanobis metric
\[
    A_x^\star:=U_x^\star(\Lambda_x^\star)^{-1}U_x^{\star\top}.
\]

Under assumptions mentioned above, the empirical Mahalanobis metric
$\widehat A_x$ converges to the ideal tangent Mahalanobis metric $A_x^\star$.
As a result, the empirical certified ellipsoid $\widehat E_x(r)$ is sandwiched
between slightly smaller and slightly larger ideal ellipsoids, showing that the
certificate induced by the empirical $k$-NN covariance converges to the ideal
local manifold certificate.

\begin{theorem}[Convergence of the certified ellipsoid]
\label{thm:main-ellipsoid-convergence}
Let the assumptions of
Theorem~\ref{thm:cov-convergence} hold, and assume
$\eta_k(x,\delta)\le\lambda_{x,m}^\star/4.$
Then, with probability at least $1-\delta$,
\[
    \left\|\widehat A_x-A_x^\star\right\|_2
    \le C_3\eta_k(x,\delta)\left(\frac{1}{(\lambda_{x,m}^\star)^2}+\frac{1}{\lambda_{x,m}^\star\gamma_x}\right).
\]
Consequently, for any ambient radius $R>0$, define
\[
    \zeta_R:=R^2 C_3\eta_k(x,\delta)\left(\frac{1}{(\lambda_{x,m}^\star)^2}+\frac{1}{\lambda_{x,m}^\star\gamma_x}\right).
\]
Let estimated and ideal certified ellipsoid be
\[
    \widehat E_x(r):=\left\{z:(z-x)^\top \widehat A_x(z-x)\le r^2\right\}
\]
and
\[
    E_x^\star(r):=\left\{z:(z-x)^\top A_x^\star(z-x)\le r^2\right\},
\]
respectively. Then, if $r^2>\zeta_R$, we have
\[
    E_x^\star\!\left(\sqrt{r^2-\zeta_R}\right)\cap B(x,R)\subseteq\widehat E_x(r)\cap B(x,R)\subseteq E_x^\star\!\left(\sqrt{r^2+\zeta_R}\right)\cap B(x,R).
\]
Thus, the empirical certificate converges to the ideal local manifold
certificate, with error controlled by the number of neighbors $k$, the
tangent--normal eigengap $\gamma_x$, and the curvature scale $\tau$.
\end{theorem}

In the usual local manifold regime,
\[
    \lambda_{x,j}^\star \asymp \rho_k(x)^2,\qquad j=1,\ldots,m,
\]
and hence also $\gamma_x\asymp \rho_k(x)^2$.
Here \(a_k\asymp b_k\) means that \(a_k\) and \(b_k\) are equal up to
constant factors, i.e., there exist constants \(c_1,c_2>0\) such that
\(c_1 b_k\le a_k\le c_2 b_k\). Thus,
\(\lambda_{x,j}^{\star}\asymp \rho_k(x)^2\) means that the local tangent
variance along direction \(j\) scales like the squared radius of the
\(k\)-nearest neighborhood.

If this is satisfied, the ellipsoid error scales heuristically as
\[
    O\left(\sqrt{\frac{m+\log(1/\delta)}{k}}+\frac{\rho_k(x)}{\tau}\right).
\]
For a training set of size $n_{\mathcal{D}}$ sampled from a regular density on an
$m$-dimensional manifold, the $k$-NN radius typically satisfies
$\rho_k(x)\asymp \left(\frac{k}{n_{\mathcal{D}}}\right)^{1/m}.$ Hence the relative error is of the form
\[
    O\left(\sqrt{\frac{m+\log(1/\delta)}{k}}+\frac{1}{\tau}\left(\frac{k}{n_{\mathcal{D}}}\right)^{1/m}\right).
\]
This displays the \textit{bias--variance tradeoff} in choosing $k$: smaller
neighborhoods produce noisier covariance estimates, while larger neighborhoods
reduce sampling error but incur more curvature bias.
We discuss the full proof next.

\subsection{Detailed Proofs for Local Ellipsoid Convergence}
\label{app:ellipsoid-convergence}

This appendix provides the proof of
Theorems~\ref{thm:cov-convergence}
and~\ref{thm:main-ellipsoid-convergence}. The proof separates the
approximation error into three components, which are the finite-sample covariance error, the curvature bias, and the inverse-metric perturbation.
We will first discuss the relevant notation.

\subsubsection{Geometric and probabilistic setup}

Let $\mathcal M\subset \mathbb R^d$ be an $m$-dimensional $C^2$ manifold with
reach at least $\tau>0$. Let $x\in \mathcal M$, and let $T_x\mathcal M$ denote
the tangent space at $x$. Let
\[
    P_T := P_{T_x\mathcal{M}}
    \quad\text{and}\quad
    P_N := I-P_T
\]
be the tangent and normal projections.

For the $k$ nearest neighbors of $x$, $N_k(x)=\{i_1,\ldots,i_k\}$, define the maximal distance to a neighbor as
\[
    \rho_k(x):=\max_{j\le k}\|x_{i_j}-x\|_2.
\]
For notational simplicity, we write $\rho := \rho_k(x)$.
We work locally on the assumption that this distance is sufficiently small with respect to the reach of the manifold, i.e. $\rho_k(x)\le c\tau,$ for $c>0$ sufficiently small. This ensures that
$\mathcal M\cap B(x,\rho_k(x))$ is well approximated by the tangent space.
As before, let the empirical local covariance be
\[
    \widehat \Sigma_{x,k}^{\mathrm{loc}}=\frac{1}{k-1}\sum_{j=1}^k(x_{i_j}-\widehat \mu_x)(x_{i_j}-\widehat \mu_x)^\top,
\]
with empirical mean $\widehat \mu_x =\frac{1}{k}\sum_{j=1}^k x_{i_j}.$
Let the ideal tangent covariance be
\[
    \Sigma_{x,\rho_k}^{\star}=\operatorname{Cov}\left(P_T(X-x)\,\middle|\,X\in \mathcal M\cap B(x,\rho_k(x))\right).
\]

\subsubsection{Local chart and curvature control}

Since $\mathcal M$ has reach at least $\tau$, for every
$y\in \mathcal M\cap B(x,\rho)$ with $\rho\le c\tau$, we may write
\[
    y = x+t+q(t),
\]
where
\[
    t\in T_x\mathcal M,
    \qquad
    q(t)\in T_x\mathcal M^\perp.
\]
The vector $t$ describes tangent displacement, while $q(t)$ describes the
normal deviation due to curvature.
Standard reach bounds imply that, for a universal constant $C>0$,
\[
    \|q(t)\|_2 \le C\frac{\|t\|_2^2}{\tau}.
\]
Since $\|y-x\|_2\le \rho$, we also have $\|t\|_2\le C\rho$, and therefore
\[
    \|q(t)\|_2 \le C\frac{\rho^2}{\tau}.
\]
Thus local normal displacement is second order in the neighborhood radius.

\subsubsection{Curvature bias of the population covariance}

Let $Y$ be a local population sample distributed as $X\in \mathcal M\cap B(x,\rho)$.
Let
\[
    Y-x = T+Q,
\]
where
\[
    T := P_T(Y-x),\qquad Q := P_N(Y-x).
\]
By construction, we have  $\Sigma_{x,\rho}^{\star} = \operatorname{Cov}(T).$
The actual population covariance of the curved local neighborhood is
\[
    \Sigma_{x,\rho}^{\mathrm{pop}} := \operatorname{Cov}(Y-x) = \operatorname{Cov}(T+Q).
\]
Expanding this covariance gives
\[
    \Sigma_{x,\rho}^{\mathrm{pop}} = \operatorname{Cov}(T) + \operatorname{Cov}(Q)
    + \operatorname{Cov}(T,Q) + \operatorname{Cov}(Q,T).
\]
Therefore,
\[
    \Sigma_{x,\rho}^{\mathrm{pop}} - \Sigma_{x,\rho}^{\star}
    = \operatorname{Cov}(Q) + \operatorname{Cov}(T,Q) + \operatorname{Cov}(Q,T).
\]
From the local chart bounds we can obtain bounds for the tangent and normal projection,
\[
    \|T\|_2 \le C\rho,\qquad \|Q\|_2 \le C\frac{\rho^2}{\tau},
\]
which provide us with
\[
    \|\operatorname{Cov}(Q)\|_2 \le C\frac{\rho^4}{\tau^2}.
\]
Moreover, by Cauchy--Schwarz,
\[
    \|\operatorname{Cov}(T,Q)\|_2 \le \left(\mathbb E\|T-\mathbb ET\|_2^2\right)^{1/2} \left(\mathbb E\|Q-\mathbb EQ\|_2^2\right)^{1/2} \le C\frac{\rho^3}{\tau}.
\]
The same bound holds for $\operatorname{Cov}(Q,T)$. Therefore, we get the curvature bias term
\[
    \left\| \Sigma_{x,\rho}^{\mathrm{pop}} - \Sigma_{x,\rho}^{\star} \right\|_2\le C\frac{\rho^3}{\tau}.
\]

If the local distribution is symmetric in tangent coordinates, the mixed terms
$\operatorname{Cov}(T,Q)$ and $\operatorname{Cov}(Q,T)$ may cancel to leading
order. In that case, the curvature bias can improve to
$O\left(\frac{\rho^4}{\tau^2}\right)$.
We use the more conservative $O\left(\frac{\rho^3}{\tau}\right)$
bound in the main theorem.

\subsubsection{Finite-sample covariance concentration}

We next compare the empirical covariance
$\widehat \Sigma_{x,k}^{\mathrm{loc}}$ to the local population covariance
$\Sigma_{x,\rho}^{\mathrm{pop}}$.
Conditioned on the event that the $k$ neighbors lie in $B(x,\rho)$, the samples
are bounded by radius $\rho$ and are intrinsically $m$-dimensional up to the
curvature perturbation controlled above. Under the stated sub-Gaussian
assumption in tangent coordinates, a standard sample covariance concentration
bound yields, with probability at least $1-\delta$,
\[
    \left\|\widehat \Sigma_{x,k}^{\mathrm{loc}}-\Sigma_{x,\rho}^{\mathrm{pop}}\right\|_2
    \le C\rho^2\left(\sqrt{\frac{m+\log(1/\delta)}{k}}+\frac{m+\log(1/\delta)}{k}\right).
\]
The factor $\rho^2$ appears because covariance scales quadratically with the
diameter of the local neighborhood.
Using the empirical mean instead of the population mean does not change this
rate. Indeed, if $\mu=\mathbb E[Y]$, then
\[
    \widehat \Sigma_{x,k}^{\mathrm{loc}}-\Sigma_{x,\rho}^{\mathrm{pop}}
    =\left[\frac{1}{k}\sum_{j=1}^k(x_{i_j}-\mu)(x_{i_j}-\mu)^\top
    -\Sigma_{x,\rho}^{\mathrm{pop}}\right]-(\widehat \mu_x-\mu)(\widehat \mu_x-\mu)^\top.
\]
The empirical mean term is also controlled at the same order under the
sub-Gaussian assumption.

\subsubsection{Covariance approximation bound}

Combining the sampling error curvature bias from the previous two sections through
the triangle inequality, we get
\[
\begin{aligned}
    \left\|\widehat \Sigma_{x,k}^{\mathrm{loc}}-\Sigma_{x,\rho}^{\star}\right\|_2
    &\le\left\|\widehat \Sigma_{x,k}^{\mathrm{loc}}-\Sigma_{x,\rho}^{\mathrm{pop}}\right\|_2+\left\|\Sigma_{x,\rho}^{\mathrm{pop}}-\Sigma_{x,\rho}^{\star}\right\|_2\\
    &\le C_1\rho^2\left(\sqrt{\frac{m+\log(1/\delta)}{k}}+\frac{m+\log(1/\delta)}{k}\right)+C_2\frac{\rho^3}{\tau}.
\end{aligned}
\]
With $\rho=\rho_k(x)$ we get
\[
    \left\|\widehat \Sigma_{x,k}^{\mathrm{loc}}-\Sigma_{x,\rho_k}^{\star}\right\|_2\le\eta_k(x,\delta),
\]
where
\[
    \eta_k(x,\delta):=C_1\rho_k(x)^2\left(\sqrt{\frac{m+\log(1/\delta)}{k}}+
    \frac{m+\log(1/\delta)}{k}\right)+C_2\frac{\rho_k(x)^3}{\tau}.
\]

\subsubsection{Eigenvalue convergence}

Let $\widehat \lambda_{x,1}\ge \cdots \ge\widehat \lambda_{x,d}$
be the eigenvalues of $\widehat \Sigma_{x,k}^{\mathrm{loc}}$, and let
\[
    \lambda_{x,1}^\star\ge \cdots \ge\lambda_{x,m}^\star>\lambda_{x,m+1}^\star=\cdots=\lambda_{x,d}^\star=0
\]
be the eigenvalues of the rank-$m$ tangent covariance
$\Sigma_{x,\rho_k}^{\star}$. By Weyl's inequality, we have
\[
    |\widehat \lambda_{x,j}-\lambda_{x,j}^\star|\le\left\|\widehat \Sigma_{x,k}^{\mathrm{loc}}-\Sigma_{x,\rho_k}^{\star}\right\|_2.
\]
Hence, with probability at least $1-\delta$,
\[
    |\widehat \lambda_{x,j}-\lambda_{x,j}^\star|\le\eta_k(x,\delta)\qquad\text{for all } j=1,\ldots,d.
\]
In particular, for $j=1,\ldots,m$,
$|\widehat \lambda_{x,j}-\lambda_{x,j}^\star|\le\eta_k(x,\delta),$
and the leading normal eigenvalue is bounded by
$\widehat \lambda_{x,m+1}\le\eta_k(x,\delta)$.

\subsubsection{Tangent subspace convergence}

Let $\widehat U_{x,m}$ contain the top $m$ eigenvectors of
$\widehat \Sigma_{x,k}^{\mathrm{loc}}$, and let $U_x^\star$ contain the top
$m$ eigenvectors of $\Sigma_{x,\rho_k}^{\star}$. The tangent--normal eigengap is
$\gamma_x = \lambda_{x,m}^\star-\lambda_{x,m+1}^\star = \lambda_{x,m}^\star.$
By the Davis--Kahan sin-theta theorem, we have
\[
    \left\|\sin\Theta(\widehat U_{x,m},U_x^\star)\right\|_2
    \le\frac{\left\|\widehat \Sigma_{x,k}^{\mathrm{loc}}-\Sigma_{x,\rho_k}^{\star}\right\|_2}{\gamma_x-\left\|\widehat \Sigma_{x,k}^{\mathrm{loc}}-\Sigma_{x,\rho_k}^{\star}\right\|_2}.
\]
If $\eta_k(x,\delta)\le \frac{\gamma_x}{4},$
then the denominator is at least $\gamma_x/2$, and therefore
\[
    \left\|\sin\Theta(\widehat U_{x,m},U_x^\star)\right\|_2\le\frac{2\eta_k(x,\delta)}{\gamma_x}.
\]
This proves the stated subspace convergence bound.
\qed

\subsubsection{Inverse Mahalanobis metric perturbation}

Our certificate uses the inverse covariance metric. We therefore compare
\[
    \widehat A_x=\widehat U_{x,m}\widehat \Lambda_{x,m}^{-1}\widehat U_{x,m}^\top
\]
with
\[
    A_x^\star=U_x^\star(\Lambda_x^\star)^{-1}U_x^{\star\top}.
\]

Assume $\eta_k(x,\delta)\le \frac{\lambda_{x,m}^\star}{4}.$
Then, by the eigenvalue convergence bound,
\[
    \widehat \lambda_{x,j}\ge\lambda_{x,j}^\star-\eta_k(x,\delta)\ge\frac{1}{2}\lambda_{x,j}^\star\ge\frac{1}{2}\lambda_{x,m}^\star
\]
for all $j=1,\ldots,m$. Hence
\[
    \left|\frac{1}{\widehat \lambda_{x,j}}-\frac{1}{\lambda_{x,j}^\star}\right|
    =\frac{|\widehat \lambda_{x,j}-\lambda_{x,j}^\star|}{\widehat \lambda_{x,j}\lambda_{x,j}^\star}
    \le\frac{2\eta_k(x,\delta)}{(\lambda_{x,m}^\star)^2}.
\]

The remaining error comes from the perturbation of the eigenspace. Since
\[
    \|(\Lambda_x^\star)^{-1}\|_2=\frac{1}{\lambda_{x,m}^\star},
\]
and since Davis--Kahan gives
\[
    \|\widehat U_{x,m}\widehat U_{x,m}^\top-U_x^\star U_x^{\star\top}\|_2
    \le C\frac{\eta_k(x,\delta)}{\gamma_x},
\]
we obtain
\[
    \left\|\widehat A_x-A_x^\star\right\|_2
    \le C_3\eta_k(x,\delta)\left(\frac{1}{(\lambda_{x,m}^\star)^2}+\frac{1}{\lambda_{x,m}^\star\gamma_x}\right).
\]
The first term is caused by perturbing the inverse eigenvalues, while the
second is caused by perturbing the PCA subspace.

\subsubsection{Set-level convergence of ellipsoids}
\label{app:theory_conv_ellipsoids}

For any $z\in B(x,R)$, from the previous section we know

\begin{align*}
    \left|(z-x)^\top(\widehat A_x-A_x^\star)(z-x)\right|&\le\|z-x\|_2^2\|\widehat A_x-A_x^\star\|_2\\
    &\le R^2 C_3\eta_k(x,\delta)\left(\frac{1}{(\lambda_{x,m}^\star)^2}+\frac{1}{\lambda_{x,m}^\star\gamma_x}\right).
\end{align*}
Now let this upper bound be $\zeta_R$, i.e.,
\[
    \zeta_R:=R^2 C_3\eta_k(x,\delta)\left(\frac{1}{(\lambda_{x,m}^\star)^2}+\frac{1}{\lambda_{x,m}^\star\gamma_x}\right).
\]
Rewriting the above, we have
\[
    \left|(z-x)^\top\widehat A_x(z-x)-(z-x)^\top A_x^\star(z-x)\right|\le\zeta_R
\]
for all $z\in B(x,R)$.
Now suppose  $z\in E_x^\star\!\left(\sqrt{r^2-\zeta_R}\right)\cap B(x,R).$
Then we get
\[
    (z-x)^\top A_x^\star(z-x)\le r^2-\zeta_R.
\]
Using the metric perturbation bound, it follows that
\[
    (z-x)^\top \widehat A_x(z-x)\le r^2,
\]
and therefore, $z\in \widehat E_x(r).$

This proves
\[
    E_x^\star\!\left(\sqrt{r^2-\zeta_R}\right)\cap B(x,R)
    \subseteq
    \widehat E_x(r)\cap B(x,R).
\]

Conversely, suppose $z\in \widehat E_x(r)\cap B(x,R).$
Then similarly as above we have
\[
    (z-x)^\top \widehat A_x(z-x)\le r^2.
\]
Again applying the perturbation bound, we get
\[
    (z-x)^\top A_x^\star(z-x)\le r^2+\zeta_R,
\]
and therefore $z\in E_x^\star\!\left(\sqrt{r^2+\zeta_R}\right)$.
This proves
\[
    \widehat E_x(r)\cap B(x,R)\subseteq
    E_x^\star\!\left(\sqrt{r^2+\zeta_R}\right)\cap B(x,R).
\]
Combining the two inclusions gives
\[
    E_x^\star\!\left(\sqrt{r^2-\zeta_R}\right)\cap B(x,R)\subseteq
    \widehat E_x(r)\cap B(x,R)\subseteq
    E_x^\star\!\left(\sqrt{r^2+\zeta_R}\right)\cap B(x,R).
\]
This completes the proof of
Theorem~\ref{thm:main-ellipsoid-convergence}. \qed

\subsubsection{Implications on choosing the neighborhood size}

The bound contains two competing terms:
\[
    \eta_k(x,\delta)=
    \underbrace{C_1\rho_k(x)^2\left(\sqrt{\frac{m+\log(1/\delta)}{k}}+\frac{m+\log(1/\delta)}{k}\right)}_{\text{finite-sample covariance error}}+
    \underbrace{
    C_2\frac{\rho_k(x)^3}{\tau}
    }_{\text{curvature bias}}.
\]
Increasing $k$ reduces the finite-sample covariance error, but it also increases the neighborhood radius $\rho_k(x)$, which increases the curvature bias.
If the training data consist of $n_{\mathcal{D}}$ samples drawn from a sufficiently regular
density on an $m$-dimensional manifold, then locally
\[
    \rho_k(x)\asymp\left(\frac{k}{n_{\mathcal{D}}}\right)^{1/m}.
\]
Substituting this into the relative error gives the heuristic rate
\[
    O\left(\sqrt{\frac{m+\log(1/\delta)}{k}}+\frac{1}{\tau}\left(\frac{k}{n_{\mathcal{D}}}\right)^{1/m}\right).
\]
Thus, the number of neighbors $k$ should be large enough to make the empirical
PCA stable, but not so large that the local neighborhood ceases to be well
approximated by a tangent ellipsoid.



\section{Implementation Details}

\paragraph{Implementation.}
We implement all experiments in Python using PyTorch. Approximate
nearest-neighbor search is performed using Annoy~\citep{bernhardsson2018annoy} with 50 trees, with
Euclidean distance for image-space experiments and angular distance for
CLIP retrieval. Local PCA is computed using scikit-learn, with eigenvalues
lower-bounded by $10^{-6}$ for numerical stability. For each input, the
neighborhood and local PCA decomposition are computed once and reused
across all Monte Carlo samples.

\label{app:experiment_details}
\subsection{Characterizing Perturbation Geometry}
\label{sec:perturbation-volume}

The noise scale $\sigma$ alone does not capture the geometric extent of
an anisotropic perturbation: two smoothing distributions with the same
$\sigma$ may occupy substantially different regions depending on their
covariance structure. We therefore use volume as a scalar summary of the
perturbation geometry, following its geometric use for comparing
anisotropic regions in randomized smoothing~\citep{eiras2022ancer}. Here,
the volume characterizes the \emph{smoothing perturbation distribution},
rather than the certified region. For a $d$-dimensional ball of radius
$r$,
\begin{equation}
    V(r) = C_d r^d,
    \qquad
    C_d =
    \frac{\pi^{d/2}}
         {\Gamma\!\left(\frac{d}{2}+1\right)},
\end{equation}
where $C_d$ is the volume of the $d$-dimensional unit ball.

Under isotropic smoothing, the perturbation scale is identical along all
$d$ directions. Setting $r=\sigma$ gives
\begin{equation}
    V_{\mathrm{I}} = C_d \sigma^d.
\end{equation}
For manifold smoothing, the local covariance $\Lambda_m$ rescales each
retained direction $m$, inducing ellipsoidal semi-axes
\begin{equation}
    a_j = \sigma\sqrt{\lambda_j},
    \qquad j=1,\ldots,m,
\end{equation}
where $\lambda_j$ is the $j$-th retained eigenvalue. The resulting
perturbation volume is
\begin{equation}
\label{eq:mani-perturbation-volume}
    V_{\mathrm{M}}
    =
    C_m \prod_{j=1}^{m} a_j
    =
    C_m \sigma^m \sqrt{\det(\Lambda_m)}.
\end{equation}

Thus, $\sigma^m$ captures the global perturbation scale, while
$\sqrt{\det(\Lambda_m)}$ captures the local anisotropic scaling induced
by the covariance spectrum. Since volume scales exponentially with
dimension, we report $\log V$ throughout the experiments.

\subsection{Algorithm: Randomized Manifold Smoothing}
\label{app:pseudocode}
Algorithm ~\ref{alg:manifold_certify} expresses the 
per-input certification procedure used in our experiments.
For each input $x$, we estimate its local geometry once and use it
throughout the Monte Carlo sampling procedure. The classification and segmentation then use the standard certification
procedures of \citet{cohen2019certified} and
\citet{Fischer2021ScalableSegmentation}, respectively.
\begin{algorithm}[H]
\caption{Randomized Manifold Smoothing}
\label{alg:manifold_certify}
\begin{algorithmic}

\Require predictor $f$, input $x$, dataset $\mathcal D$,
         $k,m,\sigma,n_0,n,\alpha$, task $t$,
         (threshold $\tau$ for segmentation)


\State $N_k(x) \gets \Call{NearestNeighbors}{\mathcal D,x,k}$

\State $\Sigma_x^{\mathrm{loc}}
       \gets \Call{LocalCovariance}{N_k(x)}$
       \Comment{Eq.~\ref{eq:covar}}

\State $(U_{x,m},\Lambda_{x,m})
       \gets \Call{LocalPCA}{\Sigma_x^{\mathrm{loc}},m}$


\Function{ManifoldSample}{$f,\sigma,x,N,U_{x,m},\Lambda_{x,m}$}

    \State $\mathrm{cnts}\gets 0$

    \For{$i\gets1,\ldots,N$}

        \State $\varepsilon_i
        \sim \mathcal N(0,\sigma^2 I_m)$

        \State $\delta_i
        \gets U_{x,m}\Lambda_{x,m}^{1/2}\varepsilon_i$ \Comment{Eq.~\ref{eq:sample}}

        \State $\hat y_i \gets f(x+\delta_i)$

        \State update $\mathrm{cnts}$ with $\hat y_i$

    \EndFor

    \State \Return $\mathrm{cnts}$

\EndFunction


\Statex \textit{\# task-specific certification with manifold sampling}

\If{$t=\mathrm{classification}$}

    \State \Return
    $\Call{Certify}
    {f,\sigma,x,n_0,n,\alpha;
     \textsc{ManifoldSample}}$
    \Comment{\citep{cohen2019certified}}

\Else
    \Statex \textit{\# segmentation}
    \State \Return
    $\Call{SegCertify}
    {f,\sigma,x,n,n_0,\tau,\alpha;
     \textsc{ManifoldSample}}$
    \Comment{\citep{Fischer2021ScalableSegmentation}}

\EndIf

\end{algorithmic}
\end{algorithm}

\subsection{Models and Training}
\label{app:model_training}

\paragraph{Image Classification and Segmentation.}
For CelebA~\citep{liu2015faceattributes} and CelebA-HQ~\citep{karras2018progressive}. classification, we train ResNet-18 smile
classifiers and keep the trained models fixed during all smoothing
experiments. For semantic segmentation, we use BiSeNet~\citep{yu2018bisenet} model with
publicly available pretrained weights from the official face-parsing
implementation\footnote{\url{https://github.com/zllrunning/face-parsing.PyTorch}}
on CelebAMask-HQ~\citep{CelebAMask-HQ}. The model predicts 19 facial
semantic classes and is used without additional fine-tuning.

\paragraph{Concept-Space Classification.}
For concept-space experiments, we use the pretrained Concept Foundation
Model (CFM)~\citet{Wittenmayer2026CFM} \footnote{\url{https://github.com/kawi19/CFM}}. CFM maps CLIP-DINO image features
through a sparse autoencoder (SAE) to an interpretable sparse concept
representation, followed by a pretrained linear classifier on ImageNet~\citep{deng2009imagenet}.
We keep both the representation and classifier fixed and apply smoothing
directly to the SAE concept vector.

\begin{table}[H]
\centering
\caption{
\textbf{Predictive performance of the base models used across experiments.}
Performance is reported prior to randomized smoothing and certification.
}
\label{tab:base_models}

\small
\begin{tabular}{lll}
\toprule
\textbf{Dataset} & \textbf{Model} & \textbf{Base performance} \\
\midrule
CelebA~\citep{liu2015faceattributes}       & ResNet-18 & 92.0\% Acc. \\
CelebA-HQ~\citep{karras2018progressive}     & ResNet-18 & 94.3\% Acc. \\
CelebAMask-HQ~\citep{CelebAMask-HQ} & BiSeNet   & 95.3\% Pixel Acc.; 71.5\% mIoU \\
ImageNet~\citep{deng2009imagenet}      & CFM       & 78.6\% Acc. \\
\bottomrule
\end{tabular}
\end{table}

\paragraph{OOD Classifier Training.}
\label{app:ood_training}
For each OOD experiment, we select a CelebA facial attribute $a$ and
construct an attribute-defined distribution shift. The smile classifier is
trained only on images where the attribute is absent,
\[
D_{\mathrm{train}}=\{(x,y)\mid a(x)=0\},
\]
and evaluated only where it is present,
\[
D_{\mathrm{test}}=\{(x,y)\mid a(x)=1\},
\]
where $y$ denotes the Smile/No-Smile label. Certification uses a disjoint
subset with $a=1$ from the original training split. To control for
differences in attribute frequency, all OOD experiments use the same
number of examples, determined by the smallest available attribute subset,
with balanced Smile/No-Smile samples for both classifier training and
certification. Table~\ref{tab:ood_classifier_acc} reports accuracy before
certification.

\begin{table}[t]
\centering
\caption{
\textbf{Base classifier test accuracy for the CelebA OOD experiments.}
Accuracy is reported before smoothing and certification for the standard
and attribute-specific classifiers.
}
\label{tab:ood_classifier_acc}

\small
\setlength{\tabcolsep}{10pt}
\renewcommand{\arraystretch}{1.0}

\begin{tabular}{lc}
\toprule
\textbf{Classifier} & \textbf{Acc.} (\%) \\
\midrule
Base (All Data)       & 92.0 \\
Eyeglasses            & 91.9 \\
High Cheekbones       & 80.0 \\
Male                  & 90.6 \\
Mouth Slightly Open   & 94.0 \\
Mustache              & 90.9 \\
No Beard              & 89.8 \\
\bottomrule
\end{tabular}
\end{table}

\section{Evaluation Metrics}
\label{app:evaluation}
\paragraph{Prediction and certification.}
For classification, we report certified accuracy (CAcc), defined as the
fraction of evaluated examples for which the smoothed classifier returns
the correct class together with a valid certificate. Examples for which
the certification test does not return a prediction are counted as
abstentions $\varnothing$. For semantic segmentation, the same quantities are evaluated
pixel-wise: certified pixel accuracy is the fraction of correctly
classified and certified pixels, while the abstention rate
$\varnothing$ is the fraction of pixels for which no certified prediction
is returned. We additionally report certified mean intersection-over-union
(mIoU), computed from the certified segmentation predictions. Manifold certification radii are reported in the corresponding
whitened smoothing space as
$r_{\mathrm{white}}$. For geometric comparisons, we report the log-volume $\log V$ of the
$\sigma$-scaled smoothing region defined in Sec.~\ref{sec:perturbation-volume}.
Since raw volume scales strongly with the representation dimension $m$,
in ablation work \ref{app:ablation_study} we report the dimension-normalized quantity $\log V/m$
when comparing different manifold dimensions. This provides a
per-dimension measure of the geometric extent of the smoothing
distribution.

\paragraph{Task-specific robustness metrics.}
For image-to-text retrieval, Recall@1 (R@1) denotes the fraction of queries
for which the correct caption is ranked first. Following RoCOCO, RSMS
measures the fraction of semantically manipulated captions that are
retrieved at rank one, with lower values indicating greater robustness.
We report the relative retrieval degradation with respect to performance
on the clean COCO test set as
$\mathrm{Drop}=(R_{\mathrm{clean}}-R_{\mathrm{corr}})/
R_{\mathrm{clean}}\times100\%$,
where $R_{\mathrm{clean}}$ is R@1 on the original COCO gallery and
$R_{\mathrm{corr}}$ is R@1 under the corresponding RoCOCO corruption.
For segmentation analyses, per-class IoU is computed independently for
each semantic class, and $\Delta\mathrm{IoU}$ denotes the difference
between manifold and isotropic smoothing for the same noise scale.

\section{Ablation Study}
\label{app:ablation_study}
We ablated the key hyperparameters and settings governing manifold
estimation and certification. Specifically, we varied the neighborhood
size $k$ and retained PCA dimension $m$ to assess sensitivity to the
estimated local geometry. We additionally varied the number of Monte
Carlo samples $n$ and evaluation subset size $N$ to assess the stability
of the certification results.

The reported radii have different geometric interpretations. Isotropic
smoothing certifies an $\ell_2$ radius $R$ in the ambient input space,
whereas manifold smoothing certifies $r_{\mathrm{white}}$ in the whitened
local PCA space, corresponding to an ellipsoidal region in the retained
subspace. Their magnitudes are therefore not directly comparable.
Similarly, we report dimension-normalized log-volume only as a geometric
diagnostic, since volumes over different-dimensional subspaces are not
directly comparable.

\paragraph{Sensitivity to local neighborhood size $k$.}
The full-rank setting in Fig.~\ref{fig:ablation_knn_k}(a) shows that at larger noise scales,
extreme neighborhood sizes ($k=128$ and $k=1000$) generally yield lower
certified accuracy and higher abstention than intermediate choices. This is
consistent with the local-geometry trade-off discussed in
App.~\ref{app:theory_conv_ellipsoids}: small neighborhoods provide fewer
samples for estimating local geometry, whereas large neighborhoods may span
regions of greater manifold curvature. Because the available PCA rank also
changes with $k$, we repeat the experiment with $m=100$ fixed
(Fig.~\ref{fig:ablation_knn_k}(b)). Sensitivity remains visible, with
moderate neighborhoods ($k=250$--$750$) generally providing more stable
certification than the extreme choices, indicating an effect of neighborhood
locality beyond the changing PCA dimension.


\definecolor{ablBlue}{RGB}{45,110,180}
\definecolor{ablOrange}{RGB}{230,120,30}
\definecolor{ablGreen}{RGB}{55,150,70}
\definecolor{ablRed}{RGB}{205,65,65}
\definecolor{ablPurple}{RGB}{130,85,175}


\newcommand{\kAblationCSV}{
data/ablation/celeba/k_table.csv
}

%

\newcommand{\plotKabl}[3]{%
\addplot[
    color=#3,
    line width=0.4pt,
    mark=*,
    mark size=1pt,
    restrict expr to domain={\thisrow{parameter}}{#1:#1},
    unbounded coords=discard
]
table[
    col sep=comma,
    x=sigma,
    y expr={#2}
]{\kAblationCSV};
}


\newcommand{\kFixedAblationCSV}{
data/ablation/celeba/k_fixed_d100_table.csv
}

\newcommand{\plotKFixedAbl}[3]{%
\addplot[
    color=#3,
    line width=0.4pt,
    mark=*,
    mark size=1pt,
    restrict expr to domain={\thisrow{parameter}}{#1:#1},
    unbounded coords=discard
]
table[
    col sep=comma,
    x=sigma,
    y expr={#2}
]{\kFixedAblationCSV};
}


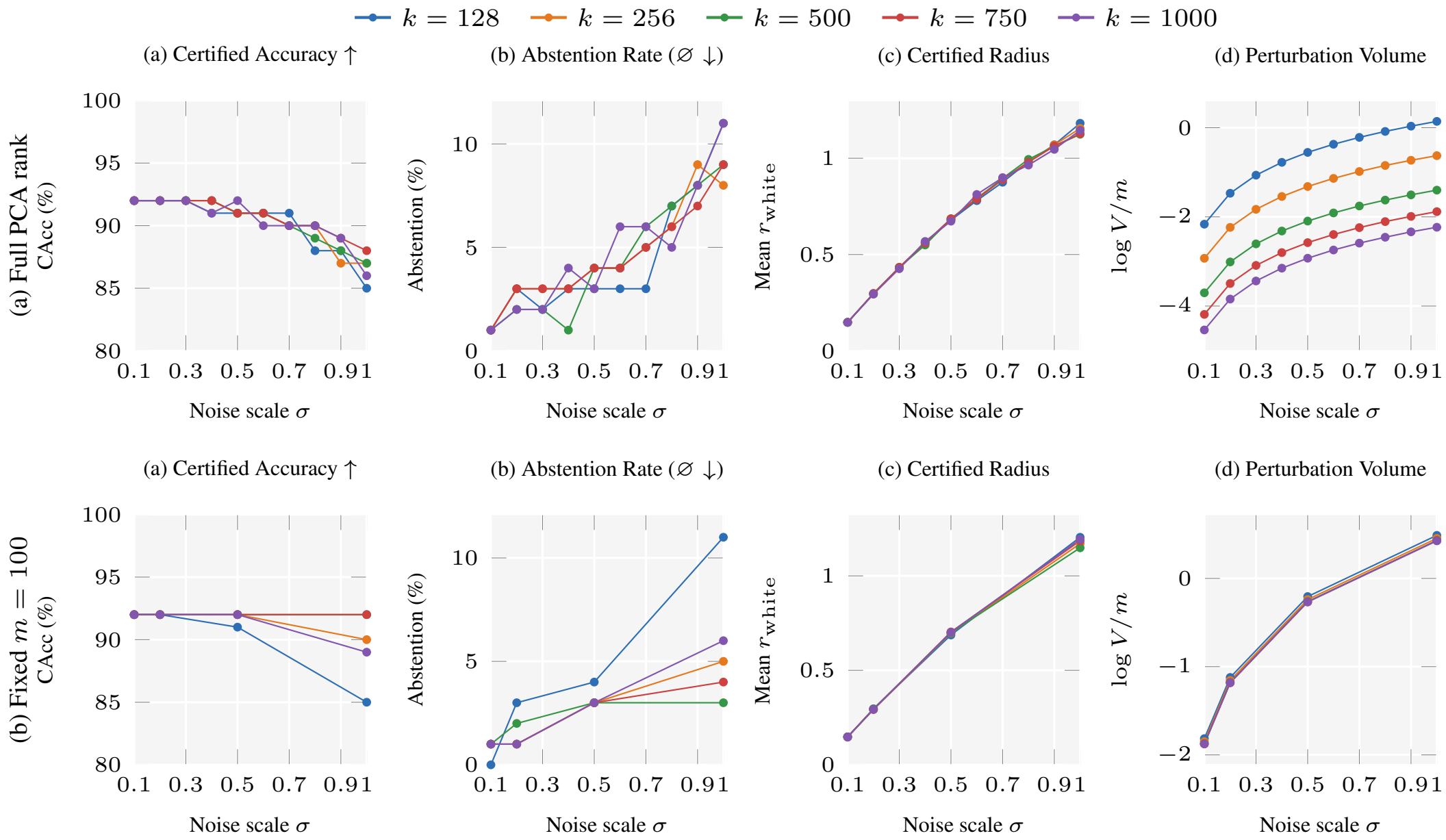
\begin{figure}[t]
\centering

\begin{tikzpicture}

\begin{groupplot}[
    group style={
    group name=fullrank,
    group size=4 by 1,
    horizontal sep=1.2cm,
},
    width=0.274\textwidth,
    height=4.0cm,
    xmin=0.1,
    xmax=1.0,
    xtick={0.1,0.3,0.5,0.7,0.9,1.0},
    xlabel={Noise scale $\sigma$},
    axis background/.style={
        fill=gray!8
    },
    grid=major,
    major grid style={
        draw=white,
        line width=0.6pt
    },
    tick label style={
        font=\tiny
    },
    label style={
        font=\tiny
    },
    title style={
        font=\tiny
    },
     axis line style={draw=white},
]


\nextgroupplot[
    title={(a) Certified Accuracy $\uparrow$},
    ylabel={CAcc (\%)},
     title style={font=\tiny},
    ymin=80,
    ymax=100,
    ytick={80,85,90,95,100},
]

\plotKabl{128}
{\thisrow{certified_accuracy_pct}}
{ablBlue}

\plotKabl{256}
{\thisrow{certified_accuracy_pct}}
{ablOrange}

\plotKabl{500}
{\thisrow{certified_accuracy_pct}}
{ablGreen}

\plotKabl{750}
{\thisrow{certified_accuracy_pct}}
{ablRed}

\plotKabl{1000}
{\thisrow{certified_accuracy_pct}}
{ablPurple}


\nextgroupplot[
    title={(b) Abstention Rate ($\varnothing \downarrow$)},
    ylabel={Abstention (\%)},
     title style={font=\tiny},
    ymin=0,
]

\plotKabl{128}
{\thisrow{abstention_pct}}
{ablBlue}

\plotKabl{256}
{\thisrow{abstention_pct}}
{ablOrange}

\plotKabl{500}
{\thisrow{abstention_pct}}
{ablGreen}

\plotKabl{750}
{\thisrow{abstention_pct}}
{ablRed}

\plotKabl{1000}
{\thisrow{abstention_pct}}
{ablPurple}


\nextgroupplot[
    title={(c) Certified Radius},
     title style={font=\tiny},
    ylabel={Mean $r_{\mathrm{white}}$},
    ymin=0,
]

\plotKabl{128}
{\thisrow{mean_radius}}
{ablBlue}

\plotKabl{256}
{\thisrow{mean_radius}}
{ablOrange}

\plotKabl{500}
{\thisrow{mean_radius}}
{ablGreen}

\plotKabl{750}
{\thisrow{mean_radius}}
{ablRed}

\plotKabl{1000}
{\thisrow{mean_radius}}
{ablPurple}


\nextgroupplot[
    title={(d) Perturbation Volume},
     title style={font=\tiny},
    ylabel={$\log V / m$},
]

\plotKabl{128}{
    \thisrow{mean_log_volume}/
    \thisrow{volume_dimension}
}{ablBlue}

\plotKabl{256}{
    \thisrow{mean_log_volume}/
    \thisrow{volume_dimension}
}{ablOrange}

\plotKabl{500}{
    \thisrow{mean_log_volume}/
    \thisrow{volume_dimension}
}{ablGreen}

\plotKabl{750}{
    \thisrow{mean_log_volume}/
    \thisrow{volume_dimension}
}{ablRed}

\plotKabl{1000}{
    \thisrow{mean_log_volume}/
    \thisrow{volume_dimension}
}{ablPurple}

\end{groupplot}

\node[
    rotate=90,
    font=\scriptsize,
    anchor=center
]
at ([xshift=-11mm]
$(fullrank c1r1.north west)!0.5!(fullrank c1r1.south west)$)
{(a) Full PCA rank};


\begin{scope}[yshift=-4cm]

\begin{groupplot}[
group style={
    group name=fixedrank,
    group size=4 by 1,
    horizontal sep=1.2cm,
},
    width=0.274\textwidth,
    height=4.0cm,
    xmin=0.1,
    xmax=1.0,
    xtick={0.1,0.3,0.5,0.7,0.9,1.0},
    xlabel={Noise scale $\sigma$},
    axis background/.style={fill=gray!8},
    grid=major,
    major grid style={
        draw=white,
        line width=0.6pt
    },
    tick label style={font=\tiny},
    label style={font=\tiny},
    title style={font=\tiny},
    axis line style={draw=white},
]


\nextgroupplot[
    title={(a) Certified Accuracy $\uparrow$},
    ylabel={CAcc (\%)},
    ymin=80,
    ymax=100,
    ytick={80,85,90,95,100},
]

\plotKFixedAbl{128}{\thisrow{certified_accuracy_pct}}{ablBlue}
\plotKFixedAbl{256}{\thisrow{certified_accuracy_pct}}{ablOrange}
\plotKFixedAbl{500}{\thisrow{certified_accuracy_pct}}{ablGreen}
\plotKFixedAbl{750}{\thisrow{certified_accuracy_pct}}{ablRed}
\plotKFixedAbl{1000}{\thisrow{certified_accuracy_pct}}{ablPurple}


\nextgroupplot[
    title={(b) Abstention Rate ($\varnothing \downarrow$)},
    ylabel={Abstention (\%)},
    ymin=0,
]

\plotKFixedAbl{128}{\thisrow{abstention_pct}}{ablBlue}
\plotKFixedAbl{256}{\thisrow{abstention_pct}}{ablOrange}
\plotKFixedAbl{500}{\thisrow{abstention_pct}}{ablGreen}
\plotKFixedAbl{750}{\thisrow{abstention_pct}}{ablRed}
\plotKFixedAbl{1000}{\thisrow{abstention_pct}}{ablPurple}


\nextgroupplot[
    title={(c) Certified Radius},
    ylabel={Mean $r_{\mathrm{white}}$},
    ymin=0,
]

\plotKFixedAbl{128}{\thisrow{mean_radius}}{ablBlue}
\plotKFixedAbl{256}{\thisrow{mean_radius}}{ablOrange}
\plotKFixedAbl{500}{\thisrow{mean_radius}}{ablGreen}
\plotKFixedAbl{750}{\thisrow{mean_radius}}{ablRed}
\plotKFixedAbl{1000}{\thisrow{mean_radius}}{ablPurple}


\nextgroupplot[
    title={(d) Perturbation Volume},
    ylabel={$\log V / m$},
]

\plotKFixedAbl{128}{
    \thisrow{mean_log_volume}/\thisrow{volume_dimension}
}{ablBlue}

\plotKFixedAbl{256}{
    \thisrow{mean_log_volume}/\thisrow{volume_dimension}
}{ablOrange}

\plotKFixedAbl{500}{
    \thisrow{mean_log_volume}/\thisrow{volume_dimension}
}{ablGreen}

\plotKFixedAbl{750}{
    \thisrow{mean_log_volume}/\thisrow{volume_dimension}
}{ablRed}

\plotKFixedAbl{1000}{
    \thisrow{mean_log_volume}/\thisrow{volume_dimension}
}{ablPurple}

\end{groupplot}
\node[
    rotate=90,
    font=\scriptsize,
    anchor=center
]
at ([xshift=-11mm]
$(fixedrank c1r1.north west)!0.5!(fixedrank c1r1.south west)$)
{(b) Fixed $m=100$};
\end{scope}

\node[
    anchor=south,
    inner sep=0pt
]
at
([yshift=6mm]
$(fullrank c1r1.north west)!0.5!(fullrank c4r1.north east)$)
{%
\begin{tabular}{
@{}
c@{\hspace{8pt}}
c@{\hspace{8pt}}
c@{\hspace{8pt}}
c@{\hspace{8pt}}
c
@{}
}

\tikz[baseline=-0.5ex]{
    \draw[ablBlue,line width=0.8pt] (0,0)--(0.35,0);
    \fill[ablBlue] (0.175,0) circle (1.1pt);
}
{\scriptsize $k=128$}
&
\tikz[baseline=-0.5ex]{
    \draw[ablOrange,line width=0.8pt] (0,0)--(0.35,0);
    \fill[ablOrange] (0.175,0) circle (1.1pt);
}
{\scriptsize $k=256$}
&
\tikz[baseline=-0.5ex]{
    \draw[ablGreen,line width=0.8pt] (0,0)--(0.35,0);
    \fill[ablGreen] (0.175,0) circle (1.1pt);
}
{\scriptsize $k=500$}
&
\tikz[baseline=-0.5ex]{
    \draw[ablRed,line width=0.8pt] (0,0)--(0.35,0);
    \fill[ablRed] (0.175,0) circle (1.1pt);
}
{\scriptsize $k=750$}
&
\tikz[baseline=-0.5ex]{
    \draw[ablPurple,line width=0.8pt] (0,0)--(0.35,0);
    \fill[ablPurple] (0.175,0) circle (1.1pt);
}
{\scriptsize $k=1000$}

\end{tabular}
};

\end{tikzpicture}


\caption{
\textbf{Effect of neighborhood size $k$ under full-rank and fixed-dimensional local PCA.}
(a): the retained PCA rank follows the available local rank.
(b): the PCA dimension is fixed to $m=100$ for all neighborhood sizes.
Certified accuracy, abstention, mean manifold certified radius, and
dimension-normalized perturbation log-volume are reported across noise scales.
}
\label{fig:ablation_knn_k}

\end{figure}

\paragraph{Retained PCA dimension $m$ at fixed $k=500$.}
Figure~\ref{fig:ablation_pca_dim} shows that certification is largely
insensitive to the retained PCA dimension at small and moderate noise,
suggesting that the leading principal directions already capture most of
the locally relevant variation. At larger $\sigma$, retaining more
components can slightly reduce certified accuracy and increase abstention.
This is consistent with lower-variance PCA directions contributing weaker
or less reliable local structure, whose cumulative effect becomes more
pronounced under stronger perturbations. Thus, using the full local PCA basis is not necessary for strong
certification, and moderate truncation can act as a regularizer at large
perturbation scales.

\definecolor{ablBlue}{RGB}{45,110,180}
\definecolor{ablOrange}{RGB}{230,120,30}
\definecolor{ablGreen}{RGB}{55,150,70}
\definecolor{ablRed}{RGB}{205,65,65}
\definecolor{ablPurple}{RGB}{130,85,175}


\newcommand{\PCAAblationCSV}{
data/ablation/celeba/pca_table.csv
}

%

\newcommand{\plotPCAabl}[3]{%
\addplot[
    color=#3,
    line width=0.4pt,
    mark=*,
    mark size=1pt,
    restrict expr to domain={\thisrow{parameter}}{#1:#1},
    unbounded coords=discard
]
table[
    col sep=comma,
    x=sigma,
    y expr={#2}
]{\PCAAblationCSV};
}


\begin{figure}[t]
\centering

\begin{tikzpicture}

\begin{groupplot}[
    group style={
        group size=4 by 1,
        horizontal sep=1.2cm,
    },
    width=0.274\textwidth,
    height=4.0cm,
    xmin=0.1,
    xmax=1.0,
    xtick={0.1,0.3,0.5,0.7,0.9,1.0},
    xlabel={Noise scale $\sigma$},
    axis background/.style={
        fill=gray!8
    },
    grid=major,
    major grid style={
        draw=white,
        line width=0.6pt
    },
    tick label style={
        font=\tiny
    },
    label style={
        font=\tiny
    },
    title style={
        font=\tiny
    },
     axis line style={draw=white},
]


\nextgroupplot[
    title={(a) Certified Accuracy $\uparrow$},
    title style={font=\tiny},
    ylabel={CAcc (\%)},
    ymin=80,
    ymax=100,
    ytick={80,85,90,95,100},
]

\plotPCAabl{128}{\thisrow{certified_accuracy_pct}}{ablBlue}
\plotPCAabl{192}{\thisrow{certified_accuracy_pct}}{ablOrange}
\plotPCAabl{256}{\thisrow{certified_accuracy_pct}}{ablGreen}
\plotPCAabl{384}{\thisrow{certified_accuracy_pct}}{ablRed}
\plotPCAabl{499}{\thisrow{certified_accuracy_pct}}{ablPurple}


\nextgroupplot[
    title={(b) Abstention Rate ($\varnothing \downarrow$)},
     title style={font=\tiny},
    ylabel={Abstention (\%)},
    ymin=0,
]

\plotPCAabl{128}{\thisrow{abstention_pct}}{ablBlue}
\plotPCAabl{192}{\thisrow{abstention_pct}}{ablOrange}
\plotPCAabl{256}{\thisrow{abstention_pct}}{ablGreen}
\plotPCAabl{384}{\thisrow{abstention_pct}}{ablRed}
\plotPCAabl{499}{\thisrow{abstention_pct}}{ablPurple}


\nextgroupplot[
    title={(c) Certified Radius},
    ylabel={Mean $r_{\mathrm{white}}$},
    ymin=0,
     title style={font=\tiny},
]

\plotPCAabl{128}{\thisrow{mean_radius}}{ablBlue}
\plotPCAabl{192}{\thisrow{mean_radius}}{ablOrange}
\plotPCAabl{256}{\thisrow{mean_radius}}{ablGreen}
\plotPCAabl{384}{\thisrow{mean_radius}}{ablRed}
\plotPCAabl{499}{\thisrow{mean_radius}}{ablPurple}


\nextgroupplot[
    title={(d) Perturbation Volume},
     title style={font=\tiny},
    ylabel={$\log V / m$},
]

\plotPCAabl{128}{
    \thisrow{mean_log_volume}/\thisrow{volume_dimension}
}{ablBlue}

\plotPCAabl{192}{
    \thisrow{mean_log_volume}/\thisrow{volume_dimension}
}{ablOrange}

\plotPCAabl{256}{
    \thisrow{mean_log_volume}/\thisrow{volume_dimension}
}{ablGreen}

\plotPCAabl{384}{
    \thisrow{mean_log_volume}/\thisrow{volume_dimension}
}{ablRed}

\plotPCAabl{499}{
    \thisrow{mean_log_volume}/\thisrow{volume_dimension}
}{ablPurple}
\end{groupplot}

\node[
    anchor=south,
    inner sep=0pt
]
at
([yshift=6mm]
$(group c1r1.north west)!0.5!(group c4r1.north east)$)
{%
\begin{tabular}{
@{}
c@{\hspace{8pt}}
c@{\hspace{8pt}}
c@{\hspace{8pt}}
c@{\hspace{8pt}}
c
@{}
}

\tikz[baseline=-0.5ex]{
    \draw[ablBlue,line width=0.8pt] (0,0)--(0.35,0);
    \fill[ablBlue] (0.175,0) circle (1.1pt);
}
{\scriptsize $m=128$}
&
\tikz[baseline=-0.5ex]{
    \draw[ablOrange,line width=0.8pt] (0,0)--(0.35,0);
    \fill[ablOrange] (0.175,0) circle (1.1pt);
}
{\scriptsize $m=192$}
&
\tikz[baseline=-0.5ex]{
    \draw[ablGreen,line width=0.8pt] (0,0)--(0.35,0);
    \fill[ablGreen] (0.175,0) circle (1.1pt);
}
{\scriptsize $m=256$}
&
\tikz[baseline=-0.5ex]{
    \draw[ablRed,line width=0.8pt] (0,0)--(0.35,0);
    \fill[ablRed] (0.175,0) circle (1.1pt);
}
{\scriptsize $m=384$}
&
\tikz[baseline=-0.5ex]{
    \draw[ablPurple,line width=0.8pt] (0,0)--(0.35,0);
    \fill[ablPurple] (0.175,0) circle (1.1pt);
}
{\scriptsize $m=499$}

\end{tabular}
};

\end{tikzpicture}

\caption{
\textbf{Effect of retained PCA dimension $m$ on CelebA certification at $k=500$.}
Certified accuracy, abstention, mean manifold certified radius, and
dimension-normalized perturbation log-volume across noise scales.
The main experiments retain the full local PCA basis.
}
\label{fig:ablation_pca_dim}

\end{figure}
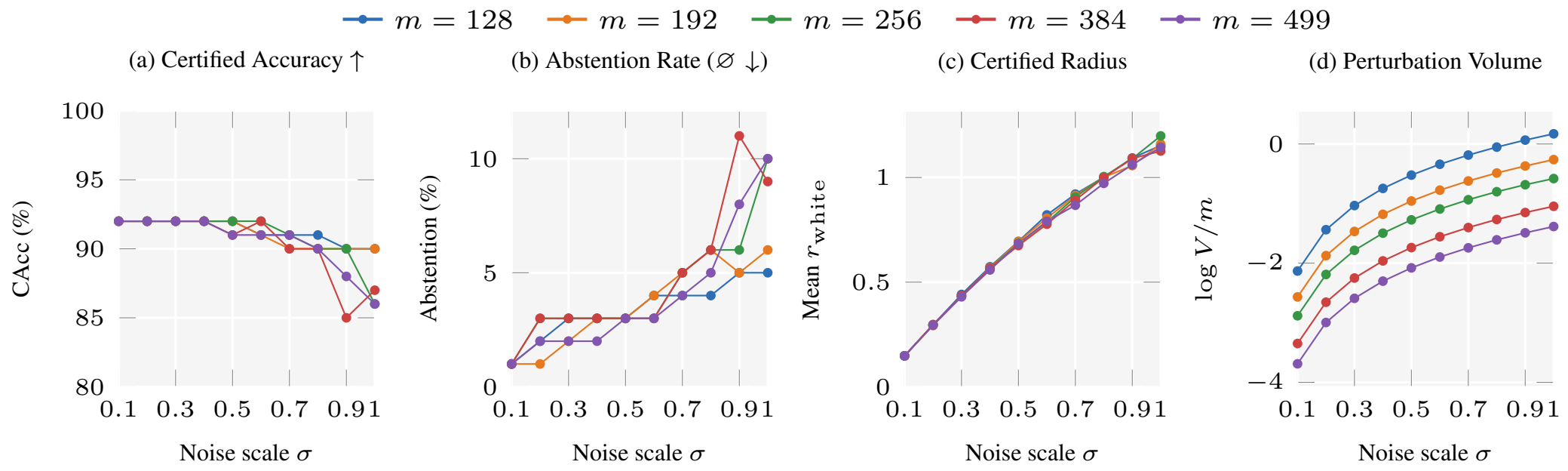

\paragraph{Discussion on perturbation volume.}
The effect of $m$ is more pronounced in the perturbation geometry.
Increasing $m$ systematically decreases the dimension-normalized
log-volume (Fig.~\ref{fig:ablation_pca_dim}), as progressively retained
low-variance directions reduce the geometric mean of the PCA eigenvalues.
In contrast, with $m=100$ fixed
(Fig.~\ref{fig:ablation_knn_k}(b)), volume varies only slightly across $k$.
Thus, the pronounced volume differences in the full-rank setting are
primarily associated with the changing retained dimension rather than
neighborhood size. 

\paragraph{Monte Carlo sample size $n$.}
Using $k=500$ and the full local PCA basis,
Fig.~\ref{fig:ablation_mc_samples} shows that increasing $n$ primarily
extends the attainable certificate through tighter probability estimation,
while the separation in certified accuracy between the two smoothing
schemes persists. To characterize this effect without averaging radii
across examples, Fig.~\ref{fig:mc_radius_ablation} reports certified
accuracy at a required radius $r$, following \citet{cohen2019certified}:
\begin{equation}
\label{app:approx_radius}
\mathrm{CAcc}_{\sigma}(r)
=
\frac{1}{N}\sum_{i=1}^{N}
\mathbf{1}\!\left[
\hat y_i=y_i,\;
\mathrm{non\text{-}abstain},\;
R_i\ge r
\right].
\end{equation}
Here, $R_i$ denotes the certificate in the corresponding geometry:
$R_{\mathrm I}$ for isotropic smoothing and $r_{\mathrm{white}}$ for
manifold smoothing. Larger $n$ extends the range over which certification
can be established. At larger noise scales, isotropic smoothing reaches
larger input-space radii while its certified accuracy approaches $45\%$;
manifold smoothing maintains substantially higher certified accuracy over
its corresponding whitened-space radius range.
%
%

\definecolor{ablBlue}{RGB}{45,110,180}
\definecolor{ablOrange}{RGB}{230,120,30}
\definecolor{ablGreen}{RGB}{55,150,70}
\definecolor{ablRed}{RGB}{205,65,65}
\definecolor{ablPurple}{RGB}{130,85,175}

\definecolor{maniBorder}{RGB}{80,170,100}


\newcommand{\mcIsoCSV}{
data/ablation/celeba/mc_isotropic_table.csv
}

\newcommand{\mcManiCSV}{
data/ablation/celeba/mc_manifold_table.csv
}

%

\newcommand{\plotMCiso}[3]{%
\addplot[
    color=#3,
    line width=0.3pt,
    mark=*,
    mark size=1pt,
    restrict expr to domain={\thisrow{parameter}}{#1:#1},
    unbounded coords=discard
]
table[
    col sep=comma,
    x=sigma,
    y expr={#2}
]{\mcIsoCSV};
}

\newcommand{\plotMCmani}[3]{%
\addplot[
    color=#3,
    line width=0.3pt,
    mark=*,
    mark size=1pt,
    restrict expr to domain={\thisrow{parameter}}{#1:#1},
    unbounded coords=discard
]
table[
    col sep=comma,
    x=sigma,
    y expr={#2}
]{\mcManiCSV};
}


\begin{figure}[H]
\centering

\begin{tikzpicture}

\begin{groupplot}[
    group style={
        group size=4 by 2,
        horizontal sep=1.2cm,
        vertical sep=1.15cm,
    },
     width=0.26\textwidth,
    height=4.0cm,
    xmin=0.1,
    xmax=1.0,
    xtick={0.1,0.3,0.5,0.7,0.9,1.0},
    xlabel={Noise scale $\sigma$},
    axis background/.style={
        fill=gray!8
    },
    grid=major,
    major grid style={
        draw=white,
        line width=0.6pt
    },
    tick label style={
        font=\tiny
    },
    label style={
        font=\tiny
    },
    title style={
        font=\tiny
    },
     axis line style={draw=white},
]



\nextgroupplot[
    title={(a) Certified Accuracy $\uparrow$},
    ylabel={CAcc (\%)},
    ymin=40,
    ymax=100,
    ytick={40,60,80,100},
]

\plotMCiso{100}{\thisrow{certified_accuracy_pct}}{ablBlue}
\plotMCiso{300}{\thisrow{certified_accuracy_pct}}{ablOrange}
\plotMCiso{500}{\thisrow{certified_accuracy_pct}}{ablGreen}
\plotMCiso{700}{\thisrow{certified_accuracy_pct}}{ablRed}
\plotMCiso{1000}{\thisrow{certified_accuracy_pct}}{ablPurple}


\nextgroupplot[
    title={(b) Abstention $\downarrow$},
    ylabel={Abstention (\%)},
    ymin=0,
]

\plotMCiso{100}{\thisrow{abstention_pct}}{ablBlue}
\plotMCiso{300}{\thisrow{abstention_pct}}{ablOrange}
\plotMCiso{500}{\thisrow{abstention_pct}}{ablGreen}
\plotMCiso{700}{\thisrow{abstention_pct}}{ablRed}
\plotMCiso{1000}{\thisrow{abstention_pct}}{ablPurple}


\nextgroupplot[
    title={(c) Certified Radius},
    ylabel={Mean $R_{\mathrm I}$},
    ymin=0,
    ymax=2.5,
    ytick={0,0.5,1.0,1.5,2.0,2.5},
]

\plotMCiso{100}{\thisrow{mean_radius}}{ablBlue}
\plotMCiso{300}{\thisrow{mean_radius}}{ablOrange}
\plotMCiso{500}{\thisrow{mean_radius}}{ablGreen}
\plotMCiso{700}{\thisrow{mean_radius}}{ablRed}
\plotMCiso{1000}{\thisrow{mean_radius}}{ablPurple}


\nextgroupplot[
    title={(d) Perturbation Volume},
    ylabel={$\log V_{\mathrm I}/d$},
    ymin=-8,
    ymax=0,
    ytick={-8,-6,-4,-2,-1},
]

\plotMCiso{100}{
    \thisrow{mean_log_volume}/
    \thisrow{volume_dimension}
}{ablBlue}

\plotMCiso{300}{
    \thisrow{mean_log_volume}/
    \thisrow{volume_dimension}
}{ablOrange}

\plotMCiso{500}{
    \thisrow{mean_log_volume}/
    \thisrow{volume_dimension}
}{ablGreen}

\plotMCiso{700}{
    \thisrow{mean_log_volume}/
    \thisrow{volume_dimension}
}{ablRed}

\plotMCiso{1000}{
    \thisrow{mean_log_volume}/
    \thisrow{volume_dimension}
}{ablPurple}



\nextgroupplot[
    ylabel={CAcc (\%)},
    ymin=40,
    ymax=100,
    ytick={40,60,80,100},
]

\plotMCmani{100}{\thisrow{certified_accuracy_pct}}{ablBlue}
\plotMCmani{300}{\thisrow{certified_accuracy_pct}}{ablOrange}
\plotMCmani{500}{\thisrow{certified_accuracy_pct}}{ablGreen}
\plotMCmani{700}{\thisrow{certified_accuracy_pct}}{ablRed}
\plotMCmani{1000}{\thisrow{certified_accuracy_pct}}{ablPurple}


\nextgroupplot[
    ylabel={Abstention (\%)},
    ymin=0,
]

\plotMCmani{100}{\thisrow{abstention_pct}}{ablBlue}
\plotMCmani{300}{\thisrow{abstention_pct}}{ablOrange}
\plotMCmani{500}{\thisrow{abstention_pct}}{ablGreen}
\plotMCmani{700}{\thisrow{abstention_pct}}{ablRed}
\plotMCmani{1000}{\thisrow{abstention_pct}}{ablPurple}


\nextgroupplot[
    ylabel={Mean $r_{\mathrm{white}}$},
    ymin=0,
    ymax=2.5,
    ytick={0,0.5,1.0,1.5,2.0,2.5},
]

\plotMCmani{100}{\thisrow{mean_radius}}{ablBlue}
\plotMCmani{300}{\thisrow{mean_radius}}{ablOrange}
\plotMCmani{500}{\thisrow{mean_radius}}{ablGreen}
\plotMCmani{700}{\thisrow{mean_radius}}{ablRed}
\plotMCmani{1000}{\thisrow{mean_radius}}{ablPurple}


\nextgroupplot[
    ylabel={$\log V_{\mathrm M}/m$},
    ymin=-8,
    ymax=0,
    ytick={-8,-6,-4,-2,-1},
]

\plotMCmani{100}{
    \thisrow{mean_log_volume}/
    \thisrow{volume_dimension}
}{ablBlue}

\plotMCmani{300}{
    \thisrow{mean_log_volume}/
    \thisrow{volume_dimension}
}{ablOrange}

\plotMCmani{500}{
    \thisrow{mean_log_volume}/
    \thisrow{volume_dimension}
}{ablGreen}

\plotMCmani{700}{
    \thisrow{mean_log_volume}/
    \thisrow{volume_dimension}
}{ablRed}

\plotMCmani{1000}{
    \thisrow{mean_log_volume}/
    \thisrow{volume_dimension}
}{ablPurple}

\end{groupplot}


\node[
    rotate=90,
    font=\scriptsize,
    anchor=south
]
at ([xshift=-10mm]
$(group c1r1.north west)!0.5!(group c1r1.south west)$)
{Isotropic};

\node[
    rotate=90,
    font=\scriptsize,
    anchor=south,
]
at ([xshift=-10mm]
$(group c1r2.north west)!0.5!(group c1r2.south west)$)
{\textbf{Manifold (Ours)}};


\draw[
    maniBorder,
    rounded corners=2pt,
    line width=0.8pt
]
([xshift=-6mm,yshift=3mm]group c1r2.north west)
rectangle
([xshift=3mm,yshift=-4mm]group c4r2.south east);


\node[
    anchor=south,
    inner sep=0pt
]
at
([yshift=6mm]
$(group c1r1.north west)!0.5!(group c4r1.north east)$)
{%
\begin{tabular}{
@{}
c@{\hspace{8pt}}
c@{\hspace{8pt}}
c@{\hspace{8pt}}
c@{\hspace{8pt}}
c
@{}
}

\tikz[baseline=-0.5ex]{
    \draw[ablBlue,line width=0.8pt] (0,0)--(0.35,0);
    \fill[ablBlue] (0.175,0) circle (1.1pt);
}
{\scriptsize $n=100$}

&

\tikz[baseline=-0.5ex]{
    \draw[ablOrange,line width=0.8pt] (0,0)--(0.35,0);
    \fill[ablOrange] (0.175,0) circle (1.1pt);
}
{\scriptsize $n=300$}

&

\tikz[baseline=-0.5ex]{
    \draw[ablGreen,line width=0.8pt] (0,0)--(0.35,0);
    \fill[ablGreen] (0.175,0) circle (1.1pt);
}
{\scriptsize $n=500$}

&

\tikz[baseline=-0.5ex]{
    \draw[ablRed,line width=0.8pt] (0,0)--(0.35,0);
    \fill[ablRed] (0.175,0) circle (1.1pt);
}
{\scriptsize $n=700$}

&

\tikz[baseline=-0.5ex]{
    \draw[ablPurple,line width=0.8pt] (0,0)--(0.35,0);
    \fill[ablPurple] (0.175,0) circle (1.1pt);
}
{\scriptsize $n=1000$}

\end{tabular}
};

\end{tikzpicture}


\caption{
\textbf{Effect of Monte Carlo sample size $n$ on CelebA certification.}
Top: isotropic smoothing; bottom: manifold smoothing.
We vary $n\in\{100,300,500,700,1000\}$ and report certified accuracy,
abstention, mean certified radius, and dimension-normalized perturbation
log-volume across representative noise scales.
The main experiments use $n=100$.
}

\label{fig:ablation_mc_samples}

\end{figure}


\definecolor{ablBlue}{RGB}{45,110,180}
\definecolor{ablOrange}{RGB}{230,120,30}
\definecolor{ablGreen}{RGB}{55,150,70}
\definecolor{ablRed}{RGB}{205,65,65}
\definecolor{ablPurple}{RGB}{130,85,175}



\newcommand{\mcIsoRadiusCSV}{
data/ablation/celeba/iso_certified_accuracy_radius_curves_mc.csv
}

\newcommand{\mcManiRadiusCSV}{
data/ablation/celeba/mani_certified_accuracy_radius_curves_mc.csv
}

%

\newcommand{\plotMCIsoRadius}[3]{%
    \addplot[
        color=#3,
        line width=0.4pt,
        const plot,
        no marks,
        restrict expr to domain={\thisrow{sigma}}{#1:#1},
        restrict expr to domain={\thisrow{parameter}}{#2:#2},
        unbounded coords=discard,
    ]
    table[
        col sep=comma,
        x=radius_threshold,
        y=certified_accuracy_pct,
    ]{\mcIsoRadiusCSV};
}

\newcommand{\plotMCManiRadius}[3]{%
    \addplot[
        color=#3,
        line width=0.4pt,
        const plot,
        no marks,
        restrict expr to domain={\thisrow{sigma}}{#1:#1},
        restrict expr to domain={\thisrow{parameter}}{#2:#2},
        unbounded coords=discard,
    ]
    table[
        col sep=comma,
        x=radius_threshold,
        y=certified_accuracy_pct,
    ]{\mcManiRadiusCSV};
}


\newcommand{\allMCIsoRadius}[1]{%
    \plotMCIsoRadius{#1}{100}{ablBlue}
    \plotMCIsoRadius{#1}{300}{ablOrange}
    \plotMCIsoRadius{#1}{500}{ablGreen}
    \plotMCIsoRadius{#1}{700}{ablRed}
    \plotMCIsoRadius{#1}{1000}{ablPurple}
}

\newcommand{\allMCManiRadius}[1]{%
    \plotMCManiRadius{#1}{100}{ablBlue}
    \plotMCManiRadius{#1}{300}{ablOrange}
    \plotMCManiRadius{#1}{500}{ablGreen}
    \plotMCManiRadius{#1}{700}{ablRed}
    \plotMCManiRadius{#1}{1000}{ablPurple}
}


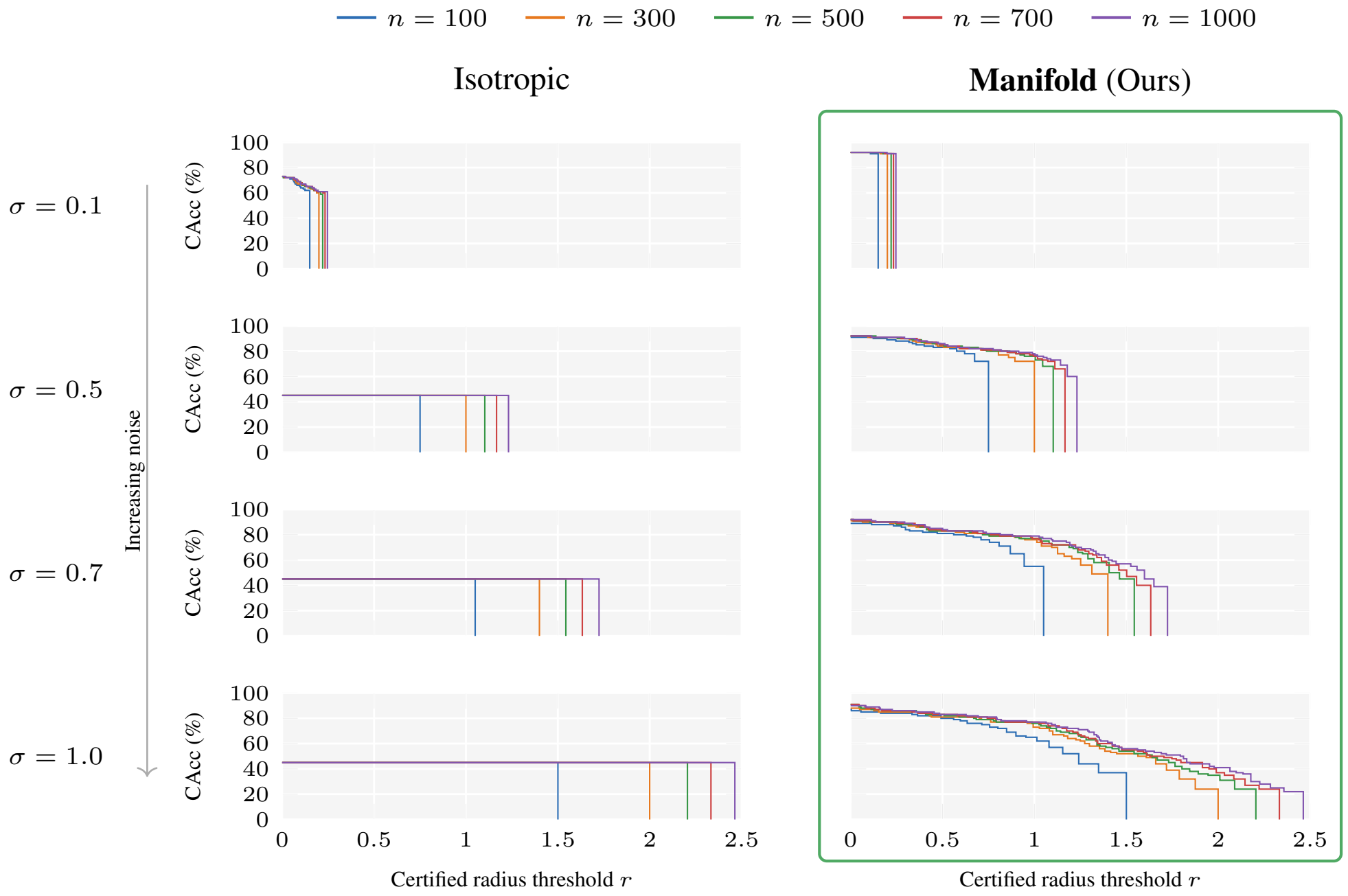
\begin{figure}[H]
\centering

\begin{tikzpicture}

\begin{groupplot}[
    group style={
        group size=2 by 4,
        horizontal sep=1.05cm,
        vertical sep=0.55cm,
    },
    width=0.43\textwidth,
    height=2.8cm,
    xmin=0,
    xmax=2.5,
    ymin=0,
    ymax=100,
    xtick={0,0.5,1.0,1.5,2.0,2.5},
    ytick={0,20,40,60,80,100},
    axis background/.style={
        fill=gray!8
    },
    grid=major,
    major grid style={
        draw=white,
        line width=0.55pt
    },
    axis line style={
        draw=white
    },
    tick style={
        draw=gray!15
    },
    tick label style={
        font=\tiny
    },
    label style={
        font=\tiny
    },
]


\nextgroupplot[
    ylabel={CAcc (\%)},
    xticklabels=\empty,
]

\allMCIsoRadius{0.1}

\nextgroupplot[
    yticklabels=\empty,
    xticklabels=\empty,
]

\allMCManiRadius{0.1}


\nextgroupplot[
    ylabel={CAcc (\%)},
    xticklabels=\empty,
]

\allMCIsoRadius{0.5}

\nextgroupplot[
    yticklabels=\empty,
    xticklabels=\empty,
]

\allMCManiRadius{0.5}


\nextgroupplot[
    ylabel={CAcc (\%)},
    xticklabels=\empty,
]

\allMCIsoRadius{0.7}

\nextgroupplot[
    yticklabels=\empty,
    xticklabels=\empty,
]

\allMCManiRadius{0.7}


\nextgroupplot[
    ylabel={CAcc (\%)},
    xlabel={Certified radius threshold $r$},
]

\allMCIsoRadius{1.0}

\nextgroupplot[
    yticklabels=\empty,
    xlabel={Certified radius threshold $r$},
]

\allMCManiRadius{1.0}

\end{groupplot}


\node[
    font=\small,
    anchor=south
]
at ([yshift=3.2mm]group c1r1.north)
{Isotropic};

\node[
    font=\small,
    anchor=south,
]
at ([yshift=3.2mm]group c2r1.north)
{\textbf{Manifold} (Ours)};


\node[
    font=\scriptsize,
    anchor=east
]
at ([xshift=-16mm]group c1r1.west)
{$\sigma=0.1$};

\node[
    font=\scriptsize,
    anchor=east
]
at ([xshift=-16mm]group c1r2.west)
{$\sigma=0.5$};

\node[
    font=\scriptsize,
    anchor=east
]
at ([xshift=-16mm]group c1r3.west)
{$\sigma=0.7$};

\node[
    font=\scriptsize,
    anchor=east
]
at ([xshift=-16mm]group c1r4.west)
{$\sigma=1.0$};


\draw[
    ->,
    gray!65,
    line width=0.5pt
]
([xshift=-13mm,yshift=2mm]group c1r1.west)
--
([xshift=-13mm,yshift=-2mm]group c1r4.west);

\node[
    rotate=90,
    font=\tiny,
    anchor=south
]
at
($(group c1r1.west)!0.5!(group c1r4.west)+(-1.2cm,0)$)
{Increasing noise};


\draw[
    maniBorder,
    rounded corners=2pt,
    line width=0.8pt
]
([xshift=-3mm,yshift=3mm]group c2r1.north west)
rectangle
([xshift=3mm,yshift=-4mm]group c2r4.south east);


\node[
    anchor=south,
    inner sep=0pt
]
at
([yshift=10mm]
$(group c1r1.north west)!0.5!(group c2r1.north east)$)
{%
\begin{tabular}{
@{}
c@{\hspace{10pt}}
c@{\hspace{10pt}}
c@{\hspace{10pt}}
c@{\hspace{10pt}}
c
@{}
}

\tikz[baseline=-0.5ex]{
    \draw[ablBlue,line width=0.8pt] (0,0)--(0.38,0);
}
{\scriptsize $n=100$}

&

\tikz[baseline=-0.5ex]{
    \draw[ablOrange,line width=0.8pt] (0,0)--(0.38,0);
}
{\scriptsize $n=300$}

&

\tikz[baseline=-0.5ex]{
    \draw[ablGreen,line width=0.8pt] (0,0)--(0.38,0);
}
{\scriptsize $n=500$}

&

\tikz[baseline=-0.5ex]{
    \draw[ablRed,line width=0.8pt] (0,0)--(0.38,0);
}
{\scriptsize $n=700$}

&

\tikz[baseline=-0.5ex]{
    \draw[ablPurple,line width=0.8pt] (0,0)--(0.38,0);
}
{\scriptsize $n=1000$}

\end{tabular}
};

\end{tikzpicture}


\caption{
\textbf{Effect of Monte Carlo sample size on certified accuracy across radii.}
Certified accuracy as a function of the radius threshold for isotropic
(left) and manifold smoothing (right) at increasing noise scales.
Each curve corresponds to a different certification sample count
$n\in\{100,300,500,700,1000\}$.
}

\label{fig:mc_radius_ablation}

\end{figure}
\paragraph{Evaluation-set size $N$.}
Figure~\ref{fig:ablation_test_size} varies the number of evaluated examples
from $N=100$ to $1000$. Manifold smoothing exhibits similar certified
accuracy across evaluation-set sizes, whereas isotropic CAcc is slightly
sensitive to $N$, particularly at larger noise scales. The qualitative
separation between the two smoothing distributions therefore persists as
the evaluation set is enlarged.

%
%

\definecolor{ablBlue}{RGB}{45,110,180}
\definecolor{ablOrange}{RGB}{230,120,30}
\definecolor{ablGreen}{RGB}{55,150,70}
\definecolor{ablRed}{RGB}{205,65,65}
\definecolor{ablPurple}{RGB}{130,85,175}

\definecolor{maniBorder}{RGB}{80,170,100}


\newcommand{\testSIsoCSV}{
data/ablation/celeba/test_size_isotropic_table.csv
}

\newcommand{\testSManiCSV}{
data/ablation/celeba/test_size_manifold_table.csv
}

%

\newcommand{\plotTSiso}[3]{%
\addplot[
    color=#3,
    line width=0.3pt,
    mark=*,
    mark size=1pt,
    restrict expr to domain={\thisrow{parameter}}{#1:#1},
    unbounded coords=discard
]
table[
    col sep=comma,
    x=sigma,
    y expr={#2}
]{\testSIsoCSV};
}

\newcommand{\plotTSmani}[3]{%
\addplot[
    color=#3,
    line width=0.3pt,
    mark=*,
    mark size=1pt,
    restrict expr to domain={\thisrow{parameter}}{#1:#1},
    unbounded coords=discard
]
table[
    col sep=comma,
    x=sigma,
    y expr={#2}
]{\testSManiCSV};
}


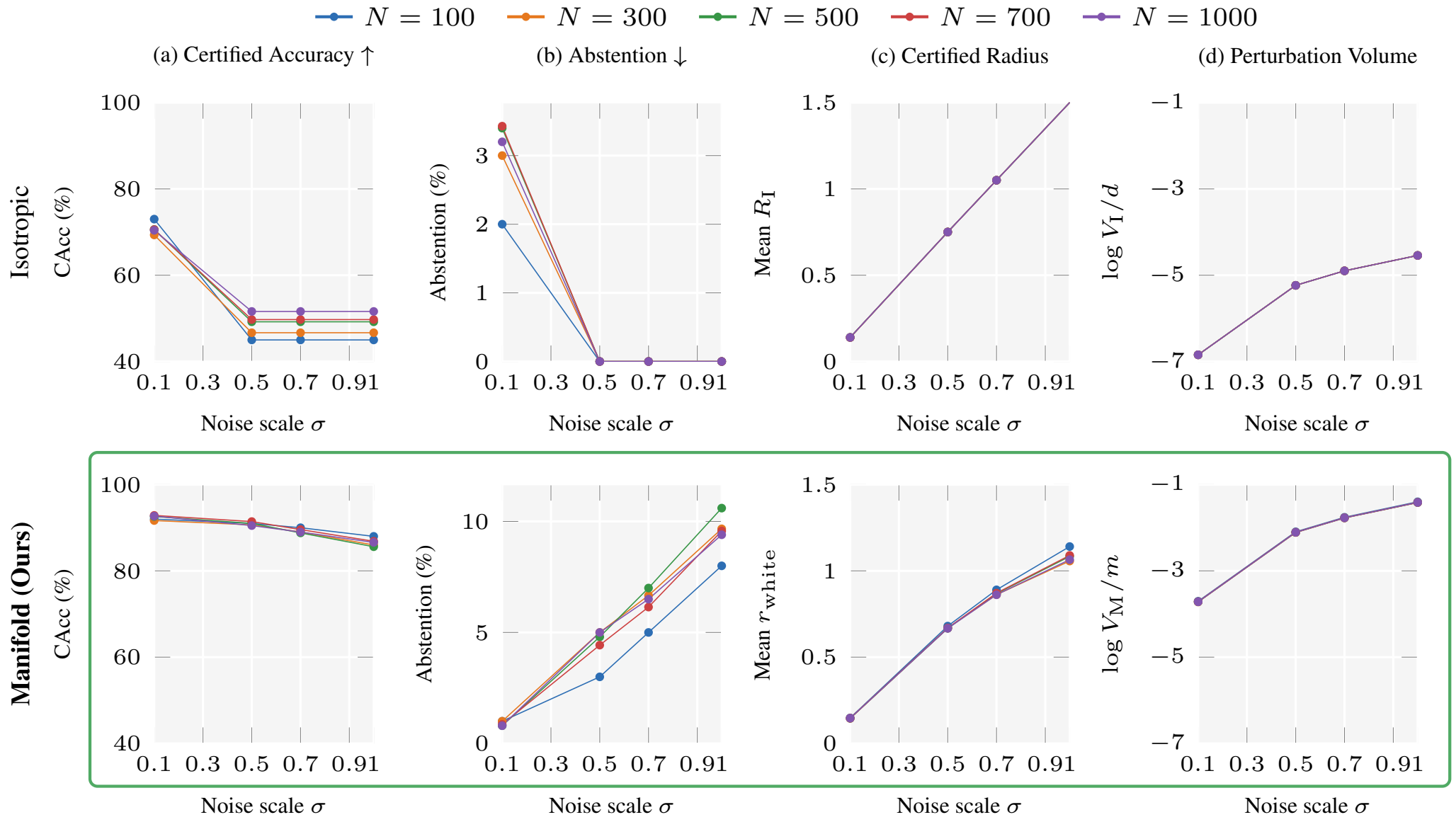
\begin{figure}[H]
\centering

\begin{tikzpicture}

\begin{groupplot}[
    group style={
        group size=4 by 2,
        horizontal sep=1.2cm,
        vertical sep=1.15cm,
    },
     width=0.26\textwidth,
    height=4.0cm,
    xmin=0.1,
    xmax=1.0,
    xtick={0.1,0.3,0.5,0.7,0.9,1.0},
    xlabel={Noise scale $\sigma$},
    axis background/.style={
        fill=gray!8
    },
    grid=major,
    major grid style={
        draw=white,
        line width=0.6pt
    },
    tick label style={
        font=\tiny
    },
    label style={
        font=\tiny
    },
    title style={
        font=\tiny
    },
     axis line style={draw=white},
]



\nextgroupplot[
    title={(a) Certified Accuracy $\uparrow$},
    ylabel={CAcc (\%)},
    ymin=40,
    ymax=100,
    ytick={40,60,80,100},
]

\plotTSiso{100}{\thisrow{certified_accuracy_pct}}{ablBlue}
\plotTSiso{300}{\thisrow{certified_accuracy_pct}}{ablOrange}
\plotTSiso{500}{\thisrow{certified_accuracy_pct}}{ablGreen}
\plotTSiso{700}{\thisrow{certified_accuracy_pct}}{ablRed}
\plotTSiso{1000}{\thisrow{certified_accuracy_pct}}{ablPurple}


\nextgroupplot[
    title={(b) Abstention $\downarrow$},
    ylabel={Abstention (\%)},
    ymin=0,
]

\plotTSiso{100}{\thisrow{abstention_pct}}{ablBlue}
\plotTSiso{300}{\thisrow{abstention_pct}}{ablOrange}
\plotTSiso{500}{\thisrow{abstention_pct}}{ablGreen}
\plotTSiso{700}{\thisrow{abstention_pct}}{ablRed}
\plotTSiso{1000}{\thisrow{abstention_pct}}{ablPurple}


\nextgroupplot[
    title={(c) Certified Radius},
    ylabel={Mean $R_{\mathrm I}$},
     ymin=0,
    ymax=1.5,
    ytick={0,0.5,1,1.5},
]

\plotTSiso{100}{\thisrow{mean_radius}}{ablBlue}
\plotTSiso{300}{\thisrow{mean_radius}}{ablOrange}
\plotTSiso{500}{\thisrow{mean_radius}}{ablGreen}
\plotTSiso{700}{\thisrow{mean_radius}}{ablRed}
\plotTSiso{1000}{\thisrow{mean_radius}}{ablPurple}


\nextgroupplot[
    title={(d) Perturbation Volume},
    ylabel={$\log V_{\mathrm I}/d$},
    ymin=-7,
    ymax=-1,
    ytick={-7,-5,-3,-1}
]

\plotTSiso{100}{
    \thisrow{mean_log_volume}/
    \thisrow{volume_dimension}
}{ablBlue}

\plotTSiso{300}{
    \thisrow{mean_log_volume}/
    \thisrow{volume_dimension}
}{ablOrange}

\plotTSiso{500}{
    \thisrow{mean_log_volume}/
    \thisrow{volume_dimension}
}{ablGreen}

\plotTSiso{700}{
    \thisrow{mean_log_volume}/
    \thisrow{volume_dimension}
}{ablRed}

\plotTSiso{1000}{
    \thisrow{mean_log_volume}/
    \thisrow{volume_dimension}
}{ablPurple}



\nextgroupplot[
    ylabel={CAcc (\%)},
    ymin=40,
    ymax=100,
    ytick={40,60,80,100},
]

\plotTSmani{100}{\thisrow{certified_accuracy_pct}}{ablBlue}
\plotTSmani{300}{\thisrow{certified_accuracy_pct}}{ablOrange}
\plotTSmani{500}{\thisrow{certified_accuracy_pct}}{ablGreen}
\plotTSmani{700}{\thisrow{certified_accuracy_pct}}{ablRed}
\plotTSmani{1000}{\thisrow{certified_accuracy_pct}}{ablPurple}


\nextgroupplot[
    ylabel={Abstention (\%)},
    ymin=0,
]

\plotTSmani{100}{\thisrow{abstention_pct}}{ablBlue}
\plotTSmani{300}{\thisrow{abstention_pct}}{ablOrange}
\plotTSmani{500}{\thisrow{abstention_pct}}{ablGreen}
\plotTSmani{700}{\thisrow{abstention_pct}}{ablRed}
\plotTSmani{1000}{\thisrow{abstention_pct}}{ablPurple}


\nextgroupplot[
    ylabel={Mean $r_{\mathrm{white}}$},
    ymin=0,
    ymax=1.5,
    ytick={0,0.5,1,1.5},
]

\plotTSmani{100}{\thisrow{mean_radius}}{ablBlue}
\plotTSmani{300}{\thisrow{mean_radius}}{ablOrange}
\plotTSmani{500}{\thisrow{mean_radius}}{ablGreen}
\plotTSmani{700}{\thisrow{mean_radius}}{ablRed}
\plotTSmani{1000}{\thisrow{mean_radius}}{ablPurple}


\nextgroupplot[
    ylabel={$\log V_{\mathrm M}/m$},
    ymin=-7,
    ymax=-1,
    ytick={-7,-5,-3,-1},
]

\plotTSmani{100}{
    \thisrow{mean_log_volume}/
    \thisrow{volume_dimension}
}{ablBlue}

\plotTSmani{300}{
    \thisrow{mean_log_volume}/
    \thisrow{volume_dimension}
}{ablOrange}

\plotTSmani{500}{
    \thisrow{mean_log_volume}/
    \thisrow{volume_dimension}
}{ablGreen}

\plotTSmani{700}{
    \thisrow{mean_log_volume}/
    \thisrow{volume_dimension}
}{ablRed}

\plotTSmani{1000}{
    \thisrow{mean_log_volume}/
    \thisrow{volume_dimension}
}{ablPurple}

\end{groupplot}


\node[
    rotate=90,
    font=\scriptsize,
    anchor=south
]
at ([xshift=-10mm]
$(group c1r1.north west)!0.5!(group c1r1.south west)$)
{Isotropic};

\node[
    rotate=90,
    font=\scriptsize,
    anchor=south,
]
at ([xshift=-10mm]
$(group c1r2.north west)!0.5!(group c1r2.south west)$)
{\textbf{Manifold (Ours)}};


\draw[
    maniBorder,
    rounded corners=2pt,
    line width=0.8pt
]
([xshift=-6mm,yshift=3mm]group c1r2.north west)
rectangle
([xshift=3mm,yshift=-4mm]group c4r2.south east);


\node[
    anchor=south,
    inner sep=0pt
]
at
([yshift=6mm]
$(group c1r1.north west)!0.5!(group c4r1.north east)$)
{%
\begin{tabular}{
@{}
c@{\hspace{8pt}}
c@{\hspace{8pt}}
c@{\hspace{8pt}}
c@{\hspace{8pt}}
c
@{}
}

\tikz[baseline=-0.5ex]{
    \draw[ablBlue,line width=0.8pt] (0,0)--(0.35,0);
    \fill[ablBlue] (0.175,0) circle (1.1pt);
}
{\scriptsize $N=100$}

&

\tikz[baseline=-0.5ex]{
    \draw[ablOrange,line width=0.8pt] (0,0)--(0.35,0);
    \fill[ablOrange] (0.175,0) circle (1.1pt);
}
{\scriptsize $N=300$}

&

\tikz[baseline=-0.5ex]{
    \draw[ablGreen,line width=0.8pt] (0,0)--(0.35,0);
    \fill[ablGreen] (0.175,0) circle (1.1pt);
}
{\scriptsize $N=500$}

&

\tikz[baseline=-0.5ex]{
    \draw[ablRed,line width=0.8pt] (0,0)--(0.35,0);
    \fill[ablRed] (0.175,0) circle (1.1pt);
}
{\scriptsize $N=700$}

&

\tikz[baseline=-0.5ex]{
    \draw[ablPurple,line width=0.8pt] (0,0)--(0.35,0);
    \fill[ablPurple] (0.175,0) circle (1.1pt);
}
{\scriptsize $N=1000$}

\end{tabular}
};

\end{tikzpicture}


\caption{
\textbf{Effect of test sample size $N$ on CelebA certification.}
Top: isotropic smoothing; bottom: manifold smoothing.
We vary $N\in\{100,300,500,700,1000\}$ and report certified accuracy,
abstention, mean certified radius, and dimension-normalized perturbation
log-volume across representative noise scales.
The main experiments use $N=100$.
}

\label{fig:ablation_test_size}

\end{figure}

\subsection{Comparable Noise Magnitude Baselines}
\label{app:basline}

Isotropic and manifold smoothing distribute Gaussian
noise differently. Consequently, using the same
nominal noise scale $\sigma$ does not imply equal
perturbation magnitudes or directly comparable
certified regions. To investigate how much of the
observed performance difference is explained by
noise magnitude rather than local covariance geometry,
we introduce two matched-magnitude baselines against manifold smoothing and
compare four smoothing schemes.

Let $U_{x,m}$ denote the retained PCA directions,
$\widetilde{\Lambda}_{x,m}$ the corresponding
eigenvalues after numerical flooring, and
$s_x=\operatorname{tr}(\widetilde{\Lambda}_{x,m})$.

\paragraph{Standard isotropic smoothing.}
The original isotropic baseline samples Gaussian
noise uniformly across all $d$ ambient dimensions:
\begin{equation}
    \delta_{\mathrm{Iso}}
    =\sigma\epsilon_d,
    \qquad
    \epsilon_d\sim\mathcal{N}(0,I_d).
\end{equation}
Its expected squared perturbation magnitude is
$\mathbb{E}\|\delta_{\mathrm{Iso}}\|_2^2
=\sigma^2d$.

\paragraph{Matched ambient isotropic smoothing.}
To control for the difference in overall noise
magnitude, we adjust the isotropic noise scale
so that its expected squared Euclidean norm matches
that of manifold smoothing:
\begin{equation}
    \delta_{\mathrm{Iso,norm}}
    =\sigma_{\mathrm Iso}\epsilon_d,
    \qquad
    \sigma_{\mathrm Iso}
    =\sigma\sqrt{\frac{s_x}{d}}.
\end{equation}
This baseline retains isotropic noise across all
ambient dimensions while matching manifold smoothing's expected
squared perturbation norm. It therefore helps
distinguish the effect of overall noise magnitude
from the effect of restricting noise to local
data directions.

\paragraph{Matched projected isotropic smoothing.}
To additionally control for the retained PCA
subspace, we sample isotropic Gaussian noise
within the same $m$-dimensional subspace as manifold smoothing:
\begin{equation}
    \delta_{\mathrm{ProjIso}}
    =\alpha U_{x,m}\epsilon_m,
    \qquad
    \alpha=\sigma\sqrt{\frac{s_x}{m}},
    \qquad
    \epsilon_m\sim\mathcal{N}(0,I_m).
\end{equation}
This baseline matches both the expected squared
perturbation norm and the retained PCA subspace,
but assigns equal variance to all retained
directions. Comparing it with manifold smoothing isolates the
additional contribution of eigenvalue-dependent
anisotropic scaling.

\paragraph{Randomized manifold smoothing (ours).}
Manifold smoothing samples Gaussian noise within the retained
PCA subspace, scaling each principal direction
according to its local covariance eigenvalue:
\begin{equation}
    \delta_{\mathrm{M}}
    =\sigma U_{x,m}
    \widetilde{\Lambda}_{x,m}^{1/2}\epsilon_m.
\end{equation}
Its expected squared perturbation norm is
\begin{equation}
    \mathbb{E}\|\delta_{\mathrm{M}}\|_2^2
    =\sigma^2s_x.
\end{equation}
Both matched baselines use this manifold smoothing magnitude
as their reference, satisfying
\begin{equation}
\begin{aligned}
    \mathbb{E}\|\delta_{\mathrm{Iso,norm}}\|_2^2
    &=\mathbb{E}\|\delta_{\mathrm{ProjIso}}\|_2^2\\
    &=\mathbb{E}\|\delta_{\mathrm{M}}\|_2^2
    =\sigma^2s_x.
\end{aligned}
\end{equation}
For each reference input, the PCA geometry and
the corresponding noise scales are held fixed
during certification.

\begin{table}[H]
\centering
\scriptsize
\setlength{\tabcolsep}{4pt}
\begin{tabular}{ccc|cccc}
\toprule
& \multicolumn{2}{c|}{(a) Perturbation magnitude}
& \multicolumn{4}{c}{(b) Cert Acc. (\%) $\uparrow$} \\
\cmidrule(lr){2-3}\cmidrule(lr){4-7}
$\sigma$
& Isotropic ($\sigma\sqrt{d}$)
& Manifold ($\sigma\sqrt{s_x}$)
& Isotropic
& \textbf{Matched ambient ISO}
& \textbf{Matched projected ISO}
& \textbf{Ours} \\
\midrule

0.02 & $7.760 \pm 0.000$ & $1.210 \pm 0.151$ & 93.0 & 92.0 & 92.0 & 92.0 \\
0.05 & $19.399 \pm 0.000$ & $3.025 \pm 0.377$ & 87.0 & 92.0 & 92.0 & \underline{92.0} \\
0.10 & $38.798 \pm 0.000$ & $6.050 \pm 0.753$ & 72.0 & 92.0 & 92.0 & \underline{92.0} \\
0.20 & $77.596 \pm 0.000$ & $12.100 \pm 1.506$ & 45.0 & 90.0 & 92.0 & \underline{92.0} \\
0.30 & $116.394 \pm 0.000$ & $18.150 \pm 2.259$ & 45.0 & 88.0 & 92.0 & \underline{92.0} \\
0.40 & $155.192 \pm 0.000$ & $24.200 \pm 3.012$ & 45.0 & 85.0 & 90.0 & \textcolor{gaincolor}{\textbf{91.0}} \\
0.50 & $193.990 \pm 0.000$ & $30.250 \pm 3.765$ & 45.0 & 80.0 & 89.0 & \textcolor{gaincolor}{\textbf{91.0}} \\
0.60 & $232.788 \pm 0.000$ & $36.300 \pm 4.518$ & 45.0 & 74.0 & 88.0 & \textcolor{gaincolor}{\textbf{91.0}} \\
0.70 & $271.586 \pm 0.000$ & $42.349 \pm 5.271$ & 45.0 & 68.0 & 87.0 & \textcolor{gaincolor}{\textbf{90.0}} \\
0.80 & $310.384 \pm 0.000$ & $48.399 \pm 6.024$ & 45.0 & 60.0 & 87.0 & \textcolor{gaincolor}{\textbf{88.0}} \\
0.90 & $349.181 \pm 0.000$ & $54.449 \pm 6.778$ & 45.0 & 51.0 & 84.0 & \textcolor{gaincolor}{\textbf{89.0}} \\
1.00 & $387.979 \pm 0.000$ & $60.499 \pm 7.531$ & 45.0 & 47.0 & 82.0 & \textcolor{gaincolor}{\textbf{86.0}} \\
2.00 & $775.959 \pm 0.000$ & $120.998 \pm 15.061$ & 45.0 & 45.0 & 69.0 & \textcolor{gaincolor}{\textbf{72.0}} \\

\bottomrule
\end{tabular}
\caption{
\textbf{Norm Matched Smoothing Baseline:} Comparison of perturbation magnitudes and certified
accuracy across four smoothing schemes at different
nominal noise scales $\sigma$.
(a) Perturbation magnitude is measured by
$\rho_x=\sqrt{\mathbb{E}_{\delta\mid x}
\|\delta\|_2^2}$.
The columns report the magnitude of standard
isotropic smoothing ($\sigma\sqrt{d}$) and manifold smoothing
($\sigma\sqrt{s_x}$), respectively, as mean
$\pm$ standard deviation across reference inputs,
where $s_x=\operatorname{tr}
(\widetilde{\Lambda}_{x,m})$.
The manifold smoothing magnitude serves as the per-input matching
target for both new baselines.
(b) Certified accuracy (\%) for standard isotropic
smoothing, matched ambient isotropic smoothing
(same expected squared noise norm as manifold smoothing across
all $d$ dimensions), matched projected isotropic
smoothing (same norm and retained PCA subspace
as manifold smoothing, but uniform variance), and manifold smoothing (local
eigenvalue-dependent variance).
}
\label{tab:matched_noise_baselines}
\end{table}

Table~\ref{tab:matched_noise_baselines}
reports (a) the perturbation magnitudes of
standard isotropic smoothing and manifold smoothing, and
(b) the certified accuracy of all four methods.
The matched baselines use the manifold smoothing perturbation
magnitude reported in part (a) as their target,
with the matching performed separately for
each reference input.

The results show that differences in perturbation
magnitude explain part of the performance gap
between standard isotropic smoothing and manifold smoothing.
However, manifold smoothing continues to outperform matched
ambient isotropic smoothing, particularly at
larger noise scales. For example, at $\sigma=1$,
manifold smoothing achieves 86\% certified accuracy compared
with 47\% for matched ambient isotropic smoothing.

Restricting noise to the same PCA subspace
substantially narrows this gap, and matched projected
isotropic smoothing is achieving 82\% at $\sigma=1$.
Nevertheless, manifold smoothing achieves consistently higher
certified accuracy at larger noise scales,
indicating an additional benefit from scaling
perturbations according to local eigenvalues.

These results suggest that manifold smoothing benefits from
both subspace restriction and eigenvalue-dependent
anisotropic scaling, beyond differences in
perturbation magnitude alone.


\subsection{Computational Cost}
\label{app:runtime}
Manifold smoothing requires retrieving a local neighborhood and estimating
its PCA geometry once per input, which is then reused across all Monte Carlo
samples. Table~\ref{tab:production_runtime_grouped} reports the runtime for certifying a single
CelebA image with $n_0=30$ and $n=100$ in seconds(s). \textbf{ANN} is the search for the
$k$ nearest-neighbor IDs, \textbf{Fetch} loads their vectors into memory,
\textbf{PCA} estimates the local geometry, and \textbf{MC} includes noise
generation, model inference, and certificate computation.
\textbf{Certification} measures the complete certification path from image
loading to the final certificate, including neighborhood construction, PCA,
and MC certification. One-time model and ANN-index initialization is excluded
because these are loaded once and reused across inputs. 

Annoy neighborhood search runs on the CPU, while Monte Carlo model inference
uses an NVIDIA Quadro RTX 8000 GPU (48 GB). We additionally compare
scikit-learn, NumPy, and PyTorch PCA implementations to characterize the
implementation-dependent cost of local geometry estimation. In our experiments, we used scikit-learn.
Tab.~\ref{tab:production_runtime_grouped} shows an increase in runtime as $k$ increases, as well as variability with implementation source. However, local PCA remains relatively inexpensive.
\begin{table*}[t]
\centering
\small
\setlength{\tabcolsep}{3.5pt}
\caption{\textbf{Runtime breakdown for certifying one CelebA image}, in seconds. MC certification includes noise generation, preprocessing, classifier inference, and certificate calculation. Certification time includes image loading and neighborhood construction, but excludes model and index initialization. }
\label{tab:production_runtime_grouped}
\begin{tabular}{llrr|rrrrr}
\toprule
\textbf{PCA} & \textbf{Sweep} & $k$ & $m$ & \textbf{ANN (s)} & \textbf{Fetch (s)} & \textbf{PCA (s)} & \textbf{MC (s)} & \textbf{Certification (s)} \\
\midrule
\multirow{11}{*}{\rotatebox{90}{scikit-learn }} & \multirow{6}{*}{$k$} & 128 & 100 & 214.537 & 1.047 & 3.227 & 7.070 & 226.078 \\
 &  & 256 & 100 & 214.537 & 2.093 & 2.370 & 6.685 & 226.071 \\
 &  & 500 & 100 & 213.983 & 4.365 & 3.220 & 7.260 & 229.575 \\
 &  & 750 & 100 & 213.983 & 6.527 & 5.976 & 4.857 & 232.483 \\
 &  & 1000 & 100 & 6.060 & 8.521 & 3.678 & 3.741 & 23.608 \\
 &  & 5000 & 100 & 13.291 & 43.734 & 13.163 & 3.770 & 81.431 \\
\cmidrule{2-9}
 & \multirow{5}{*}{$m$} & 500 & 128 & 217.772 & 4.424 & 4.340 & 7.982 & 235.280 \\
 &  & 500 & 192 & 217.773 & 4.373 & 6.219 & 6.513 & 235.647 \\
 &  & 500 & 256 & 5.490 & 4.373 & 5.671 & 4.487 & 20.821 \\
 &  & 500 & 384 & 5.478 & 4.486 & 8.697 & 6.014 & 25.564 \\
 &  & 500 & 499 & 5.431 & 4.425 & 7.854 & 7.543 & 26.074 \\
\midrule
\multirow{11}{*}{\rotatebox{90}{NumPy}} & \multirow{6}{*}{$k$} & 128 & 100 & 6.121 & 1.033 & 6.526 & 3.796 & 17.667 \\
 &  & 256 & 100 & 6.117 & 2.103 & 9.057 & 4.629 & 22.307 \\
 &  & 500 & 100 & 240.701 & 4.411 & 14.076 & 4.938 & 264.895 \\
 &  & 750 & 100 & 240.701 & 6.485 & 22.147 & 4.725 & 275.198 \\
 &  & 1000 & 100 & 6.266 & 8.787 & 99.628 & 3.925 & 120.206 \\
 &  & 5000 & 100 & 6.179 & 44.825 & 437.246 & 4.226 & 499.792 \\
\cmidrule{2-9}
 & \multirow{5}{*}{$m$} & 500 & 128 & 5.986 & 4.407 & 13.274 & 4.755 & 29.228 \\
 &  & 500 & 192 & 5.892 & 4.882 & 18.152 & 7.504 & 37.337 \\
 &  & 500 & 256 & 5.379 & 4.491 & 14.734 & 4.518 & 29.918 \\
 &  & 500 & 384 & 6.006 & 4.394 & 13.617 & 6.053 & 30.880 \\
 &  & 500 & 499 & 5.557 & 4.421 & 15.300 & 6.957 & 33.007 \\
\midrule
\multirow{11}{*}{\rotatebox{90}{PyTorch}} & \multirow{6}{*}{$k$} & 128 & 100 & 6.986 & 1.159 & 4.442 & 5.042 & 17.911 \\
 &  & 256 & 100 & 5.491 & 2.108 & 12.149 & 8.663 & 28.841 \\
 &  & 500 & 100 & 7.672 & 4.443 & 6.432 & 4.673 & 24.004 \\
 &  & 750 & 100 & 7.677 & 6.594 & 8.778 & 5.103 & 29.327 \\
 &  & 1000 & 100 & 6.100 & 8.626 & 11.023 & 3.677 & 31.019 \\
 &  & 5000 & 100 & 6.033 & 47.171 & 155.247 & 6.175 & 222.342 \\
\cmidrule{2-9}
 & \multirow{5}{*}{$m$} & 500 & 128 & 6.103 & 4.418 & 6.212 & 4.645 & 22.155 \\
 &  & 500 & 192 & 8.087 & 4.315 & 6.480 & 5.400 & 25.077 \\
 &  & 500 & 256 & 8.087 & 4.439 & 6.455 & 4.940 & 24.736 \\
 &  & 500 & 384 & 6.112 & 4.415 & 6.445 & 7.084 & 24.839 \\
 &  & 500 & 499 & 6.110 & 4.403 & 6.394 & 7.277 & 24.943 \\
\bottomrule
\end{tabular}
\end{table*}

\section{Extended Results Analysis}
\label{app:quantitative}

\subsection{Robustness Across Classes (Binary Classification Case)}
\label{app:class_wise_celeba}

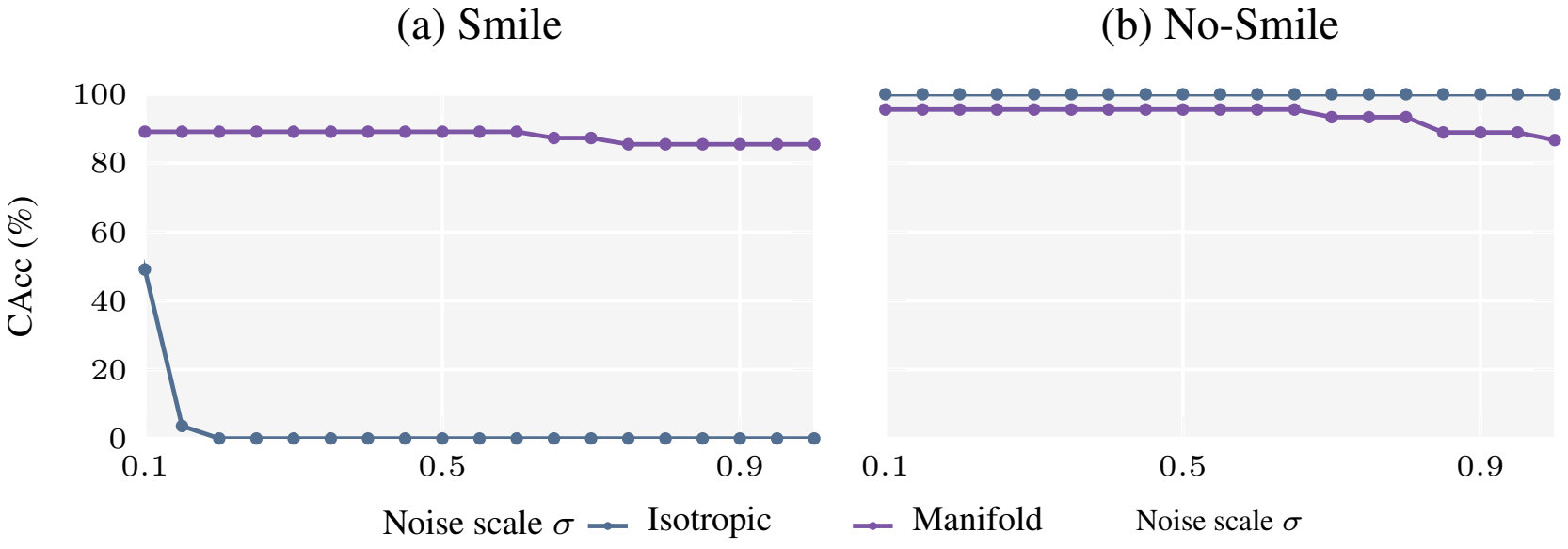
\begin{figure}[H]
    \centering

\newcommand{\classPlot}[3]{%
    \addplot[
        line width=0.9pt,
        mark=*,
        mark size=0.8pt,
        #3,
        restrict expr to domain={\thisrow{SmoothID}}{#1:#1},
        unbounded coords=discard,
    ]
    table[
        col sep=comma,
        x=sigma,
        y expr={100*\thisrow{#2}},
    ] {data/celeba_perclass_cert.csv};
}


\begin{tikzpicture}

\begin{axis}[
    name=smile,
    width=0.45\linewidth,
    height=4cm,
    xmin=0.1,
    xmax=1.0,
    ymin=0,
    ymax=100,
    xtick={0.1,0.5,0.9},
    ytick={0,20,40,60,80,100},
    xlabel={Noise scale $\sigma$},
    ylabel={CAcc (\%)},
    title={(a) Smile},
    axis background/.style={fill=gray!8},
    grid=major,
    major grid style={draw=white,line width=0.7pt},
    axis line style={draw=white},
    tick style={draw=gray!10},
    tick label style={font=\tiny},
    label style={font=\scriptsize},
    title style={font=\footnotesize},
]

\classPlot{0}{smile_acc}{isoBlue}
\classPlot{1}{smile_acc}{maniPurple}

\end{axis}

\begin{axis}[
    at={(smile.east)},
    anchor=west,
    xshift=0.5cm,
    width=0.45\linewidth,
    height=4cm,
    xmin=0.1,
    xmax=1.0,
    ymin=0,
    ymax=100,
    xtick={0.1,0.5,0.9},
    ytick={0,20,40,60,80,100},
    xlabel={Noise scale $\sigma$},
    title={(b) No-Smile},
    axis background/.style={fill=gray!8},
    grid=major,
    major grid style={draw=white,line width=0.7pt},
    axis line style={draw=white},
    tick style={draw=gray!10},
    tick label style={font=\tiny},
    label style={font=\tiny},
    title style={font=\footnotesize},
    yticklabels=\empty,
]

\classPlot{0}{no_smile_acc}{isoBlue}
\classPlot{1}{no_smile_acc}{maniPurple}

\end{axis}

\node[
    anchor=south,
    font=\scriptsize
] at ([yshift=-0.82cm]smile.south east)
{
    \tikz{
        \draw[isoBlue,line width=0.9pt] (0,0)--(0.28,0);
        \fill[isoBlue] (0.14,0) circle (0.8pt);
    }
    \ Isotropic
    \qquad
    \tikz{
        \draw[maniPurple,line width=0.9pt] (0,0)--(0.28,0);
        \fill[maniPurple] (0.14,0) circle (0.8pt);
    }
    \ Manifold
};

\end{tikzpicture}
   \caption{
\textbf{Class-wise certification on CelebA.}
Certified accuracy for Smile and No-Smile under isotropic and manifold
smoothing across increasing perturbation levels.
}
    \label{fig:celeba_perclass}
\end{figure}

Figure~\ref{fig:celeba_perclass} reveals a failure mode that is hidden by
aggregate certified accuracy in the binary prediction task: isotropic
smoothing increasingly certifies No-Smile while losing certification for
Smile as the perturbation scale grows. Consequently, the apparent $45\%$
CAcc at large $\sigma$ does not represent uniform robustness, but largely
reflects guarantees for only one side of the decision. This is problematic
when small, plausible variations in input can alter the
predicted class. Manifold smoothing instead maintains useful certification
for both outcomes over larger perturbation scales, yielding a more balanced
robustness guarantee for the predictive task.




\subsection{Extended Certification Results Across Noise Scales}

We report the complete certification results across all evaluated noise
scales $\sigma$, extending the representative settings reported in the
main paper. Results are averaged over five random seeds, where randomness
arises solely from the Monte Carlo sampling used during certification.
Model, evaluation data sample, and certification set remain fixed across
seeds. The tables \ref{tab:full_celeba} - \ref{tab:full_conceptspace} report the mean and standard deviation across these
certification runs.
\label{app:noise_sweeps_all}


\definecolor{gaincolor}{RGB}{55,160,75}
\definecolor{losscolor}{RGB}{130,130,130}
\definecolor{groupgray}{RGB}{242,242,242}




\begin{sidewaystable}
\centering

\caption{
\textbf{Certification over an extended perturbation range on CelebA.}
Results are reported as mean $\pm$ standard deviation over five
certification seeds. $\Delta$ denotes the absolute difference between
the mean manifold and isotropic results. Certified radii are reported
for the corresponding smoothing geometry.
}
\label{tab:full_celeba}

\scriptsize
\setlength{\tabcolsep}{10pt}
\renewcommand{\arraystretch}{1.1}

\setlength{\tabcolsep}{0pt}

\begin{tabular}{
c
@{\hspace{10pt}}
c@{\hspace{3pt}}c@{\hspace{3pt}}c
@{\hspace{14pt}}
c@{\hspace{3pt}}c@{\hspace{3pt}}c
@{\hspace{14pt}}
c@{\hspace{4pt}}c
@{\hspace{14pt}}
c@{\hspace{4pt}}c
}

\toprule

\rowcolor{groupgray}
\multicolumn{11}{c}{
\textbf{Classification on CelebA  test set ($N=100$)}
}
\\

\midrule

\multirow{2}{*}{$\sigma$}
&
\multicolumn{3}{c}{Certified Accuracy $\uparrow$}
&
\multicolumn{3}{c}{Abstention $\varnothing$ $\downarrow$}
&
\multicolumn{2}{c}{Radius}
&
\multicolumn{2}{c}{Volume}
\\

\cmidrule(lr){2-4}
\cmidrule(lr){5-7}
\cmidrule(lr){8-9}
\cmidrule(lr){10-11}

&
$\mathrm{CAcc}_{\mathrm{I}}$
&
$\mathrm{CAcc}_{\mathrm{M}}$
&
$\Delta$
&
$\varnothing_{\mathrm{I}}$
&
$\varnothing_{\mathrm{M}}$
&
$\Delta$
&
$R_{\mathrm{I}}$
&
$r_{\mathrm{white},\mathrm{M}}$
&
$\log V_{\mathrm{I}}$
&
$\log V_{\mathrm{M}}$
\\

\midrule
0.02
& $93.00{\scriptstyle\pm0.00}$
& $92.00{\scriptstyle\pm0.00}$
& {\textcolor{mutedred}{$\downarrow1.00$}}
& $0.00{\scriptstyle\pm0.00}$
& $0.00{\scriptstyle\pm0.00}$
& {\textcolor{black}{$\rightarrow0.00$}}
& $0.029{\scriptstyle\pm0.000}$
& $0.030{\scriptstyle\pm0.000}$
& $-1272575.76$
& $-2656.71$
\\

0.05
& $87.00{\scriptstyle\pm0.00}$
& $\mathbf{92.00{\scriptstyle\pm0.00}}$
& {\textcolor{gaincolor}{$\uparrow5.00$}}
& $3.00{\scriptstyle\pm0.00}$
& $\mathbf{0.25{\scriptstyle\pm0.50}}$
& {\textcolor{gaincolor}{$\downarrow2.75$}}
& $0.073{\scriptstyle\pm0.000}$
& $0.074{\scriptstyle\pm0.000}$
& $-1134648.35$
& $-2198.57$
\\

0.10
& $73.00{\scriptstyle\pm0.00}$
& $\mathbf{92.00{\scriptstyle\pm0.00}}$
& {\textcolor{gaincolor}{$\uparrow19.00$}}
& $1.75{\scriptstyle\pm0.50}$
& $\mathbf{1.00{\scriptstyle\pm0.00}}$
& {\textcolor{gaincolor}{$\downarrow0.75$}}
& $0.140{\scriptstyle\pm0.001}$
& $0.148{\scriptstyle\pm0.000}$
& $-1030310.29$
& $-1852.20$
\\

0.20
& $45.00{\scriptstyle\pm0.00}$
& $\mathbf{92.00{\scriptstyle\pm0.00}}$
& {\textcolor{gaincolor}{$\uparrow47.00$}}
& $0.00{\scriptstyle\pm0.00}$
& $2.20{\scriptstyle\pm0.45}$
& {\textcolor{mutedred}{$\uparrow2.20$}}
& $0.300{\scriptstyle\pm0.000}$
& $0.295{\scriptstyle\pm0.001}$
& $-925972.24$
& $-1505.57$
\\

0.30
& $45.00{\scriptstyle\pm0.00}$
& $\mathbf{92.00{\scriptstyle\pm0.00}}$
& {\textcolor{gaincolor}{$\uparrow47.00$}}
& $0.00{\scriptstyle\pm0.00}$
& $2.40{\scriptstyle\pm0.55}$
& {\textcolor{mutedred}{$\uparrow2.40$}}
& $0.450{\scriptstyle\pm0.000}$
& $0.432{\scriptstyle\pm0.003}$
& $-864938.38$
& $-1302.90$
\\

0.40
& $45.00{\scriptstyle\pm0.00}$
& $\mathbf{92.00{\scriptstyle\pm0.00}}$
& {\textcolor{gaincolor}{$\uparrow47.00$}}
& $0.00{\scriptstyle\pm0.00}$
& $3.00{\scriptstyle\pm0.00}$
& {\textcolor{mutedred}{$\uparrow3.00$}}
& $0.600{\scriptstyle\pm0.000}$
& $0.564{\scriptstyle\pm0.002}$
& $-821634.18$
& $-1159.00$
\\

0.50
& $45.00{\scriptstyle\pm0.00}$
& $\mathbf{92.00{\scriptstyle\pm0.00}}$
& {\textcolor{gaincolor}{$\uparrow47.00$}}
& $0.00{\scriptstyle\pm0.00}$
& $3.00{\scriptstyle\pm0.00}$
& {\textcolor{mutedred}{$\uparrow3.00$}}
& $0.750{\scriptstyle\pm0.000}$
& $0.678{\scriptstyle\pm0.002}$
& $-788044.82$
& $-1047.43$
\\

0.60
& $45.00{\scriptstyle\pm0.00}$
& $\mathbf{91.60{\scriptstyle\pm0.55}}$
& {\textcolor{gaincolor}{$\uparrow46.60$}}
& $0.00{\scriptstyle\pm0.00}$
& $3.40{\scriptstyle\pm0.55}$
& {\textcolor{mutedred}{$\uparrow3.40$}}
& $0.900{\scriptstyle\pm0.000}$
& $0.783{\scriptstyle\pm0.003}$
& $-760600.33$
& $-956.27$
\\

0.70
& $45.00{\scriptstyle\pm0.00}$
& $\mathbf{90.00{\scriptstyle\pm0.00}}$
& {\textcolor{gaincolor}{$\uparrow45.00$}}
& $0.00{\scriptstyle\pm0.00}$
& $5.60{\scriptstyle\pm0.55}$
& {\textcolor{mutedred}{$\uparrow5.60$}}
& $1.050{\scriptstyle\pm0.000}$
& $0.897{\scriptstyle\pm0.006}$
& $-737396.33$
& $-879.01$
\\

0.80
& $45.00{\scriptstyle\pm0.00}$
& $\mathbf{89.20{\scriptstyle\pm0.45}}$
& {\textcolor{gaincolor}{$\uparrow44.20$}}
& $0.00{\scriptstyle\pm0.00}$
& $6.40{\scriptstyle\pm0.89}$
& {\textcolor{mutedred}{$\uparrow6.40$}}
& $1.200{\scriptstyle\pm0.000}$
& $0.979{\scriptstyle\pm0.013}$
& $-717296.12$
& $-811.50$
\\

0.90
& $45.00{\scriptstyle\pm0.00}$
& $\mathbf{87.60{\scriptstyle\pm1.34}}$
& {\textcolor{gaincolor}{$\uparrow42.60$}}
& $0.00{\scriptstyle\pm0.00}$
& $8.40{\scriptstyle\pm1.52}$
& {\textcolor{mutedred}{$\uparrow8.40$}}
& $1.350{\scriptstyle\pm0.000}$
& $1.071{\scriptstyle\pm0.016}$
& $-699566.47$
& $-752.49$
\\

1.00
& $45.00{\scriptstyle\pm0.00}$
& $\mathbf{85.60{\scriptstyle\pm1.52}}$
& {\textcolor{gaincolor}{$\uparrow40.60$}}
& $0.00{\scriptstyle\pm0.00}$
& $10.60{\scriptstyle\pm1.52}$
& {\textcolor{mutedred}{$\uparrow10.60$}}
& $1.500{\scriptstyle\pm0.000}$
& $1.157{\scriptstyle\pm0.020}$
& $-683706.77$
& $-700.83$
\\

2.00
& $45.00{\scriptstyle\pm0.00}$
& $\mathbf{75.25{\scriptstyle\pm2.63}}$
& {\textcolor{gaincolor}{$\uparrow30.25$}}
& $0.00{\scriptstyle\pm0.00}$
& $21.25{\scriptstyle\pm2.22}$
& {\textcolor{mutedred}{$\uparrow21.25$}}
& $3.001{\scriptstyle\pm0.000}$
& $1.605{\scriptstyle\pm0.047}$
& $-579368.71$
& $-351.51$
\\

\bottomrule
\end{tabular}
\end{sidewaystable}


\begin{sidewaystable}
\centering

\caption{
\textbf{Certification over an extended perturbation range on CelebA-HQ.}
Results are reported as mean $\pm$ standard deviation over five
certification seeds. $\Delta$ denotes the absolute difference between
the mean manifold and isotropic results. Certified radii are reported
for the corresponding smoothing geometry.
}
\label{tab:full_celebahq}

\scriptsize
\setlength{\tabcolsep}{10pt}
\renewcommand{\arraystretch}{1.1}

\setlength{\tabcolsep}{0pt}

\begin{tabular}{
c
@{\hspace{10pt}}
c@{\hspace{3pt}}c@{\hspace{3pt}}c
@{\hspace{14pt}}
c@{\hspace{3pt}}c@{\hspace{3pt}}c
@{\hspace{14pt}}
c@{\hspace{4pt}}c
@{\hspace{14pt}}
c@{\hspace{4pt}}c
}

\toprule

\rowcolor{groupgray}
\multicolumn{11}{c}{
\textbf{Classification on CelebA-HQ  test set ($N=100$)}
}
\\

\midrule

\multirow{2}{*}{$\sigma$}
&
\multicolumn{3}{c}{Certified Accuracy $\uparrow$}
&
\multicolumn{3}{c}{Abstention $\varnothing$ $\downarrow$}
&
\multicolumn{2}{c}{Radius}
&
\multicolumn{2}{c}{Volume}
\\

\cmidrule(lr){2-4}
\cmidrule(lr){5-7}
\cmidrule(lr){8-9}
\cmidrule(lr){10-11}

&
$\mathrm{CAcc}_{\mathrm{I}}$
&
$\mathrm{CAcc}_{\mathrm{M}}$
&
$\Delta$
&
$\varnothing_{\mathrm{I}}$
&
$\varnothing_{\mathrm{M}}$
&
$\Delta$
&
$R_{\mathrm{I}}$
&
$r_{\mathrm{white},\mathrm{M}}$
&
$\log V_{\mathrm{I}}$
&
$\log V_{\mathrm{M}}$
\\

\midrule

0.02
& $94.00{\scriptstyle\pm0.00}$
& $93.00{\scriptstyle\pm0.00}$
& {\textcolor{mutedred}{$\downarrow1.00$}}
& $2.00{\scriptstyle\pm0.00}$
& $\mathbf{0.00{\scriptstyle\pm0.00}}$
& {\textcolor{gaincolor}{$\downarrow2.00$}}
& $0.029{\scriptstyle\pm0.000}$
& $0.029{\scriptstyle\pm0.000}$
& $-7298658.81$
& $-2184.34$
\\

0.05
& $91.25{\scriptstyle\pm0.50}$
& $\mathbf{92.25{\scriptstyle\pm0.50}}$
& {\textcolor{gaincolor}{$\uparrow1.00$}}
& $0.75{\scriptstyle\pm0.50}$
& $0.75{\scriptstyle\pm0.50}$
& {\textcolor{black}{$\rightarrow0.00$}}
& $0.071{\scriptstyle\pm0.001}$
& $0.073{\scriptstyle\pm0.000}$
& $-6578058.46$
& $-1726.32$
\\

0.10
& $83.75{\scriptstyle\pm0.50}$
& $\mathbf{92.00{\scriptstyle\pm0.00}}$
& {\textcolor{gaincolor}{$\uparrow8.25$}}
& $0.25{\scriptstyle\pm0.50}$
& $2.75{\scriptstyle\pm0.50}$
& {\textcolor{mutedred}{$\uparrow2.50$}}
& $0.141{\scriptstyle\pm0.001}$
& $0.146{\scriptstyle\pm0.001}$
& $-6032945.33$
& $-1379.72$
\\

0.20
& $53.00{\scriptstyle\pm0.00}$
& $\mathbf{93.00{\scriptstyle\pm0.00}}$
& {\textcolor{gaincolor}{$\uparrow40.00$}}
& $1.00{\scriptstyle\pm0.00}$
& $3.00{\scriptstyle\pm0.82}$
& {\textcolor{mutedred}{$\uparrow2.00$}}
& $0.272{\scriptstyle\pm0.001}$
& $0.290{\scriptstyle\pm0.003}$
& $-5487832.21$
& $-1033.18$
\\

0.30
& $50.00{\scriptstyle\pm0.00}$
& $\mathbf{93.00{\scriptstyle\pm0.00}}$
& {\textcolor{gaincolor}{$\uparrow43.00$}}
& $0.00{\scriptstyle\pm0.00}$
& $3.25{\scriptstyle\pm0.50}$
& {\textcolor{mutedred}{$\uparrow3.25$}}
& $0.450{\scriptstyle\pm0.000}$
& $0.428{\scriptstyle\pm0.002}$
& $-5168961.47$
& $-830.31$
\\

0.40
& $50.00{\scriptstyle\pm0.00}$
& $\mathbf{93.25{\scriptstyle\pm0.50}}$
& {\textcolor{gaincolor}{$\uparrow43.25$}}
& $0.00{\scriptstyle\pm0.00}$
& $3.50{\scriptstyle\pm0.58}$
& {\textcolor{mutedred}{$\uparrow3.50$}}
& $0.600{\scriptstyle\pm0.000}$
& $0.552{\scriptstyle\pm0.003}$
& $-4942719.09$
& $-686.47$
\\

0.50
& $50.00{\scriptstyle\pm0.00}$
& $\mathbf{92.75{\scriptstyle\pm0.50}}$
& {\textcolor{gaincolor}{$\uparrow42.75$}}
& $0.00{\scriptstyle\pm0.00}$
& $4.00{\scriptstyle\pm0.82}$
& {\textcolor{mutedred}{$\uparrow4.00$}}
& $0.750{\scriptstyle\pm0.000}$
& $0.657{\scriptstyle\pm0.005}$
& $-4767231.86$
& $-574.90$
\\

0.60
& $50.00{\scriptstyle\pm0.00}$
& $\mathbf{90.50{\scriptstyle\pm1.29}}$
& {\textcolor{gaincolor}{$\uparrow40.50$}}
& $0.00{\scriptstyle\pm0.00}$
& $6.50{\scriptstyle\pm1.29}$
& {\textcolor{mutedred}{$\uparrow6.50$}}
& $0.900{\scriptstyle\pm0.000}$
& $0.754{\scriptstyle\pm0.011}$
& $-4623848.35$
& $-483.42$
\\

0.70
& $50.00{\scriptstyle\pm0.00}$
& $\mathbf{89.75{\scriptstyle\pm1.26}}$
& {\textcolor{gaincolor}{$\uparrow39.75$}}
& $0.00{\scriptstyle\pm0.00}$
& $7.25{\scriptstyle\pm1.26}$
& {\textcolor{mutedred}{$\uparrow7.25$}}
& $1.050{\scriptstyle\pm0.000}$
& $0.832{\scriptstyle\pm0.006}$
& $-4502619.32$
& $-406.99$
\\

0.80
& $50.00{\scriptstyle\pm0.00}$
& $\mathbf{88.50{\scriptstyle\pm0.58}}$
& {\textcolor{gaincolor}{$\uparrow38.50$}}
& $0.00{\scriptstyle\pm0.00}$
& $8.50{\scriptstyle\pm0.58}$
& {\textcolor{mutedred}{$\uparrow8.50$}}
& $1.200{\scriptstyle\pm0.000}$
& $0.896{\scriptstyle\pm0.011}$
& $-4397605.96$
& $-340.18$
\\

0.90
& $50.00{\scriptstyle\pm0.00}$
& $\mathbf{87.25{\scriptstyle\pm1.50}}$
& {\textcolor{gaincolor}{$\uparrow37.25$}}
& $0.00{\scriptstyle\pm0.00}$
& $10.00{\scriptstyle\pm1.83}$
& {\textcolor{mutedred}{$\uparrow10.00$}}
& $1.350{\scriptstyle\pm0.000}$
& $0.942{\scriptstyle\pm0.021}$
& $-4304977.61$
& $-280.77$
\\

1.00
& $50.00{\scriptstyle\pm0.00}$
& $\mathbf{84.25{\scriptstyle\pm0.50}}$
& {\textcolor{gaincolor}{$\uparrow34.25$}}
& $0.00{\scriptstyle\pm0.00}$
& $12.25{\scriptstyle\pm0.96}$
& {\textcolor{mutedred}{$\uparrow12.25$}}
& $1.500{\scriptstyle\pm0.000}$
& $1.010{\scriptstyle\pm0.014}$
& $-4222118.73$
& $-229.39$
\\

2.00
& $50.00{\scriptstyle\pm0.00}$
& $\mathbf{65.75{\scriptstyle\pm0.96}}$
& {\textcolor{gaincolor}{$\uparrow15.75$}}
& $0.00{\scriptstyle\pm0.00}$
& $26.75{\scriptstyle\pm2.06}$
& {\textcolor{mutedred}{$\uparrow26.75$}}
& $3.001{\scriptstyle\pm0.000}$
& $1.267{\scriptstyle\pm0.065}$
& $-3677005.61$
& $118.75$
\\

\bottomrule
\end{tabular}
\end{sidewaystable}


\begin{table}[t]
\centering

\caption{\textbf{Certification over an extended perturbation range for semantic
segmentation on CelebAMask-HQ.}
Results are reported as mean $\pm$ standard deviation over five
certification seeds. $\Delta$ denotes the absolute difference between
the mean manifold and isotropic results. Segmentation uses
$\tau=0.75$, yielding a fixed certification radius for each noise scale
$\sigma$.
}
\label{tab:full_segmentation}

\tiny
\setlength{\tabcolsep}{10pt}
\renewcommand{\arraystretch}{1.1}

\setlength{\tabcolsep}{0pt}

\begin{tabular}{
c
@{\hspace{10pt}}
c@{\hspace{3pt}}c@{\hspace{3pt}}c
@{\hspace{14pt}}
c@{\hspace{3pt}}c@{\hspace{3pt}}c
@{\hspace{14pt}}
c@{\hspace{4pt}}c
@{\hspace{14pt}}
c@{\hspace{4pt}}c
}
\toprule

\rowcolor{groupgray}
\multicolumn{11}{c}{
\textbf{Semantic Segmentation on CelebAMask-HQ  test set ($N=100$)}
}
\\

\midrule

&
\multicolumn{3}{c}{Certified Accuracy $\uparrow$}
&
\multicolumn{3}{c}{Abstention $\varnothing$ $\downarrow$}
&
\multicolumn{2}{c}{Radius}
&
\multicolumn{2}{c}{$\mathrm{mIOU}$ $\uparrow$}
\\

\cmidrule(lr){2-4}
\cmidrule(lr){5-7}
\cmidrule(lr){8-9}
\cmidrule(lr){10-11}

$\sigma$
&
$\mathrm{CAcc}_{\mathrm{I}}$
&
$\mathrm{CAcc}_{\mathrm{M}}$
&
$\Delta$
&
$\varnothing_{\mathrm{I}}$
&
$\varnothing_{\mathrm{M}}$
&
$\Delta$
&
$R_{\mathrm{I}}$
&
$r_{\mathrm{white},\mathrm{M}}$
&
$\mathrm{mIOU}_{\mathrm{I}}$
&
$\mathrm{mIOU}_{\mathrm{M}}$
\\

\midrule
0.01
& $94.78{\scriptstyle\pm0.01}$
& $\mathbf{95.20{\scriptstyle\pm0.00}}$
& {\textcolor{gaincolor}{$\uparrow0.42$}}
& $1.11{\scriptstyle\pm0.01}$
& $\mathbf{0.24{\scriptstyle\pm0.00}}$
& {\textcolor{gaincolor}{$\downarrow0.87$}}
& $0.007$
& $0.007$
& $0.740{\scriptstyle\pm0.000}$
& $0.746{\scriptstyle\pm0.000}$
\\

0.02
& $94.29{\scriptstyle\pm0.01}$
& $\mathbf{95.09{\scriptstyle\pm0.00}}$
& {\textcolor{gaincolor}{$\uparrow0.80$}}
& $1.89{\scriptstyle\pm0.01}$
& $\mathbf{0.48{\scriptstyle\pm0.00}}$
& {\textcolor{gaincolor}{$\downarrow1.41$}}
& $0.013$
& $0.013$
& $0.729{\scriptstyle\pm0.001}$
& $0.746{\scriptstyle\pm0.000}$
\\

0.03
& $93.81{\scriptstyle\pm0.00}$
& $\mathbf{94.98{\scriptstyle\pm0.00}}$
& {\textcolor{gaincolor}{$\uparrow1.17$}}
& $2.53{\scriptstyle\pm0.00}$
& $\mathbf{0.71{\scriptstyle\pm0.00}}$
& {\textcolor{gaincolor}{$\downarrow1.82$}}
& $0.020$
& $0.020$
& $0.720{\scriptstyle\pm0.002}$
& $0.745{\scriptstyle\pm0.000}$
\\

0.05
& $92.82{\scriptstyle\pm0.01}$
& $\mathbf{94.74{\scriptstyle\pm0.01}}$
& {\textcolor{gaincolor}{$\uparrow1.92$}}
& $3.60{\scriptstyle\pm0.01}$
& $\mathbf{1.19{\scriptstyle\pm0.01}}$
& {\textcolor{gaincolor}{$\downarrow2.42$}}
& $0.034$
& $0.034$
& $0.694{\scriptstyle\pm0.002}$
& $0.741{\scriptstyle\pm0.001}$
\\

0.10
& $90.49{\scriptstyle\pm0.03}$
& $\mathbf{93.99{\scriptstyle\pm0.03}}$
& {\textcolor{gaincolor}{$\uparrow3.50$}}
& $5.83{\scriptstyle\pm0.04}$
& $\mathbf{2.44{\scriptstyle\pm0.03}}$
& {\textcolor{gaincolor}{$\downarrow3.38$}}
& $0.067$
& $0.067$
& $0.608{\scriptstyle\pm0.004}$
& $0.732{\scriptstyle\pm0.000}$
\\

0.20
& $83.93{\scriptstyle\pm0.02}$
& $\mathbf{92.31{\scriptstyle\pm0.02}}$
& {\textcolor{gaincolor}{$\uparrow8.38$}}
& $11.22{\scriptstyle\pm0.03}$
& $\mathbf{4.76{\scriptstyle\pm0.02}}$
& {\textcolor{gaincolor}{$\downarrow6.46$}}
& $0.135$
& $0.135$
& $0.454{\scriptstyle\pm0.001}$
& $0.710{\scriptstyle\pm0.001}$
\\

0.30
& $68.94{\scriptstyle\pm0.09}$
& $\mathbf{90.03{\scriptstyle\pm0.04}}$
& {\textcolor{gaincolor}{$\uparrow21.09$}}
& $23.59{\scriptstyle\pm0.05}$
& $\mathbf{7.45{\scriptstyle\pm0.04}}$
& {\textcolor{gaincolor}{$\downarrow16.15$}}
& $0.202$
& $0.202$
& $0.299{\scriptstyle\pm0.001}$
& $0.676{\scriptstyle\pm0.002}$
\\

0.40
& $41.77{\scriptstyle\pm0.01}$
& $\mathbf{87.34{\scriptstyle\pm0.06}}$
& {\textcolor{gaincolor}{$\uparrow45.57$}}
& $42.65{\scriptstyle\pm0.05}$
& $\mathbf{10.49{\scriptstyle\pm0.05}}$
& {\textcolor{gaincolor}{$\downarrow32.16$}}
& $0.270$
& $0.270$
& $0.137{\scriptstyle\pm0.000}$
& $0.632{\scriptstyle\pm0.002}$
\\

0.50
& $26.87{\scriptstyle\pm0.11}$
& $\mathbf{84.25{\scriptstyle\pm0.12}}$
& {\textcolor{gaincolor}{$\uparrow57.38$}}
& $39.91{\scriptstyle\pm0.15}$
& $\mathbf{13.81{\scriptstyle\pm0.13}}$
& {\textcolor{gaincolor}{$\downarrow26.10$}}
& $0.337$
& $0.337$
& $0.054{\scriptstyle\pm0.000}$
& $0.575{\scriptstyle\pm0.003}$
\\

0.60
& $26.03{\scriptstyle\pm0.11}$
& $\mathbf{81.12{\scriptstyle\pm0.05}}$
& {\textcolor{gaincolor}{$\uparrow55.10$}}
& $24.58{\scriptstyle\pm0.16}$
& $\mathbf{17.11{\scriptstyle\pm0.06}}$
& {\textcolor{gaincolor}{$\downarrow7.47$}}
& $0.405$
& $0.405$
& $0.029{\scriptstyle\pm0.000}$
& $0.519{\scriptstyle\pm0.003}$
\\

0.70
& $27.57{\scriptstyle\pm0.04}$
& $\mathbf{78.07{\scriptstyle\pm0.10}}$
& {\textcolor{gaincolor}{$\uparrow50.50$}}
& $13.07{\scriptstyle\pm0.12}$
& $20.25{\scriptstyle\pm0.10}$
& {\textcolor{mutedred}{$\uparrow7.18$}}
& $0.472$
& $0.472$
& $0.024{\scriptstyle\pm0.000}$
& $0.475{\scriptstyle\pm0.001}$
\\

0.80
& $27.94{\scriptstyle\pm0.02}$
& $\mathbf{74.89{\scriptstyle\pm0.09}}$
& {\textcolor{gaincolor}{$\uparrow46.94$}}
& $8.60{\scriptstyle\pm0.06}$
& $23.46{\scriptstyle\pm0.08}$
& {\textcolor{mutedred}{$\uparrow14.86$}}
& $0.540$
& $0.540$
& $0.023{\scriptstyle\pm0.000}$
& $0.429{\scriptstyle\pm0.001}$
\\

0.90
& $27.55{\scriptstyle\pm0.02}$
& $\mathbf{71.31{\scriptstyle\pm0.16}}$
& {\textcolor{gaincolor}{$\uparrow43.76$}}
& $8.05{\scriptstyle\pm0.06}$
& $27.05{\scriptstyle\pm0.17}$
& {\textcolor{mutedred}{$\uparrow19.00$}}
& $0.607$
& $0.607$
& $0.022{\scriptstyle\pm0.000}$
& $0.393{\scriptstyle\pm0.002}$
\\

1.00
& $27.00{\scriptstyle\pm0.02}$
& $\mathbf{67.84{\scriptstyle\pm0.17}}$
& {\textcolor{gaincolor}{$\uparrow40.84$}}
& $8.80{\scriptstyle\pm0.06}$
& $30.50{\scriptstyle\pm0.20}$
& {\textcolor{mutedred}{$\uparrow21.71$}}
& $0.674$
& $0.674$
& $0.021{\scriptstyle\pm0.000}$
& $0.358{\scriptstyle\pm0.003}$
\\

\bottomrule
\end{tabular}
\end{table}


\begin{sidewaystable}
\centering
\tiny
\caption{
\textbf{Certification over an extended perturbation range in the sparse concept representation space on ImageNet.}
Results are reported as mean $\pm$ standard deviation over five
certification seeds. $\Delta$ denotes the absolute difference between the mean
manifold and isotropic results. Certified radii are reported
for the corresponding smoothing geometry.
}
\label{tab:full_conceptspace}

\scriptsize
\setlength{\tabcolsep}{0pt}
\renewcommand{\arraystretch}{1.12}

\begin{tabular}{
@{\hspace{6pt}}c@{\hspace{7pt}}
c@{\hspace{3pt}}c@{\hspace{3pt}}c
@{\hspace{10pt}}
c@{\hspace{3pt}}c@{\hspace{3pt}}c
@{\hspace{10pt}}
c@{\hspace{4pt}}c
@{\hspace{10pt}}
c@{\hspace{4pt}}c
}

\toprule

\rowcolor{groupgray}
\multicolumn{11}{c}{
\textbf{Sparse Concept Space on ImageNet ($N=5000$)}
}
\\

\midrule

&
\multicolumn{3}{c}{Certified Accuracy $\uparrow$}
&
\multicolumn{3}{c}{Abstention $\varnothing$ $\downarrow$}
&
\multicolumn{2}{c}{Radius}
&
\multicolumn{2}{c}{Volume}
\\

\cmidrule(lr){2-4}
\cmidrule(lr){5-7}
\cmidrule(lr){8-9}
\cmidrule(lr){10-11}

\addlinespace[2pt]

$\sigma$
&
\(\mathrm{CAcc}_{\mathrm I}\)
&
\(\mathrm{CAcc}_{\mathrm M}\)
&
\(\Delta\)
&
\(\varnothing_{\mathrm I}\)
&
\(\varnothing_{\mathrm M}\)
&
\(\Delta\)
&
\(R_{\mathrm I}\)
&
\(r_{\mathrm{white},\mathrm M}\)
&
\(\log V_{\mathrm I}\)
&
\(\log V_{\mathrm M}\)
\\

\midrule
0.10 & $45.72{\scriptstyle\pm0.33}$ & $\mathbf{68.60{\scriptstyle\pm0.00}}$ & {\textcolor{gaincolor}{$\uparrow22.88$}} & $48.08{\scriptstyle\pm0.30}$ & $\mathbf{3.00{\scriptstyle\pm0.20}}$ & {\textcolor{gaincolor}{$\downarrow45.08$}} & $0.083{\scriptstyle\pm0.001}$ & $0.195{\scriptstyle\pm0.000}$ & $-44152.61$ & $-2958.18$ \\
0.20 & $6.52{\scriptstyle\pm0.41}$ & $\mathbf{67.44{\scriptstyle\pm0.09}}$ & {\textcolor{gaincolor}{$\uparrow60.92$}} & $93.12{\scriptstyle\pm0.41}$ & $\mathbf{7.96{\scriptstyle\pm0.17}}$ & {\textcolor{gaincolor}{$\downarrow85.16$}} & $0.100{\scriptstyle\pm0.005}$ & $0.356{\scriptstyle\pm0.001}$ & $-38474.35$ & $-2614.11$ \\
0.30 & $0.80{\scriptstyle\pm0.00}$ & $\mathbf{66.76{\scriptstyle\pm0.17}}$ & {\textcolor{gaincolor}{$\uparrow65.96$}} & $99.20{\scriptstyle\pm0.00}$ & $\mathbf{10.96{\scriptstyle\pm0.55}}$ & {\textcolor{gaincolor}{$\downarrow88.24$}} & $0.114{\scriptstyle\pm0.007}$ & $0.471{\scriptstyle\pm0.003}$ & $-35152.78$ & $-2412.84$ \\
0.40 & $0.16{\scriptstyle\pm0.09}$ & $\mathbf{65.24{\scriptstyle\pm0.30}}$ & {\textcolor{gaincolor}{$\uparrow65.08$}} & $99.84{\scriptstyle\pm0.09}$ & $\mathbf{15.64{\scriptstyle\pm0.38}}$ & {\textcolor{gaincolor}{$\downarrow84.20$}} & $0.040{\scriptstyle\pm0.029}$ & $0.554{\scriptstyle\pm0.004}$ & $-32796.09$ & $-2270.04$ \\
0.50 & $0.00{\scriptstyle\pm0.00}$ & $\mathbf{63.96{\scriptstyle\pm0.17}}$ & {\textcolor{gaincolor}{$\uparrow63.96$}} & $100.00{\scriptstyle\pm0.00}$ & $\mathbf{19.72{\scriptstyle\pm0.46}}$ & {\textcolor{gaincolor}{$\downarrow80.28$}} & $\text{--}$ & $0.601{\scriptstyle\pm0.004}$ & $-30968.10$ & $-2159.27$ \\
0.60 & $0.00{\scriptstyle\pm0.00}$ & $\mathbf{61.68{\scriptstyle\pm0.27}}$ & {\textcolor{gaincolor}{$\uparrow61.68$}} & $100.00{\scriptstyle\pm0.00}$ & $\mathbf{23.76{\scriptstyle\pm0.55}}$ & {\textcolor{gaincolor}{$\downarrow76.24$}} & $\text{--}$ & $0.620{\scriptstyle\pm0.007}$ & $-29474.52$ & $-2068.77$ \\
0.70 & $0.00{\scriptstyle\pm0.00}$ & $\mathbf{58.56{\scriptstyle\pm0.43}}$ & {\textcolor{gaincolor}{$\uparrow58.56$}} & $100.00{\scriptstyle\pm0.00}$ & $\mathbf{30.08{\scriptstyle\pm0.73}}$ & {\textcolor{gaincolor}{$\downarrow69.92$}} & $\text{--}$ & $0.641{\scriptstyle\pm0.009}$ & $-28211.72$ & $-1992.25$ \\
0.80 & $0.00{\scriptstyle\pm0.00}$ & $\mathbf{54.72{\scriptstyle\pm0.46}}$ & {\textcolor{gaincolor}{$\uparrow54.72$}} & $100.00{\scriptstyle\pm0.00}$ & $\mathbf{36.00{\scriptstyle\pm0.51}}$ & {\textcolor{gaincolor}{$\downarrow64.00$}} & $\text{--}$ & $0.648{\scriptstyle\pm0.006}$ & $-27117.83$ & $-1925.97$ \\
0.90 & $0.00{\scriptstyle\pm0.00}$ & $\mathbf{51.04{\scriptstyle\pm0.52}}$ & {\textcolor{gaincolor}{$\uparrow51.04$}} & $100.00{\scriptstyle\pm0.00}$ & $\mathbf{41.76{\scriptstyle\pm0.70}}$ & {\textcolor{gaincolor}{$\downarrow58.24$}} & $\text{--}$ & $0.647{\scriptstyle\pm0.009}$ & $-26152.95$ & $-1867.50$ \\
1.00 & $0.00{\scriptstyle\pm0.00}$ & $\mathbf{47.24{\scriptstyle\pm0.26}}$ & {\textcolor{gaincolor}{$\uparrow47.24$}} & $100.00{\scriptstyle\pm0.00}$ & $\mathbf{46.76{\scriptstyle\pm0.26}}$ & {\textcolor{gaincolor}{$\downarrow53.24$}} & $\text{--}$ & $0.635{\scriptstyle\pm0.004}$ & $-25289.83$ & $-1815.20$ \\

\bottomrule
\end{tabular}
\end{sidewaystable}

\subsection{Class-wise Segmentation Gains}
Figure~\ref{fig:segmentation_delta_iou_heatmap} reveals consistent class-wise gains from
manifold smoothing. Improvements already appear at low noise for small,
localized fine semantic facial regions and become widespread under stronger perturbations,
where manifold smoothing substantially outperforms isotropic smoothing across
most classes. 
\label{app:segmentation_iou_class_wise}
\input{tikz_fig/segmentation_per_class_iou}

\subsection{Image-to-Text Alignment Results}
\label{app:rococo_extended}
\textbf{Results across RoCOCO caption banks.}
We extend the evaluation to all RoCOCO caption banks, including Danger,
Same-Concept, Diff-Concept, and Rand-Vocab in Tab. \ref{tab:rococo_smoothing}. Across all settings and
perturbation levels, manifold smoothing consistently retains higher
R@1 than isotropic smoothing, with the difference becoming particularly
pronounced as $\sigma$ increases. The same trend is reflected in the
smaller retrieval-performance drop under manifold smoothing. These results
suggest that restricting perturbations to locally supported directions
better preserves the image--text alignment relevant to retrieval, rather
than disrupting the representation along arbitrary ambient directions.

\begin{table}[t]
\centering
\caption{\textbf{Image-to-Text retrieval robustness on RoCOCO} using CLIP ViT-B/32 with isotropic and manifold smoothing. Drop (\%) is computed relative to the clean CLIP performance on the original MS-COCO benchmark.} 

\label{tab:rococo_smoothing}
\resizebox{\textwidth}{!}{
\begin{tabular}{ll|c|ccc|ccc|ccc|ccc}
\toprule
& & \textbf{Base COCO}
& \multicolumn{3}{c|}{\textbf{Danger}}
& \multicolumn{3}{c|}{\textbf{Same-Concept}}
& \multicolumn{3}{c|}{\textbf{Diff-Concept}}
& \multicolumn{3}{c}{\textbf{Rand-Voca}} \\
\cmidrule(lr){3-3}
\cmidrule(lr){4-6}
\cmidrule(lr){7-9}
\cmidrule(lr){10-12}
\cmidrule(lr){13-15}
$\sigma$ & Mode
& R@1 $\uparrow$
& R@1 $\uparrow$ & RSMS $\downarrow$ & Drop $\downarrow$
& R@1 $\uparrow$ & RSMS $\downarrow$ & Drop $\downarrow$
& R@1 $\uparrow$ & RSMS $\downarrow$ & Drop $\downarrow$
& R@1 $\uparrow$ & RSMS $\downarrow$ & Drop $\downarrow$ \\
\midrule

-- & CLIP
& 50.16
& 41.26 & 23.00 & 17.70
& 39.00 & 29.30 & 22.20
& 40.28 & 26.52 & 19.70
& 39.18 & 28.52 & 21.90 \\

\addlinespace[2mm]

\multirow{2}{*}{$0.02$}
& ISO
& 43.30
& 35.24 & 26.46 & 29.70
& 32.38 & 34.62 & 35.40
& 33.40 & 33.54 & 33.40
& 33.16 & 33.52 & 33.90 \\

&
Mani
& 50.16
& 41.26 & 22.98 & 17.70
& 39.00 & 29.12 & 22.20
& 40.48 & 26.38 & 19.30
& 39.18 & 28.62 & 21.90 \\

\addlinespace[2mm]

\multirow{2}{*}{$0.05$}
& ISO
& 22.74
& 17.06 & 32.20 & 66.00
& 13.44 & 51.58 & 73.20
& 14.28 & 50.20 & 71.50
& 15.06 & 46.62 & 70.00 \\

&
Mani
& 50.50
& 41.34 & 22.90 & 17.60
& 38.78 & 29.50 & 22.70
& 40.04 & 26.52 & 20.20
& 39.28 & 28.28 & 21.70 \\

\addlinespace[2mm]

\multirow{2}{*}{$0.07$}
& ISO
& 11.70
& 8.54 & 32.90 & 83.00
& 7.14 & 55.20 & 85.80
& 6.88 & 58.02 & 86.30
& 7.30 & 53.20 & 85.40 \\

&
Mani
& 49.90
& 41.24 & 23.26 & 17.80
& 38.72 & 29.60 & 22.80
& 40.20 & 26.68 & 19.90
& 39.02 & 28.52 & 22.20 \\

\addlinespace[2mm]

\multirow{2}{*}{$0.10$}
& ISO
& 4.26
& 3.08 & 32.96 & 93.90
& 2.52 & 57.94 & 95.00
& 2.64 & 60.50 & 94.70
& 2.60 & 53.66 & 94.80 \\

&
Mani
& 50.16
& 41.52 & 22.80 & 17.20
& 38.20 & 29.58 & 23.80
& 39.68 & 26.44 & 20.90
& 38.84 & 28.90 & 22.60 \\

\addlinespace[2mm]

\multirow{2}{*}{$0.20$}
& ISO
& 0.82
& 0.48 & 29.04 & 99.00
& 0.22 & 56.46 & 99.60
& 0.22 & 59.30 & 99.60
& 0.14 & 50.92 & 99.70 \\

&
Mani
& 48.98
& 40.24 & 23.36 & 19.80
& 37.54 & 30.54 & 25.20
& 38.92 & 26.92 & 22.40
& 38.22 & 29.26 & 23.80 \\

\addlinespace[2mm]

\multirow{2}{*}{$0.50$}
& ISO
& 0.16
& 0.02 & 27.48 & 100.00
& 0.06 & 55.28 & 99.90
& 0.04 & 55.14 & 99.90
& 0.06 & 49.04 & 99.90 \\

&
Mani
& 43.22
& 35.14 & 25.26 & 29.90
& 31.46 & 33.94 & 37.30
& 33.48 & 30.92 & 33.30
& 32.82 & 30.74 & 34.60 \\

\addlinespace[2mm]

\multirow{2}{*}{$0.70$}
& ISO
& 0.02
& 0.06 & 26.44 & 99.90
& 0.00 & 54.92 & --
& 0.00 & 55.54 & --
& 0.04 & 48.00 & 99.90 \\

&
Mani
& 37.46
& 29.70 & 26.04 & 40.80
& 26.72 & 37.92 & 46.70
& 27.68 & 34.64 & 44.80
& 27.98 & 34.02 & 44.20 \\

\addlinespace[2mm]

\multirow{2}{*}{$1.00$}
& ISO
& 0.06
& 0.02 & 25.00 & 100.00
& 0.00 & 53.82 & --
& 0.02 & 54.68 & 100.00
& 0.04 & 47.20 & 99.90 \\

&
Mani
& 26.06
& 21.32 & 28.14 & 57.50
& 18.96 & 43.08 & 62.20
& 18.62 & 39.90 & 62.90
& 19.72 & 36.70 & 60.70 \\

\bottomrule
\end{tabular}
}
\end{table}

\subsection{Qualitative Results}
\label{app:qualitative}
\begin{figure}[t]
\centering

%
%
\begin{minipage}[l]{0.55\textwidth}
    \centering
    \includegraphics[
        width=\linewidth,
        height=2.0cm,
        keepaspectratio
    ]{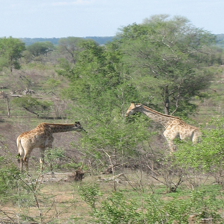}
\end{minipage}
\hspace{0.5em}
\begin{minipage}[c]{0.65\textwidth}
    \small
   
    \vspace{0.2em}
    
     \text{
    Query Image Caption: Two giraffes standing next to each other in a forest.
    }

\end{minipage}

\vspace{0.6em}

%
%
\setlength{\tabcolsep}{2pt}
\renewcommand{\arraystretch}{1.0}

{\footnotesize\text{CLIP}}

\vspace{0.85em}

\begin{tabularx}{\textwidth}{
    >{\centering\arraybackslash}m{0.105\textwidth}
    >{\centering\arraybackslash}m{0.105\textwidth}
    >{\centering\arraybackslash}m{0.105\textwidth}
    >{\centering\arraybackslash}m{0.105\textwidth}
    >{\centering\arraybackslash}m{0.105\textwidth}
    >{\raggedright\arraybackslash}X
}
{\scriptsize\text{Top-1}} &
{\scriptsize\text{Top-2}} &
{\scriptsize\text{Top-3}} &
{\scriptsize\text{Top-4}} &
{\scriptsize\text{Top-5}} &
{\scriptsize\text{Retrieved captions}} \\[0.15em]

\retimg{} &
\retimg{} &
\retimg{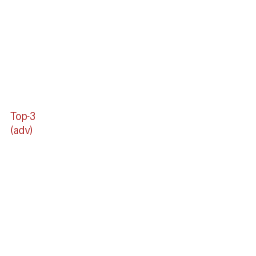} &
\retimg{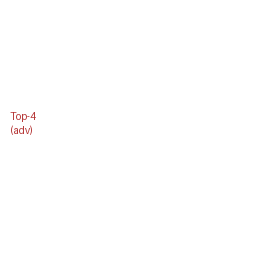} &
\retimg{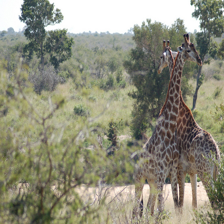} &
\begin{minipage}[c][1.85cm][c]{\linewidth}
\scriptsize
\textcolor{otherblue}{[1] three adult giraffes in a field among trees}\\
\textcolor{otherblue}{[2] Two giraffe are standing near each other in tall brush.}\\
\textcolor{advred}{[3] Two giraffes in a minefield feeding on trees}\\
\textcolor{advred}{[4] three giraffes walking in a minefield with trees in the background}\\
\textcolor{otherblue}{[5] Two giraffes standing neck to neck near some brush.}
\end{minipage}
\end{tabularx}

\vspace{0.80em}

{\footnotesize\text{Isotropic}}

\vspace{0.85em}

\begin{tabularx}{\textwidth}{
    >{\centering\arraybackslash}m{0.105\textwidth}
    >{\centering\arraybackslash}m{0.105\textwidth}
    >{\centering\arraybackslash}m{0.105\textwidth}
    >{\centering\arraybackslash}m{0.105\textwidth}
    >{\centering\arraybackslash}m{0.105\textwidth}
    >{\raggedright\arraybackslash}X
}

\retimg{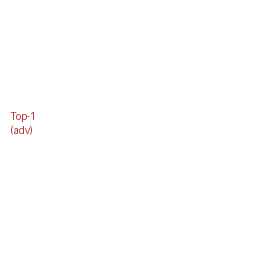} &
\retimg{} &
\retimg{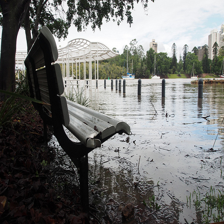} &
\retimg{} &
\retimg{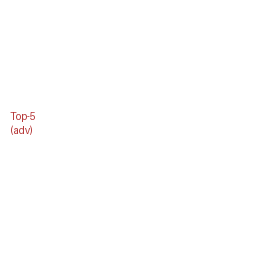} &
\begin{minipage}[c][1.85cm][c]{\linewidth}
\scriptsize
\textcolor{advred}{[3] A messy living room with various weapon and some celebratory balloons}\\
\textcolor{otherblue}{[2] a messy living room with various items and some celebratory balloons}\\
\textcolor{otherblue}{[3] A steel bench overlooks a flooded swamp like area.}\\
\textcolor{otherblue}{[4] A makeshift bedroom is decorated in plastic and cardboard.}\\
\textcolor{advred}{[5] A wooden bench in the lush minefield of a cottage.}
\end{minipage}
\end{tabularx}

\vspace{0.80em}
\vspace{0.85em}

{\footnotesize\text{\textbf{Manifold} (Ours)}}

\vspace{0.85em}
\vspace{0.85em}
\begin{tabularx}{\textwidth}{
    >{\centering\arraybackslash}m{0.105\textwidth}
    >{\centering\arraybackslash}m{0.105\textwidth}
    >{\centering\arraybackslash}m{0.105\textwidth}
    >{\centering\arraybackslash}m{0.105\textwidth}
    >{\centering\arraybackslash}m{0.105\textwidth}
    >{\raggedright\arraybackslash}X
}

\retimg{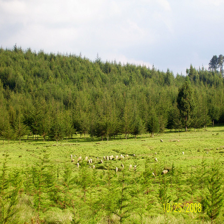} &
\retimg{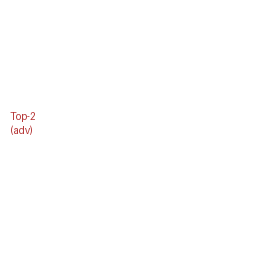} &
\retimg{} &
\retimg{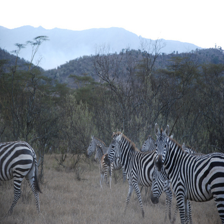} &
\retimg{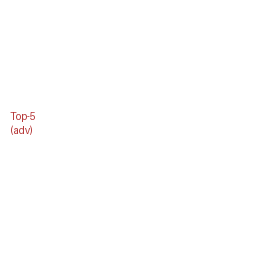} &
\begin{minipage}[c][1.85cm][c]{\linewidth}
\scriptsize
\textcolor{otherblue}{[1] A herd of grazing animals in an open field with trees in the background}\\
\textcolor{advred}{[2] A herd of grazing animals in an open minefield with trees in the background}\\
\textcolor{otherblue}{[3] Trees mark the far side of a fence that encloses a large environment space with two giraffes}\\
\textcolor{otherblue}{[4] Several zebras walking the terrain of hills and mountains}\\
\textcolor{advred}{[5] A group of zebras stand together on the rocky minefield.}
\vspace{0.85em}

\end{minipage}
\end{tabularx}

\caption{
\textbf{Qualitative image-to-text retrieval on the RoCOCO \emph{Danger} split.}
For the query image, we retrieve the Top-$5$ captions using the original
CLIP representation and its isotropic- and manifold-smoothed variants at
$\sigma=0.2$. For visualization, each retrieved caption is paired with its
closest matching image. Green denotes the ground-truth caption, red a
RoCOCO adversarial caption, and blue another retrieved caption.
Isotropic smoothing can substantially alter the semantic neighborhood of
the retrieved captions, whereas manifold smoothing retains captions that
remain more closely related to the query.}

\label{fig:rococo_qualitative}
\end{figure}

\paragraph{Effect on inactive encoded concepts.}
As a complementary analysis to Fig.~\ref{fig:sae_concept_comparison}, where concepts
are ranked by their activation after smoothing, we examine the least-active
concepts in the smooth representation while preserving their original representation
ordering. Figure~\ref{fig:sae_least_concept_comparison} shows that isotropic smoothing
can substantially activate several initially inactive concepts, whereas
manifold smoothing keeps their activations close to the original levels.
\input{tikz_fig/fig_concept_least_og}
\label{app:top_least_concepts}
\paragraph{RoCOCO retrieval example.}
Figure~\ref{fig:rococo_qualitative} visualizes how smoothing changes the retrieval semantics. We first retrieve the Top-$5$ captions for
each query representation and then associate each retrieved caption with
its closest matching image for visualization. Under isotropic smoothing,
the retrieved captions can move toward semantically unrelated content,
while manifold smoothing tends to retain captions within the semantic
neighborhood of the original query. The displayed images therefore
visualize the content associated with the retrieved captions; they are
not themselves the direct retrieval targets.
\label{app:rococo_caption_qual_eg}

\end{document}